\documentclass[sigconf, nonacm]{acmart}

\include{amssymb}
\usepackage{bbold}
\usepackage[ruled,linesnumbered,noend]{algorithm2e}
\usepackage{caption}
\usepackage{subcaption}
\usepackage{multirow}
\usepackage{arydshln}
\usepackage{mathtools}
\usepackage[a-2b]{pdfx}

\newcommand{\vect}[1]{\boldsymbol{#1}}

\SetKw{Def}{def}
\SetKwFor{For}{for (}{)}{}
\SetKwFor{Foreach}{foreach (}{)}{}
\SetKwComment{Comment}{$//$}{}
\SetKw{Continue}{continue}
\SetKw{Break}{break}

\newcommand{\rvline}{\hspace*{-\arraycolsep}\vline\hspace*{-\arraycolsep}}
\newcommand{\bigzero}{\mbox{\normalfont\Large\bfseries 0}}

\newcommand\vldbdoi{10.14778/3583140.3583155}
\newcommand\vldbpages{1399 - 1412}
\newcommand\vldbvolume{16}
\newcommand\vldbissue{6}
\newcommand\vldbyear{2023}
\newcommand\vldbauthors{Xiao He, Ye Li, Jian Tan, Bin Wu, Feifei Li}
\newcommand\vldbtitle{\shorttitle} 
\newcommand\vldbavailabilityurl{}
\newcommand\vldbpagestyle{empty} 

\begin{document}
\title{OneShotSTL: One-Shot Seasonal-Trend Decomposition For Online Time Series Anomaly Detection And Forecasting}

\author{Xiao He}
\affiliation{
		\institution{Alibaba Group}
	}
\email{xiao.hx@alibaba-inc.com}

\author{Ye Li}
\affiliation{
	\institution{Alibaba Group}
}
\email{liye.li@alibaba-inc.com}

\author{Jian Tan}
\affiliation{
	\institution{Alibaba Group}
}
\email{j.tan@alibaba-inc.com}

\author{Bin Wu}
\affiliation{
	\institution{Alibaba Group}
}
\email{binwu.wb@alibaba-inc.com}

\author{Feifei Li}
\affiliation{
	\institution{Alibaba Group}
}
\email{lifeifei@alibaba-inc.com}


\begin{abstract}
Seasonal-trend decomposition is one of the most fundamental concepts in time series
analysis that supports various downstream tasks, including 
time series anomaly detection and forecasting. However, existing
decomposition methods rely on batch processing with a time complexity
of $O(W)$, where $W$ is the number of data points within a time
window. Therefore, they cannot always efficiently
support real-time analysis that demands low processing delay.
To address this challenge, we propose OneShotSTL, an efficient and accurate
algorithm that can decompose time series online with an update time
complexity of $O(1)$. OneShotSTL is more than $1,000$ times faster than the batch
methods, with accuracy comparable to the best counterparts. 
Extensive experiments on real-world benchmark
datasets for downstream time series anomaly detection and forecasting tasks
demonstrate that OneShotSTL is from $10$ to over $1,000$ times faster
than the state-of-the-art methods, while
still providing comparable or even better accuracy.
\end{abstract}

\maketitle

\pagestyle{\vldbpagestyle}
\begingroup\small\noindent\raggedright\textbf{PVLDB Reference Format:}\\
\vldbauthors. \vldbtitle. PVLDB, \vldbvolume(\vldbissue): \vldbpages, \vldbyear.\\
\href{https://doi.org/\vldbdoi}{doi:\vldbdoi}
\endgroup
\begingroup
\renewcommand\thefootnote{}\footnote{\noindent
Xiao He is the corresponding author.\newline
This work is licensed under the Creative Commons BY-NC-ND 4.0 International License. Visit \url{https://creativecommons.org/licenses/by-nc-nd/4.0/} to view a copy of this license. For any use beyond those covered by this license, obtain permission by emailing \href{mailto:info@vldb.org}{info@vldb.org}. Copyright is held by the owner/author(s). Publication rights licensed to the VLDB Endowment. \\
\raggedright Proceedings of the VLDB Endowment, Vol. \vldbvolume, No. \vldbissue\ %
ISSN 2150-8097. \\
\href{https://doi.org/\vldbdoi}{doi:\vldbdoi} \\
}\addtocounter{footnote}{-1}\endgroup

\ifdefempty{\vldbavailabilityurl}{}{
}

\section{Introduction}
\label{sec:intro}
With the rapid advancement in data collection and storage techniques, time
series are witnessing an enormous increase in
applications, especially in areas such as Internet-of-Things (IoT~\cite{iot}) and IT
Operations (AIOps~\cite{aiops}).  For instance, cloud database systems
collect large sets of metrics,including CPU usage
and request rate, to facilitate the monitoring of complex systems.
Driven by this trend, cloud companies (e.g., AWS
Cloudwatch~\cite{cloudwatch} and
Alibaba CloudMonitor~\cite{cloudmonitor}) have provided online 
services to improve the observability of servers and applications for their users.

From an algorithmic perspective, 
real-time analysis tasks on Time Series fall into either Anomaly Detection (TSAD)
or Forecasting (TSF).  For a time series $\{y_1,y_2,...,y_t,...\}$, at
each time point $t$, an online TSAD method outputs an anomaly score
for the current data $y_t$, while an online TSF method predicts the
future value of data $\hat{y}_{t+i}$ at time $t+i$ for $i>0$.  The
former is used to identify performance issues of the system based on
historical observations (e.g., through alerting or diagnosis), while
the latter is to predict the future to prevent such issues by taking
certain actions in advance (e.g., by automatically scaling up the
computation resources). The significant volume of data generated from
large-scale systems, together with the requirement of low service
delays, impose great challenges to TSAD and TSF, due to the online
nature of the processing.

There have been a large number of
TSAD~\cite{DBLP:conf/kdd/AudibertMGMZ20,DBLP:journals/vldb/BoniolLRPMR21,DBLP:journals/datamine/YehZUBDDZSMK18,DBLP:conf/kdd/Lu00ZK22,DBLP:journals/pvldb/TuliCJ22}
and
TSF~\cite{10.1109/UKSim.2014.67,DBLP:journals/peerjpre/TaylorL17,salinas2020deepar,DBLP:conf/iclr/OreshkinCCB20,zhou2021informer,DBLP:conf/icml/ZhouMWW0022}
methods proposed in the literature. In this paper, we focus on an
important class of algorithms based on Seasonal-Trend Decomposition
(STD). STD is one of the most fundamental operations in time series analysis, which has been extensively studied and used in
economics and finance~\cite{cleveland90,dokumentov2015str}. 
Many studies have
shown that STD methods are effective for
TSAD~\cite{dokumentov2015str,DBLP:journals/corr/abs-1812-01767,DBLP:conf/kdd/WenZL020}
and TSF~\cite{de2011forecasting,DBLP:conf/iclr/OreshkinCCB20}, since very often practical time
series data exhibit trends and seasonal
patterns~\cite{DBLP:journals/corr/abs-1812-01767}. However, most of
the existing STD methods, e.g.,
STL~\cite{cleveland90} and
RobustSTL~\cite{DBLP:journals/corr/abs-1812-01767}, are based on batch
processing, and thus essentially offline by nature, which does not
support incremental updates. The only exception is a recent method
OnlineSTL~\cite{DBLP:journals/pvldb/MishraSZ22}, which decomposes time
series in an online fashion by effectively utilizing the fast
computation from explicit mathematical closed-form
expressions. Remarkably, OnlineSTL can process more than $100$ times
faster than traditional STD methods. However, 
only simple trend and seasonal filtering
methods (e.g., tri-cube and exponential smoothing) are
supported. Therefore, the decomposition quality of OnlineSTL is worse
than its full-fledged batch counterparts, e.g.,
RobustSTL~\cite{DBLP:journals/corr/abs-1812-01767}, which introduces
$l_1$-norm regularizer.
Moreover,
although OnlineSTL is fast, its single update complexity
is still $O(T)$ with $T$ being the length of the seasonal
period. Thus, its performance degrades with the increase of $T$. 

To address the aforementioned challenges, in this paper, we propose a
new online STD algorithm, termed OneShotSTL, that significantly
accelerates the decomposition with an update complexity of $O(1)$. 
The needed information from the past observations to update the effect by observing a newly arrived data point is characterized by a fixed number of state parameters of OneShotSTL, whose values are carried forward as the online processing keeps going.  
Thus, OneShotSTL processes each data point only once. 
To achieve high accuracy, we adopt a $l_1$-norm regularizer
and adaptively adjust the seasonal component in our online
computation. Notably, OneShotSTL can yield accurate decomposition
results comparable to  RobustSTL~\cite{DBLP:journals/corr/abs-1812-01767} (the best batch method that we are aware of in the literature)
and in the
meanwhile can greatly reduce the processing time.  
Extensive experiments on both TSAD and TSF tasks demonstrate the superior performance of OneShotSTL.



\noindent\textbf{Contributions.} This work makes the following contributions:
\begin{itemize}
	\item To the best of our knowledge, the proposed OneShotSTL is
          the first online STD method with an update complexity of
          $O(1)$, which greatly improves the existing ones
          ($O(T)$).
	
	\item OneShotSTL is ultra-fast in practice. It takes around
          $20 \mu s$ to process each data point on a typical commodity
          server 
          using a single CPU core, which is more than $1,000$ times
          faster than the batch STD methods and $20$ times faster than
          OnlineSTL on datasets with a seasonal period of $12,800$.
	
	
	\item OneShotSTL is also accurate.
		  Extensive experiments on benchmark datasets for TSAD and
          TSF show that OneShotSTL is from $10$ to more than $1,000$
          times faster than the SOTA methods with
          comparable or even better accuracy.
\end{itemize}

\begin{table}[t]
	\caption{Comparison of different STD algorithms}
	\vspace{-4mm}
	\begin{center}
		\small
		\begin{tabular}{lllllll}
			\hline
			\multirow{2}{*}{Algorithm} & Batch or & Trend  & Seasonality & Online \\
			& Online  & Change & Shifts  & Complexity\\
			\hline    	
			STL \cite{cleveland90}& Batch &No&No&-\\
			TATBS \cite{de2011forecasting}& Batch &No&No&-\\
			STR \cite{dokumentov2015str}& Batch &No&No&  -\\
			RobustSTL \cite{DBLP:journals/corr/abs-1812-01767}& Batch &Yes&Yes& -\\			
			FastRobustSTL \cite{DBLP:conf/kdd/WenZL020}& Batch &Yes&Yes&- \\
			\hline
			OnlineRobustSTL \cite{sreworks}&Online&Yes&Yes & O(T)\\					
			OnlineSTL \cite{DBLP:journals/pvldb/MishraSZ22}&Online&No&No & O(T)\\		
			\textbf{OneShotSTL}&\textbf{Online}&\textbf{Yes}&\textbf{Yes}& $\textbf{O(1)}$\\				
			\hline		
		\end{tabular}
	\end{center}
	\vspace{-4mm}
	\label{tab:comparison}
\end{table}

\section{Problem Setting and Related Works}
\label{preliminaries}
Since our online STD framework is derived from the batch STD, we
formally introduce both the batch and the online STD problems.  We also review the related existing
works in this area. Throughout the paper, we use bold
lowercase/capital letters to represent vectors/matrices, respectively.

\subsection{Batch STD}\label{ss:batch}
For batch processing, consider the following model with trend and
seasonality for a time series $\vect{y} \in \mathbb{R}^N$ of a finite
batch size $N$:
\begin{align}
	\label{eq:stl}		
	y_i = \tau_i + s_i + r_i, i=1,2,...N,
\end{align}
where $y_i$, $\tau_i$, $s_i$ and $r_i$ denote the observation, the
decomposed trend, seasonal and residual components at time point $i$,
respectively. We only consider the decomposition of data with a single
seasonal cycle of length $T < N$, which can be readily
estimated by period detection methods~\cite{DBLP:conf/sigmod/WenH0ZKX21}.
This can be easily extended to the case with multiple seasons as shown
in previous
studies~\cite{DBLP:journals/corr/abs-1812-01767,DBLP:conf/kdd/WenZL020}. In
this paper, we treat data with multiple seasons as a single seasonal
sequence with $T$ being the longest seasonal period.

\subsection{Online STD}
For online processing, data points arrive in a streaming fashion,
denoted by $\{y_1,y_2,...,y_t,...\}$ where the sequence can be even
unbounded. 
At time $t$, based on all
the past observations $\{y_1,...,y_{t-1}\}$ and the latest one $y_t$,
online methods only decompose $y_t$ into trend $\tau_t$, seasonal
$s_t$ and residual $r_t$ with $y_t = \tau_t + s_t + r_t$. Similarly to
OnlineSTL~\cite{DBLP:journals/pvldb/MishraSZ22}, we divide the online
processing into two phases, a single offline initialization phase and
a lasting online phase. Formally, at time $t$ we observe
$\{y_1,y_2,...,y_{t_0},y_{t_0+1},...,y_t\}$. The first $t_0$ points
$\{y_1,...,y_{t_0}\}$ are used for initialization.
The online processing starts from time $t_0+1$ and keeps computing $\tau_t$, $s_t$ and $r_t$ for $t
> t_0$ by using only data points observed before $t$. The seasonal cycle $T$, as a fixed parameter, can be estimated from $\{y_1,...,y_{t_0}\}$ using a period detection method, e.g.,~\cite{DBLP:conf/sdm/VlachosYC05,DBLP:conf/sigmod/WenH0ZKX21}. 
During the online computation phase, as data points keep arriving, we observe new seasonal periods, and each has its own cycle length $T_c$. These cycle lengths are not necessarily equal.  
We assume that each $T_c$ of the underlying cycle lengths from the streaming data sequence, although unknown, only slightly deviates from $T$. Specifically, $T_c$ can be an arbitrary integer value within a small interval decided by a hyper-parameter $H$, i.e.,  $T_c \in [T-H, T+H]$. This assumption prevents cycle lengths from varying drastically, e.g., $T_c$ gradually changing from $1000$ to $2000$ across multiple periods. A more general treatment of the varying cycle lengths deserves a dedicated study given the vast existing literature on this topic~\cite{DBLP:conf/sdm/VlachosYC05,DBLP:conf/pkdd/PuechBDM19,DBLP:journals/datamine/TollerSK19,DBLP:conf/sigmod/WenH0ZKX21}. 

\vspace{-6mm}
\subsection{Existing STD Methods}
Batch STD has been extensively studied in economics and finance for
decades, with a variety of algorithms proposed in the literature
\cite{cleveland90,de2011forecasting,dokumentov2015str,DBLP:journals/corr/abs-1812-01767,DBLP:journals/pvldb/MishraSZ22}. Among
them, the most popular one is STL~\cite{cleveland90}, which adopts an
alternating algorithm to estimate the trend and seasonal components
using LOESS (LOcal regrESSion) smoothing. Later,
TATBS~\cite{de2011forecasting} and STR~\cite{dokumentov2015str} are
proposed with confidence intervals of the decomposition
results. However, they need extensive computations, especially for
data with long seasonality.  In addition, these methods may encounter
problems in handling complex patterns, e.g., with abrupt changes of
trend (see Figure~\ref{fig:synthetic}) and seasonality shift (see
Figure~\ref{fig:seasonality_shift}), which are common in metrics for
monitoring in AIOps. As a way to improve these issues,
RobustSTL~\cite{DBLP:journals/corr/abs-1812-01767} and its accelerated
version FastRobustSTL~\cite{DBLP:conf/kdd/WenZL020} are proposed using
$l_1$-norm trend filtering and non-local seasonal filtering.

All these batch methods can be used for the online setting on a sliding
window of length $W>T$ that contains the most recent data points. A
common approach is to set $W$ equal to the length of a few seasonal
cycles, e.g., $W=4T$.
Thus, the time
complexity of these batch methods is at least $O(T)$. However, as
pointed out in the recent paper
OnlineSTL~\cite{DBLP:journals/pvldb/MishraSZ22}, most of these methods
cannot always be directly and efficiently utilized in real-time
applications due to long processing delays. OnlineSTL is the first
online decomposition algorithm that alternatively applies tri-cube
(trend) and exponential smoothing (seasonal) filtering, which utilizes
fast computation from explicit mathematical closed-form
expressions. OnlineSTL focuses on speed and can process $10,000$ time
series data points per second using a single CPU core on a typical
commodity server. However, it sacrifices quality/accuracy by using simple
trend and seasonal filtering and cannot deal with complex patterns.
More importantly, the online update complexity of OnlineSTL and other
existing online algorithms, e.g., online
RobustSTL~\cite{DBLP:conf/kdd/WenZL020,sreworks} (abbreviated as
OnlineRobustSTL), is still $O(T)$. Therefore, the
performance degrades as the length of the seasonality cycle increases.
 

We summarize the comparisons of batch and online STD methods in Table \ref{tab:comparison}, including our proposed OneShotSTL. Note that OneShotSTL is the only work with $O(1)$ update
complexity.

\section{Model}
\label{sec:model}
\begin{figure}[t]
	\vspace{-2mm}	
	\includegraphics[width=0.41\textwidth]{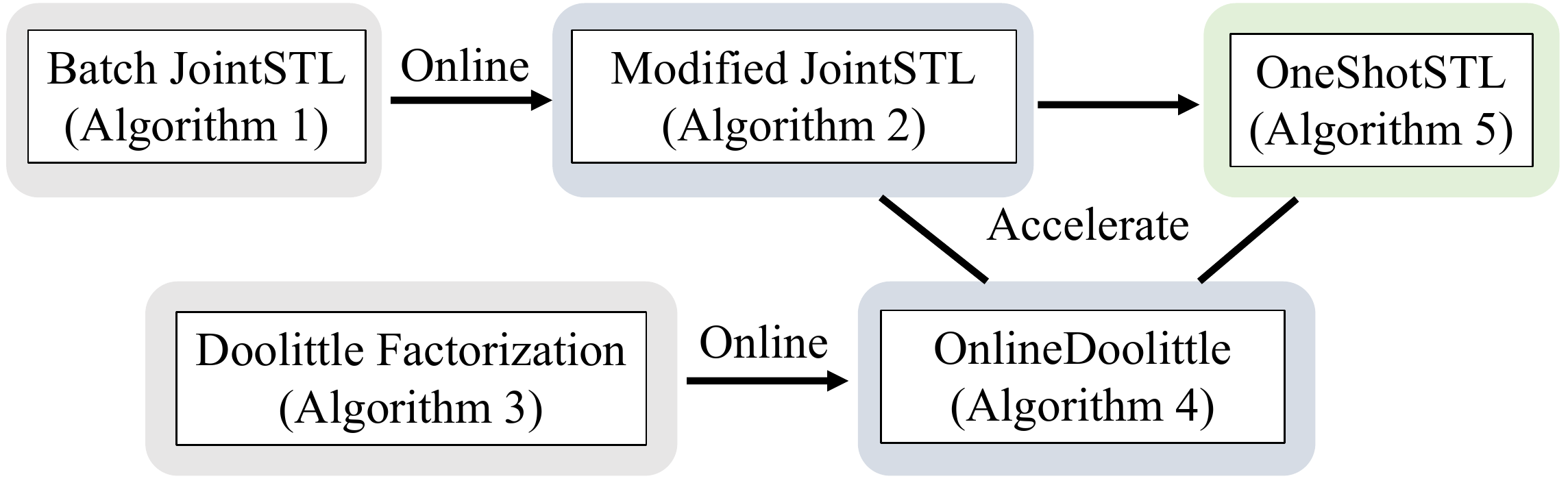}		
	\vspace{-2mm}
	\caption{Roadmap of developing OneShotSTL algorithm.}
	\vspace{-4mm}	
	\label{fig:roadmap}
\end{figure}

In this section, we begin with a new batch model formulation, JointSTL (Algorithm \ref{algo:JointSTL}), that derives seasonal and trend signals simultaneously in Section \ref{sec:jointopt}. 
This batch framework is the starting point for us to develop an efficient online algorithm. 
However, requiring an online algorithm to exactly solve the batch JointSTL problem is difficult. To this end, in Section \ref{sec:modified}, we introduce certain approximations, which pave the road towards an online algorithm by modifying JointSTL (Algorithm \ref{algo:ModifiedSTL}). Then, Section~\ref{sec:OneShotSTL} 
presents OneShotSTL (Algorithm \ref{algo:OneShotSTL}) that provides exactly the same results as Algorithm \ref{algo:ModifiedSTL} but greatly accelerates the computation by using an online Doolittle factorization algorithm (Algorithm \ref{algo:online_Doolittle}) with $O(1)$ complexity. Last, we describe the method to handle seasonality shift in Section \ref{sec:shift}. Figure \ref{fig:roadmap} depicts the roadmap for developing the OneShotSTL algorithm. To simplify the presentation, we first consider the case $H=0$ (i.e., $T_c=T$) in Sections \ref{sec:jointopt}, \ref{sec:modified},  and \ref{sec:OneShotSTL}.  In this case, we may abuse the notation and use $T$ and $T_c$ interchangeably. 
The general case when $T_c$ can be any integer in $[T-H, T+H], H\geq 1$ is investigated in Section \ref{sec:shift}.

\subsection{JointSTL for Batch Setting }
\label{sec:jointopt}
Given a time series $\vect{y} \in \mathbb{R}^N$ and the input seasonal period parameter $T$, we model the batch STD problem of JointSTL as follows:
\begin{align}
	\begin{aligned}		
		\min_{\vect{\tau}, \vect{s}} &\left(\sum_{t=1}^N \left(\tau_t + s_t - y_t\right)^2 + \sum_{t=T}^N \left(s_t - s_{t-T}\right)^2\right.  \\
		&\left. + \lambda_1 \sum_{t=2}^N \left|\tau_t - \tau_{t-1}\right| + \lambda_2 \sum_{t=3}^N \left|\tau_t - 2\tau_{t-1} +\tau_{t-2}\right|\right)
		\label{eq:joint_opt}	
	\end{aligned}		
\end{align}
To improve robustness, the $l_1$-norm is used on the first and second order of differences on the trend components with $\lambda_1$ and $\lambda_2$ as hyper-parameters to control the smoothness. 

JointSTL can be seen as a simple extension of the famous $l_1$ trend filtering~\cite{kim2009ell_1}. In other words, it reduces to $l_1$ trend filtering by setting $\vect{s}=\vect{0}$. To solve the optimization problem (\ref{eq:joint_opt}), we adopt the Iteratively Reweighted Least Square (IRLS)~\cite{10.2307/2345503} method, which plays an important role in further extensions to the online setting. 
Following the idea of IRLS~\cite{10.2307/2345503}, we introduce auxiliary variables $\vect{p} \in \mathbb{R}^{N-1}$, $\vect{q} \in \mathbb{R}^{N-2}$ and rewrite Problem (\ref{eq:joint_opt}) as:
\begin{align}
	\vspace{-4mm}
	\begin{aligned}
		\min_{\vect{\tau}, \vect{s}, \vect{p}, \vect{q}} & \left( \sum_{t=1}^N \left( \tau_t + s_t - y_t \right)^2 +\sum_{t=T}^N \left( s_t - s_{t-T} \right)^2 \right.\\
		&+ \lambda_1 \sum_{t=2}^N \left(p_t\left( \tau_t - \tau_{t-1} \right)^2 + \frac{1}{4}p_t^{-1} \right)	 \\
		&\left. + \lambda_2 \sum_{t=3}^N \left(q_t\left( \tau_t - 2\tau_{t-1} + \tau_{t-2} \right)^2 + \frac{1}{4}q_t^{-1} \right)\right)
		\label{eq:joint_opt_aux}		
	\end{aligned}				
\end{align}
where $p_t$ and $q_t$ denotes the value of $\vect{p}$ and $\vect{q}$ at time $t$.

Problem (\ref{eq:joint_opt_aux}) is equivalent to Problem (\ref{eq:joint_opt}) with respect to $\vect{\tau}$ and $\vect{s}$ when $p_t$ and $q_t$ are calculated as follows:
\begin{align}
	\label{eq:pt}
	p_{t} &= \frac{1}{2\left|\tau_t-\tau_{t-1}\right|} 
\end{align}
\begin{align}
	\label{eq:qt}
	q_{t} &= \frac{1}{2\left|\tau_t-2\tau_{t-1}+\tau_{t-2}\right|}
\end{align}
This can be simply proved by substituting Eq. (\ref{eq:pt}) and (\ref{eq:qt}) into Problem (\ref{eq:joint_opt_aux}) to obtain Problem (\ref{eq:joint_opt}).

Problem (\ref{eq:joint_opt_aux}) can be solved by alternating optimization, which is shown in Algorithm \ref{algo:JointSTL}. 
According to~\cite{beck2015convergence,DBLP:conf/aaai/HeAS19} on IRLS,
Algorithm \ref{algo:JointSTL} converges to a global optimum of Problem (\ref{eq:joint_opt}). 
When $\vect{\tau}$ and $\vect{s}$ are fixed, the update rules are shown in Eq. (\ref{eq:pt}) and (\ref{eq:qt}). When $\vect{p}$ and $\vect{q}$ are fixed,  we concatenate $\vect{\tau}$ and $\vect{s}$ to get $\vect{x} = [\vect{\tau};\vect{s}]$ and substitute it into Problem (\ref{eq:joint_opt_aux}),  resulting in a least square problem with its optimal solution determined by a linear system $\vect{A}\vect{x} = \vect{b}$ where
\begin{align}
	\begin{aligned}	
		\vect{A} &= \vect{B_1}^T\vect{B_1} + \vect{B_2}^T\vect{B_2} + \lambda_1 (\vect{p}\cdot \vect{B_3})^T\vect{B_3} + \lambda_2 (\vect{q}\cdot \vect{B_4})^T\vect{B_4}
		\\		
		\vect{b} &= \vect{B_1}^T\vect{y}
		\\
		\vect{B_1} &= 
		\begin{bmatrix}
			\begin{matrix}
				1 & & \\
				& \ddots & \\
				& & 1
			\end{matrix}
			& \rvline &
			\begin{matrix}
				1 & & \\
				& \ddots & \\
				& & 1
			\end{matrix}	  
		\end{bmatrix} \in \mathbb{R}^{N\times 2N}
		\\
		\vect{B_2} &=
		\begin{bmatrix}
			\bigzero
			& \rvline &
			\begin{matrix}
				-1 & \cdots & 1& \\
				& \ddots & & \ddots & \\
				& & -1 & \cdots & 1
			\end{matrix}				
		\end{bmatrix} \in \mathbb{R}^{(N-T)\times 2N}	
		\\
		\vect{B_3} &= 
		\begin{bmatrix}
			\begin{matrix}
				1 & -1& &\\
				&  \ddots & \ddots  &  \\
				& & 1 & -1 
			\end{matrix}
			& \rvline &
			\bigzero
		\end{bmatrix} \in \mathbb{R}^{(N-1)\times 2N}
		\\
		\vect{B_4} &= 
		\begin{bmatrix}
			\begin{matrix}
				1 & -2 & 1 & &\\
				&  \ddots & \ddots & \ddots  &  \\
				& & 1 & -2 & 1 
			\end{matrix}
			& \rvline &
			\bigzero
		\end{bmatrix} \in \mathbb{R}^{(N-2)\times 2N}
	\end{aligned}	
	\label{eq:BatchABCD}
\end{align}

In Problem (\ref{eq:joint_opt}), we use the squared $l_2$-norm in the first two terms for the residual and seasonal components. In principle, they can be replaced by the $l_1$-norm as in RobustSTL. However, using the squared $l_2$-norm for the first term (residual) and $l_1$-norm for the trend terms is consistent with the classic $l_1$ 
trend filtering~\cite{kim2009ell_1} and also our practical evaluations show more robust results with (\ref{eq:joint_opt}).
\begin{algorithm}[t]
	\small
	\caption{Batch JointSTL}
	\label{algo:JointSTL}
	\KwIn{$\vect{y} \in \mathbb{R}^{N}$, $T$, $\lambda$,  maximum iterations $I$}
	\KwOut{$\vect{\tau}$, $\vect{s}$, $\vect{r}$}
	$\vect{p}=\vect{q}=\vect{1}$\;
	\For{i = 0; i < I; i ++}
	{
		Construct $\vect{A}$ and $\vect{b}$ based on Eq. (\ref{eq:BatchABCD})\;
		Obtain $\vect{x}$ by solving the linear system $\vect{A}\vect{x}=\vect{b}$\;
		$\vect{\tau}$ = $\vect{x}_{1:N}$;	$\vect{s}$ = $\vect{x}_{N:2N}$; $\vect{r}=\vect{y}-\vect{\tau}-\vect{s}$\;
		Update $\vect{p}$, $\vect{q}$ based on Eq. (\ref{eq:pt}) and (\ref{eq:qt})\;		
	}	
\end{algorithm}

\subsection{Modified JointSTL for Online Setting}
\label{sec:modified}
Problem (\ref{eq:joint_opt}) and Algorithm \ref{algo:JointSTL} are designed for the batch setting. In this section, we modify them to adapt to the online setting.

For the online setting, the whole data processing procedure consists of an initialization phase and an online updating phase. Specifically, for a data sequence $\{y_1,y_2,...,y_{t_0},y_{t_0+1},...,y_t\}$, the first $t_0$ data points $\{y_1,y_2,...,y_{t_0}\}$ are used for initialization, $t_0\geq W$, e.g., $W=4T$ with $T$ being the longest seasonality period as in OnlineSTL. We can apply a batch model, e.g., STL,  processing on $\{y_1,y_2,...,y_{t_0}\}$ and obtain $\{\tau_1,\tau_2,...,\tau_{t_0}\}$ and $\{s_1,s_2,...,s_{t_0}\}$. Therefore, when decomposing the first data point $t_0+1$
in the online phase, we already have the seasonal component of the previous data points. Further, after each online decomposition, we can always keep the most recent period of the seasonal component in a sliding window. We name the seasonal buffer as $\vect{v} \in \mathbb{R}^T$ and initialize it with $\vect{v}=[s_{t_0-T},...,s_{t_0}]^T$. 
 Then we modify the second term of Problem (\ref{eq:joint_opt}) on seasonal component and rewrite the problem as:
\begin{align}
	\begin{aligned}	
		\min_{\vect{\tau}, \vect{s}} & \left( \sum_{j=t_0+1}^{t_0+M} \left(\tau_j + s_j - y_j\right)^2 + \sum_{j=t_0+1}^{t_0+M} \left(s_j - v_{j \% T}\right)^2 \right.\\
		&\left. + \lambda_1 \sum_{j=t_0+1}^{t_0+M} \left|\tau_j - \tau_{j-1}\right| + \lambda_2 \sum_{j=t_0+1}^{t_0+M} \left|\tau_j - 2\tau_{j-1} +\tau_{j-2}\right|\right)
		\label{eq:joint_opt_online}	
	\end{aligned}
\end{align}
where for each newly arrived data point $y_t$ at timestamp $t=t_0+M$, we decompose all the $y_j$ into $\tau_j$ and $s_j$ for $j \in [t_0+1,t_0+M]$ but only output the last decomposition result $\tau_{t_0+M}$ and $s_{t_0+M}$. 

Similarly, we can introduce auxiliary variables $\vect{p} \in \mathbb{R}^{M}$ and $\vect{q} \in \mathbb{R}^{M}$ and use IRLS~\cite{10.2307/2345503} for solving Problem (\ref{eq:joint_opt_online}). The update rules of $p_t$ and $q_t$ are the same as Eq. (\ref{eq:pt}) and (\ref{eq:qt}). But when $\vect{p}$ and $\vect{q}$  are fixed, we solve a different linear system to derive $\vect{\tau}$ and $\vect{s}$. Specifically, we concatenate $\vect{\tau}$ and $\vect{s}$ to form $\vect{x} = \{\tau_{t_0+1},s_{t_0+1},...,\tau_{t_0+M},s_{t_0+M}\}$.
The optimal $\vect{x}$ is the solution of a linear system $\vect{A}\vect{x} = \vect{b}$, satisfying
\begin{align}
	\begin{aligned}
		\vect{A} &= \vect{C_1}^T\vect{C_1} + \vect{C_2}^T\vect{C_2} + \lambda_1 (\vect{p}\cdot \vect{C_3})^T\vect{C_3} + \lambda_2 (\vect{q}\cdot \vect{C_4})^T\vect{C_4} 
		\\
		\vect{b} &= \vect{C_1}^T\vect{y} + \vect{C_2}^T\vect{u}, \text{where } \vect{u} = [\vect{v}^T,...,\vect{v}^T]^T \in \mathbb{R}^M
		\\
		\vect{C_1} &= 
		\begin{bmatrix}
			1 & 1 \\
			&        & \ddots \\
			&        &              & 1 & 1
		\end{bmatrix} \in \mathbb{R}^{M\times 2M}
		\\
		\vect{C_2} &= 
		\begin{bmatrix}
			0 & 1 \\
			&        & \ddots \\
			&        &              & 0 & 1
		\end{bmatrix} \in \mathbb{R}^{M\times 2M} 
		\\
		\vect{C_3} &=
		\begin{bmatrix}
			1  & 0 & -1 \\
			&  \ddots  & \ddots & \ddots \\
			&     &   1& 0 & -1  
		\end{bmatrix} \in \mathbb{R}^{(M-1)\times 2M}	
		\\
		\vect{C_4} &=
		\begin{bmatrix}
			1  & 0 & -2 & 0 & 1 \\
			&  \ddots & \ddots & \ddots & \ddots & \ddots \\
			&     &   1& 0 & -2 & 0 & 1  
		\end{bmatrix} \in \mathbb{R}^{(M-2)\times 2M}
	\end{aligned}
\label{eq:online_ABCD}
\end{align}

\begin{algorithm}[t]
	\caption{Modified Batch JointSTL for online setting}
	\label{algo:ModifiedSTL}
	\Comment{Initialization phase}
	\KwIn{$y_1,...,y_{t_0}$, $T$, $\lambda$, maximum iterations $I$}
	Obtain $\vect{\tau}, \vect{s}, \vect{r}$ by STL or JointSTL; 	Set $\vect{v}=[s_{t_0-T},...,s_{t_0}]$\;
	$\vect{p}^0=\vect{q}^0=[1]$; $\vect{p}^i=\vect{q}^i=[]$, for $i \in [1,I]$\;	
	
	\vspace{2mm}
	\Comment{Online phase}
	\KwIn{$\vect{y}=\{y_1,y_2,...,y_{t}\}$, $T$, $\lambda$, $I$}
	\KwOut{$\tau_t$, $s_t$, $r_t$}
	\For{i = 0; i < I; i ++}
	{
		Construct $\vect{A}$ and $\vect{b}$ based on Eq. (\ref{eq:online_ABCD}) using $\vect{p}^i, \vect{q}^i$\;
		Obtain $\vect{x}$ by solving the linear system $\vect{A}\vect{x}=\vect{b}$\;
		$\tau_t = x_{-2}$; $s_t = x_{-1}$; $r_t=y_t-\tau_t-s_t$\;
		Update $p_t$ and $q_t$ based on Eq. (\ref{eq:pt}) and (\ref{eq:qt})\;		
		$\vect{p}^{i+1}=[\vect{p}^{i+1}, p_t];\vect{q}^{i+1}=[\vect{q}^{i+1}, q_t]$\;
	}	
	$\vect{p}^0=[\vect{p}^0, 1];\vect{q}^0=[\vect{q}^0, 1]$\;
	$v_{t\%T}=s_t$\;
\end{algorithm}

However, the above method
still solves a batch problem. Moreover, it is difficult to derive an efficient online algorithm based on it.
Note that, in the online setting, we target the decomposition of the newly arrived data point.
Thus, we modify JointSTL for the online setting that can well approximate the last decomposition, as shown in Algorithm \ref{algo:ModifiedSTL}.

During the initialization phase of Algorithm \ref{algo:ModifiedSTL}, we use STL or JointSTL on $\{y_1,...,y_{t_0}\}$ and set $\vect{v}$ as introduced before. Then we set 
$\vect{p}=\vect{p}^i$ and $\vect{q}=\vect{q}^i$ for the $i$-th iteration, which will be used for constructing $\vect{A}$ and $\vect{b}$ in Eq. (\ref{eq:online_ABCD}). Because, during the online phase, we have different vectors of $\vect{p}^i$ and $\vect{q}^i$ updated using Eq. (\ref{eq:pt}) and (\ref{eq:qt}) for each iteration $i$. Unlike JointSTL which updates the whole vector of $\vect{p}$ and $\vect{q}$, while decomposing $y_t$, we only append $\vect{p}^i$ and $\vect{q}^i$ with the latest $p_t$ and $q_t$ calculated from the decomposition of $\tau_t$ and $s_t$ for iteration $i$. Finally, we output the decomposition results from the last iteration. In practice (see experiments in Section~\ref{sec:experiment}), we find that this approximation provides sufficiently good decomposition.


However, Algorithm \ref{algo:ModifiedSTL} is still unrealistic since for each time point $t=t_0+M$, one needs to solve a linear system $\vect{A}\vect{x}=\vect{b}$ in each iteration of Algorithm \ref{algo:ModifiedSTL} (lines 4-8), where $\vect{A} \in \mathbb{R}^{2M \times 2M}$ and its size is increasing while processing more data points online. Recall that in the online setting, at time point $t$, one only needs to partially solve $\vect{A}\vect{x}=\vect{b}$ and outputs the latest decomposition $\tau_t$ and $s_t$. Moreover, from Eq. (\ref{eq:online_ABCD}) we can see that $\vect{A}$ is a symmetric banded matrix with constant bandwidth of $9$ in regardless of the growing of $M$ and the values of $\vect{p}$ and $\vect{q}$, since $\vect{C_1}^T\vect{C_1}$, $\vect{C_2}^T\vect{C_2}$, $\vect{C_3}^T\vect{C_3}$ and $\vect{C_4}^T\vect{C_4}$ are all banded matrices with constant bandwidth of $3$, $3$, $5$ and $9$, respectively, for any $M\geq 5$. Therefore, the linear system $\vect{A}\vect{x}=\vect{b}$ can be efficiently solved using Cholesky/Doolittle factorization and Gaussian elimination. The crux of the matter is that the Cholesky/Doolittle factorization and the forward substitution of Gaussian elimination can be made online. Specifically, at each time point and for each iteration of Algorithm \ref{algo:ModifiedSTL}, we only need to conduct the first few steps of the backward substitution and output the last two elements of $\vect{x}$, i.e., $\tau_t$ and $s_t$. 
Following this idea, we propose OneShotSTL detailed in the next section.

\subsection{OneShotSTL Aglorithm}
\label{sec:OneShotSTL} 
This section presents the online symmetric Doolittle factorization algorithm with $O(1)$ update complexity to accelerate Algorithm \ref{algo:ModifiedSTL}, which leads to the full OneShotSTL algorithm. 

\subsubsection{\textbf{Online Symmetric Doolittle Factorization}}
For each time point $t$, Algorithm \ref{algo:ModifiedSTL} needs to sequentially solve $I$ different linear systems of  $\vect{A}\vect{x}=\vect{b}$, where $I$ is a configured parameter on the number of iterations. Note that this number $I$ is prefixed and independent of the input data size, e.g., $I=8$.  The reason is that the optimization of Problem (\ref{eq:joint_opt_online}) using Algorithm \ref{algo:ModifiedSTL} based on IRLS usually converges very fast~\cite{10.2307/2345503}.
Therefore, we only need a few iterations to obtain satisfactory results (see experiments in Section~\ref{sec:experiment}). This algorithm has two steps. The first step is to initialize the first iteration, and the second step is to repeat the rest $I-1$ iterations based on the result from the previous iteration.

\begin{algorithm}[htb!]
	\small
	\caption{Symmetric Doolittle Factorization}
	\label{algo:Doolittle}
	\KwIn{$\vect{A} \in \mathbb{R}^{M\times M}$}
	\KwOut{$\vect{L}\in \mathbb{R}^{M\times M}$, $\vect{D} \in \mathbb{R}^{M\times M}$}
	\For {k = 1; k < M; k ++} 
	{
		$L_{kk} = 1$\;
		$D_{kk} = A_{kk} - \sum_{i=1}^{k-1}D_{ii}\cdot L_{ki}^2$\;
		\For {j = k + 1; j < M; j ++} 
		{
			$L_{kj} = 0$\;
			$L_{jk} = (A_{jk} - \sum_{i=1}^{k-1}L_{ji}\cdot D_{ii}\cdot L_{ki})~/~D_{kk}$\;
		}			
	}
\end{algorithm}

\begin{figure}[htb!]
	\centering
	\vspace{-3mm}
	\begin{subfigure}[t]{0.212\textwidth}
		\caption{$\vect{A^t}$ ($M=5$),~$\vect{A^o} \in \mathbb{R}^{4\times 4}$}
		\includegraphics[width=\textwidth]{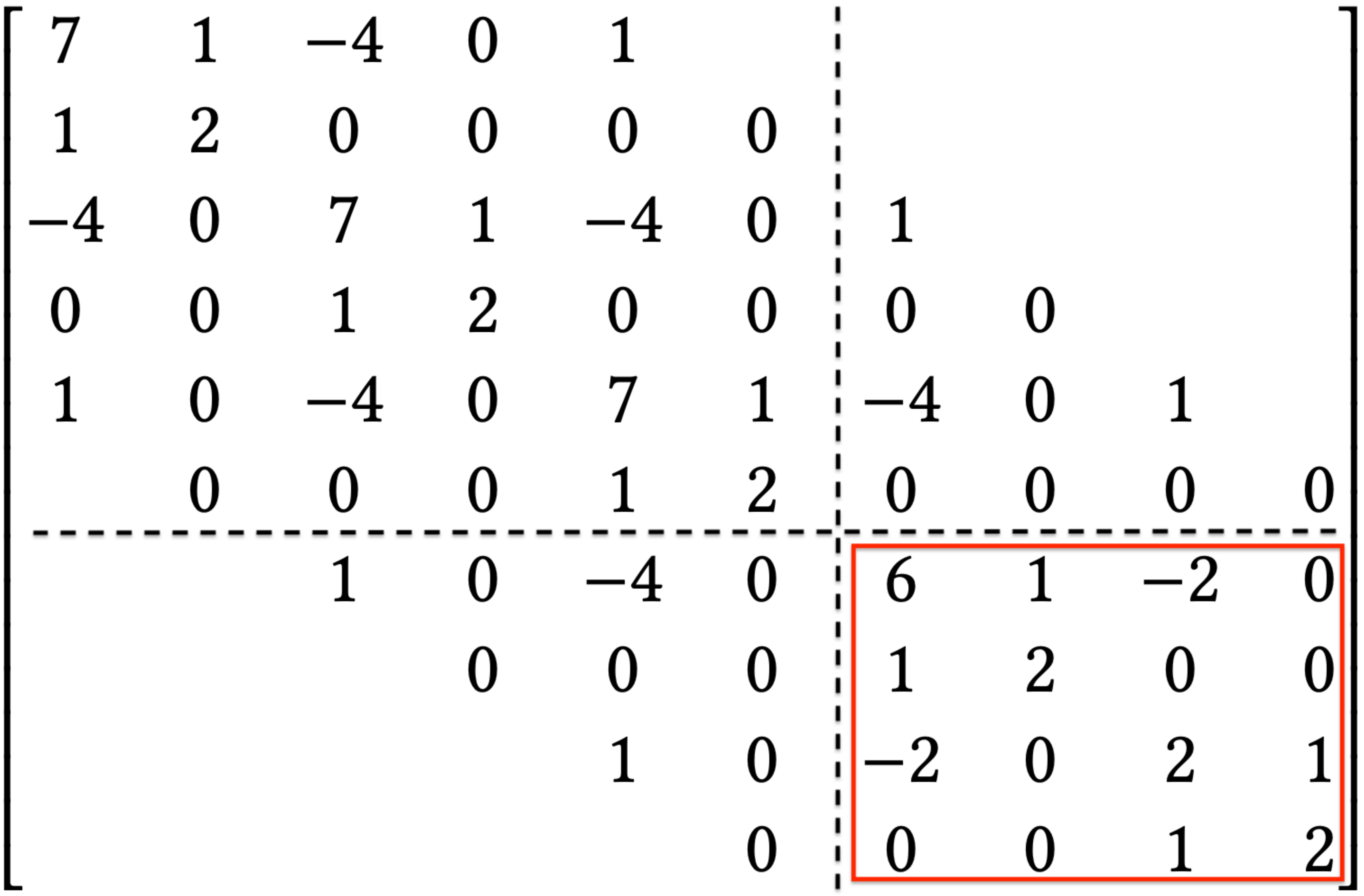}		
	\end{subfigure}
	\begin{subfigure}[t]{0.258\textwidth}
		\caption{$\vect{A^{t+1}}$ ($M=6$),~$\vect{A^*} \in \mathbb{R}^{6\times 6}$}
		\includegraphics[width=\textwidth]{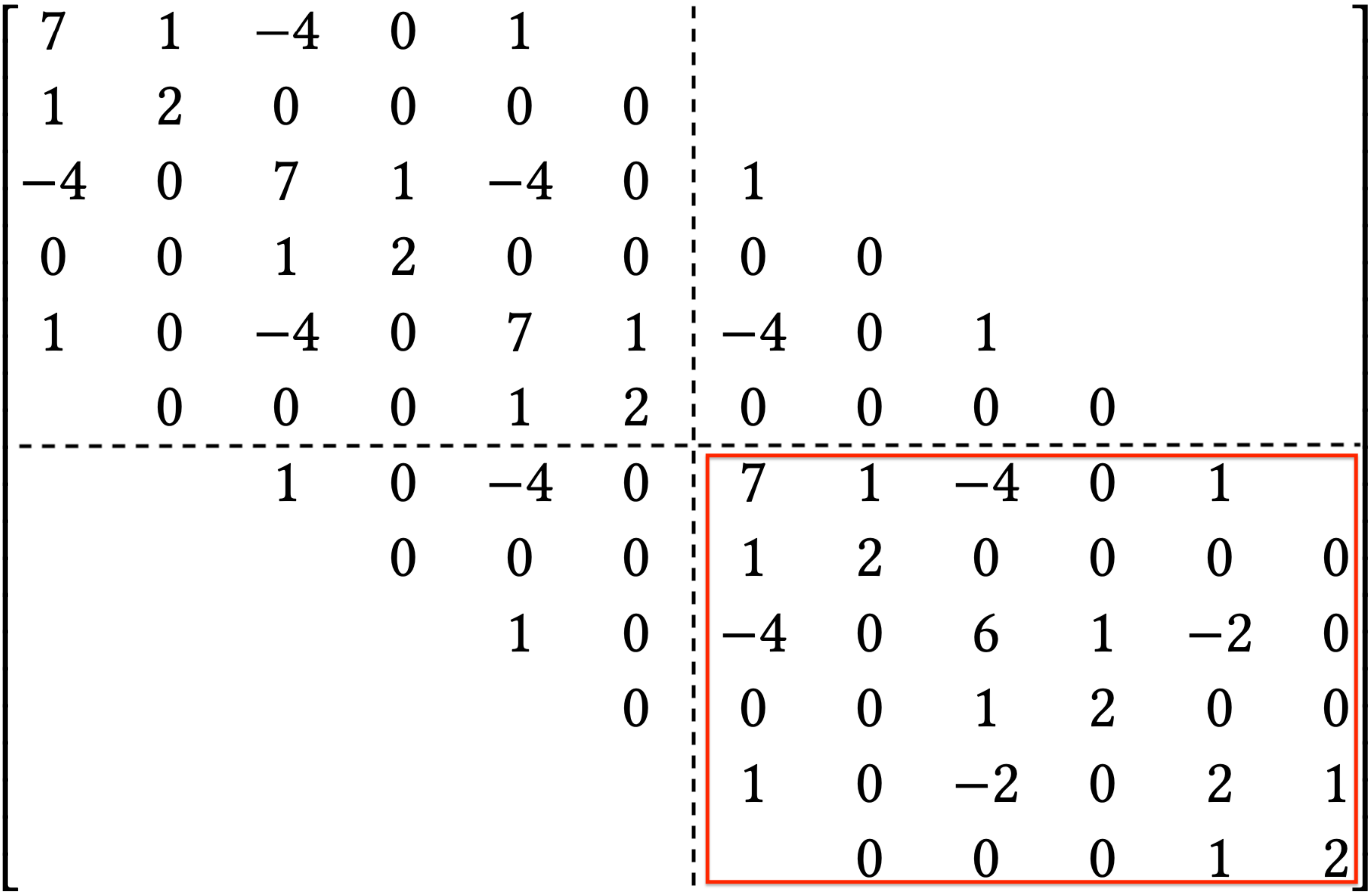}
	\end{subfigure}	
	\vspace{-3mm}
	\caption{Example of $\vect{A}$ with $\vect{p}$=$\vect{q}$=$\vect{1}$ and $\lambda_1=\lambda_2=1$.}
	\vspace{-2mm}	
	\label{fig:Aexamples}
\end{figure}

At time point $t=t_0+M$, we can construct $\vect{A^t} \in \mathbb{R}^{2M\times 2M}$ using Eq. (\ref{eq:online_ABCD}) with $\vect{p}=\vect{q}=\vect{1}$ for the first iteration of Algorithm \ref{algo:ModifiedSTL}. Similarly, for the next time point $t+1$, we can construct $\vect{A^{t+1}}$. Interestingly, some sub-matrices of $\vect{A^t}$ and $\vect{A^{t+1}}$ are identical. Figure \ref{fig:Aexamples} shows two examples of $\vect{A^t}$ and $\vect{A^{t+1}}$ with $M=5$ and $M=6$, respectively. As can be seen from the figure, we can split each matrix into $4$ parts according to dashed lines. The top-left sub-matrices of $\vect{A^t}$ and $\vect{A^{t+1}}$ are identical, and the top-right/bottom-left sub-matrices of $\vect{A^t}$ and $\vect{A^{t+1}}$ are also the same after appending rows and columns of zeros. The only differences lie in the bottom-right sub-matrix in the red box, where $\vect{A^o} \in \mathbb{R}^{4\times 4}$ of $\vect{A^t}$ is replaced by $\vect{A^*} \in \mathbb{R}^{6\times 6}$ of $\vect{A^{t+1}}$. One can easily verify that the same situation occurs for any $M\geq 5$. 

Next, we show that the common parts of $\vect{A^t}$ and $\vect{A^{t+1}}$ from adjacent timestamps can be utilized for online processing. Firstly, we can adopt Symmetric Doolittle Factorization (SDF) shown in Algorithm \ref{algo:Doolittle} to factorize $\vect{A}=\vect{L}\vect{D}\vect{L}^T$, where $\vect{D}$ is a diagonal matrix and $\vect{L}$ is a lower triangular matrix. Then, while factorizing $\vect{A^t}=\vect{L^t}\vect{D^t}\vect{L^t}^T$ using SDF, we can buffer the latest $\vect{L^t}$ and $\vect{D^t}$ before processing $\vect{A^o}$ (Figure \ref{fig:Aexamples}(a)). Further, we can continue to process $\vect{A^*}$ (Figure \ref{fig:Aexamples}(b)) for time point $t+1$ while factorizing $\vect{A^{t+1}}$ since the top-left parts of $\vect{L^{t+1}}$ and $\vect{D^{t+1}}$ are the same with $\vect{L^t}$ and $\vect{D^t}$. Lastly, thanks to the fact that $\vect{A}$ is a banded matrix of length $9$, we only need to buffer sub-matrices of $\vect{L}$ and $\vect{D}$ with a size of $\mathbb{R}^{10\times 10}$ since the others are either $0$ or not used in the computation.

\begin{algorithm}[t]
	\small
	\caption{OnlineDoolittle}
	\label{algo:online_Doolittle}
	\KwIn{$\vect{A^*} \in \mathbb{R}^{6\times 6}$, $\vect{b^*} \in \mathbb{R}^6$, ($\vect{L^o}\in \mathbb{R}^{8\times 4}$, $\vect{D^o}\in \mathbb{R}^{4\times 4}$ and $\vect{b^o}\in \mathbb{R}^4$ are needed for the first call)}
	\KwOut{$\tau_t$, $s_t$, $r_t$}
	$\vect{L^o}$, $\vect{D^o}$ and $\vect{b^o}$ are initialized for the first call and updated online\;
	$\vect{\bar{L}} = \text{zeros}(10,10)$; $\vect{\bar{L}}_{1:8,1:4}=\vect{L^o}$\; 
	$\vect{\bar{D}} = \text{zeros}(10,10)$; $\vect{\bar{D}}_{1:4,1:4}=\vect{D^o}$\;
	$\vect{\bar{b}} = [\vect{b^o}; \vect{b^*}]$\;
	\For{k = 5; k $\leq$ 10; k ++}
	{
		$\bar{L}_{kk} = 1$\;
		$\bar{D}_{kk} = A^*_{k-4,k-4} - \sum_{i=k-4}^{k}\bar{D}_{ii}\cdot \bar{L}_{ki}^2$\;		
		$\bar{b}_k = - \sum_{i=k-4}^{k}\bar{L}_{ki} \cdot \bar{b}_i$;   \Comment{Forward substitution part1}
		\For{j = k + 1; j $\leq$ 10; j ++}
		{
			$\bar{L}_{jk} = (A^*_{j-4,k-4} - \sum_{i=k-4}^{4}\bar{L}_{ji}\cdot \bar{D}_{ii}\cdot \bar{L}_{ki})~/~\bar{D}_{kk}$\;
		}
	}
	\For{j = 5; j $\leq$ 10; j ++}
	{
		\For{i = j + 1; i < j + 5; i ++}
		{
				$\bar{b}_i$ -= $\bar{L}_{ij} \cdot \bar{b}_j$;				\Comment{Forward substitution part2}	
		}
	}
	\For{j = 10; j $\leq$ 5; j - -}
	{
		\For{i = j + 1; i $\leq$ 10; i ++}
		{
				$\bar{b}_i$ -= $\bar{D}_{ii} \cdot \bar{L}_{ji} \cdot \bar{b}_j$;				\Comment{Backward substitution}	
		}
	}
	$\vect{L^o} = \vect{\bar{L}}_{3:10,3:6}$; $\vect{D^o} = \vect{\bar{D}}_{3:6,3:6}$; $\vect{b}^o=\vect{\bar{b}}_{3:6}$\;
	$\tau_t = \bar{b}_9$; $s_t = \bar{b}_{10}$; $r_t=y_t-\tau_t-s_t$\;
\end{algorithm}

%

The complete OnlineDoolittle algorithm for partially solving $\vect{A}\vect{x}=\vect{b}$ online is presented in Algorithm \ref{algo:online_Doolittle}, including the forward and backward substitutions of Gaussian elimination that can be modified with the same principle. Note that OnlineDoolittle is almost the same as the original SDF (Algorithm \ref{algo:Doolittle}), where we keep updating the most recent $\vect{L^o}$, $\vect{D^o}$ and $\vect{b}^o$ with the factorization and the forward/backward substitutions conducted online. Finally, it outputs the decomposition result of $\tau_t$ and $s_t$ for each timestamp $t$. The number of calculations needed in Algorithm \ref{algo:online_Doolittle} is always a constant, and thus its runtime complexity is $O(1)$.


Next, we show that OnlineDoolittle algorithm can also be used for partially solving $\vect{A}\vect{x}=\vect{b}$ for the $i$-th iteration of Algorithm \ref{algo:ModifiedSTL}. To construct $\vect{A}$ and $\vect{b}$ using Eq. (\ref{eq:online_ABCD}) for iteration $i$, we need $\vect{p}^{i-1}$ and $\vect{q}^{i-1}$ updated from the previous iteration. Note that we only append $p_t$ and $q_t$ at each time point $t$ calculated in iteration $i-1$ to $\vect{p}^{i-1}$ and $\vect{q}^{i-1}$. Therefore, the values of $\vect{p}_{1:t-1}$ and $\vect{q}_{1:t-1}$ for all the previous timestamps are fixed while processing time point $t$ at iteration $i$. Thus, the $\vect{A^t}$ and $\vect{A^{t+1}}$ from adjacent timestamps for iteration $i$ also share the same property, as shown in Figure \ref{fig:Aexamples}.
Then, we can build a linear system with growing $\vect{A}$ and $\vect{b}$ for each iteration, whose construction depends on the results of the previous iteration. Eventually, they can be partially solved online using the OnlineDoolittle algorithm to obtain the final decomposition results.



\subsubsection{\textbf{OneShotSTL algorithm details}} 
Combining all the ingredients introduced before, we have the complete OneShotSTL algorithm, shown in Algorithm \ref{algo:OneShotSTL}. It solves the same problem as Algorithm \ref{algo:ModifiedSTL} in an online fashion.
During the initialization phase, we conduct STL or batch JointSTL on the existing data to obtain one period of seasonal component $\vect{v}$. During the online phase, at time point $t$ OneShotSTL processes a new observation $y_t$ and outputs the decomposition results $\tau_t$ and $s_t$. Specifically, for each iteration $i$, it sequentially updates $p_t$ and $q_t$ using $\tau_t$ and $s_t$ from previous iterations and constructs $\vect{A^*}$ and $\vect{b^*}$ as in Figure \ref{fig:Aexamples} using Eq. (\ref{eq:online_ABCD}). Then, each iteration incrementally applies the symmetric Doolittle factorization using $\vect{A^*}$ and $\vect{b^*}$ with OnlineDolittle in Algorithm \ref{algo:online_Doolittle}, denoted as OnlineDolittle$_i$ for iteration $i$. Finally, we output the decomposition result $\tau_t$ and $s_t$ from the last iteration and update the corresponding value in $\vect{v}$ with the decomposed seasonal $s_t$. 

\begin{algorithm}[t]
	\small
	\caption{OneShotSTL}
	\label{algo:OneShotSTL}
	\Comment{Initialization phase}
	\KwIn{$y_1,...,y_{t_0}$, $T$, $\lambda$, maximum iterations $I$}
	$M = 0$; Obtain $\vect{\tau}, \vect{s}, \vect{r}$ by STL or JointSTL; Set $\vect{v}=[s_{t_0-T},...,s_{t_0}]$\;
	$\vect{p}^0=\vect{q}^0=[1]$; $\vect{p}^i=\vect{q}^i=[]$, for $i \in [1,I]$\;	
	
	\vspace{2mm}
	\Comment{Online phase}
	\KwIn{$y_t$, $t$, $n$, $\lambda$, $I$}
	\KwOut{$\tau_t$, $s_t$, $r_t$}
	\For{i = 0; i < I; i++}
	{
		Construct $\vect{A^*}$ and $\vect{b^*}$ as in Figure \ref{fig:Aexamples} using Eq. (\ref{eq:online_ABCD})\;
			$\tau_t$, $s_t$, $r_t$ = OnlineDoolittle$_i$($\vect{A^*}$, $\vect{b^*}$)\;
			Update $p_t$ and $q_t$ using Eq. (\ref{eq:pt}) and (\ref{eq:qt})\;
			$\vect{p}^{i+1}=[\vect{p}^{i+1}, p_t];\vect{q}^{i+1}=[\vect{q}^{i+1}, q_t]$\;			
	}	
	$\vect{p}^0=[\vect{p}^0, 1];\vect{q}^0=[\vect{q}^0, 1]$\;			
	$v_{t\%T}=s_t$\;
\end{algorithm}

The time complexity of OnlineDolittle (Algorithm \ref{algo:online_Doolittle}) is $O(1)$. Thus, the worst-case online update complexity of OneShotSTL is $O(I)$, where $I$ is the maximum number of iterations. 
Note that $I$ is a fixed constant, independent of the problem size, implying that the online update complexity of OneShotSTL is $O(1)$. In practice, it is the main factor that affects the speed of OneShotSTL. But, a small value, e.g., $I=8$, is sufficient to produce satisfactory results; see experiments in Section~\ref{sec:experiment}).

\subsection{\textbf{Handle Seasonality Shift}}
\label{sec:shift}
In this section, we handle the case of seasonality shift when the underlying $T_c$ can be any integer in $[T-H, T+H], H\geq 1$.
\begin{figure}[h]
	\centering
	\vspace{-2mm}		
	\includegraphics[width=0.4\textwidth]{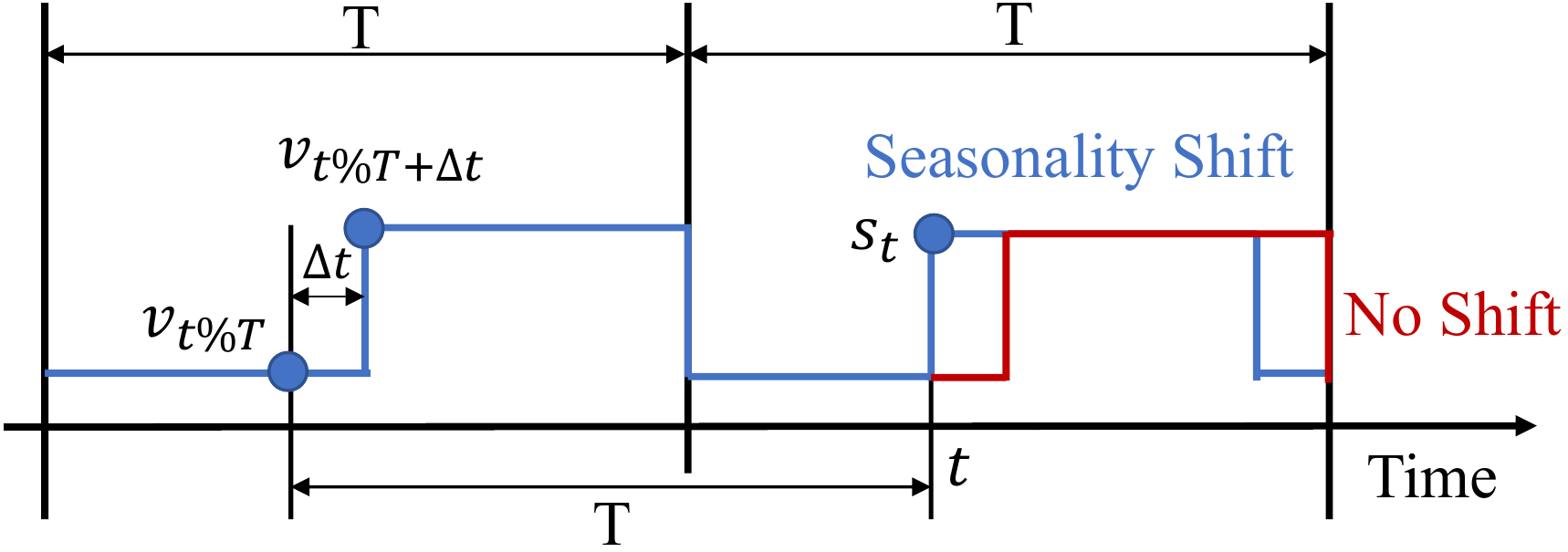}
	\vspace{-2mm}	
	\caption{Seasonality Shift of $\Delta t$ time points.}
	\vspace{-2mm}	
	\label{fig:seasonality_shift}
\end{figure}

Figure \ref{fig:seasonality_shift} depicts an example of seasonality shift of $\Delta t$ time points, which happens commonly in the metric monitoring applications of AIOps. The blue line represents the ground truth of $2$ periods of seasonal component of a time series data. From time point $t$ there is a shift of $\Delta t$ time points. The red dashed line represents the imaginary data if there is no shift. For simplicity, until time point $t$ we assume $\vect{v}$ equals the first seasonal period in Figure \ref{fig:seasonality_shift}. Clearly, OneShotSTL in Algorithm \ref{algo:OneShotSTL}  will not decompose it well after time point $t$, since $v_{t\% T}$ is significantly different from the ground truth $s_t$. If we know $\Delta t$, we can replace $\left(s_t-v_{t\% T}\right)$ in Problem (\ref{eq:joint_opt_online}) with $\left(s_t-v_{t\% T+\Delta t}\right)$ to tackle the seasonality shift issue. However, we do not know the correct $\Delta t$ for each time point $t$ in practice. 

\begin{algorithm}[t]
	\small
	\caption{NSigma}
	\label{algo:spike}
	\KwIn{$r_t$, $n$}
	\KwOut{isAnomaly, score}
	isAnomaly = False\;
	\If{not initialized}
	{
		sum = sumSquared = count = 0\;
	}
	\If{count $> 0$} 
	{
		mean = sum / count; std = sqrt(sumSquared / count - mean$^2$)\;
		score = abs($r_t$ - mean) / std\;
		\If{score > $n$}
		{
			isAnomaly = True\;
		}
	}
	count += $1$; sum += $r_t$; sumSquared += $r_t^2$\;
\end{algorithm}

To solve this problem, we first use the hyper-parameter $H$ to form neighborhoods $\mathcal{E}$ with all possible $\Delta t$ within $H$, e.g.,  $\mathcal{E}=\{0\}$ for $H=0$ and $\mathcal{E}=\{-1,0,1\}$ for $H=1$.
Further, we observe that when a seasonality shift happens, there will be a spike anomaly point in the decomposed residual $r_t$. Therefore, we monitor the residual $r_t$ and use the streaming NSigma algorithm shown in Algorithm \ref{algo:spike} to detect whether there is an anomaly in $r_t$ for each time point $t$. In Algorithm \ref{algo:spike}, $n$ is an input parameter indicating the level of abnormal, e.g., $n=5$. Once we found an anomaly, we repeat lines $3-8$  of Algorithm \ref{algo:OneShotSTL} for each $\Delta t \in \mathcal{E}$ that is used to construct $\vect{A^*}$ and $\vect{b^*}$ (line $4$). Then we pick $\Delta t$ with the smallest absolute value of $r_t$ and output the corresponding $\tau_t$ and $s_t$. This method gives the most accurate decomposition results in our simulations.

With the above method to handle seasonality shift, the computation complexity of OneShotSTL is $O(I\cdot H)$, where $H$
is the prefixed parameter that indicates the maximum variations allowed for the cycle length shift, e.g., $H=20$. Since both $I$ and $H$ are fixed constant, independent of each period length $T_c$, implying that the online update complexity of OneShotSTL is still $O(1)$. Moreover, it also does not affect the speed of OneShotSTL much in practice since anomalies (e.g., with $n > 5$) in the residual signal are rare (e.g., usually less than $1\%$).

\section{Extension For Downstream tasks} 
\label{sec:extension}
OneShotSTL and other online STD methods can be easily extended for downstream tasks of online TSAD and TSF as  follows:
\begin{enumerate}
	\item \textbf{Univariate TSAD}: For an unbounded time series, 
	at each time point $t$, an online TSAD method observes all the historical data $\{y_1,y_2,...,y_{t}\}$ and outputs an anomaly score for $y_t$. OnlineSTL and OneShotSTL can 
	be extended to online TSAD methods by combining with streaming NSigma shown in Algorithm \ref{algo:spike}. Besides the decomposition results, they can output anomaly scores using NSigma on the decomposed residual signals. 
	\item \textbf{Univariate TSF}: Similarly, an online TSF method observes $\{y_1,...,y_t\}$ at time point $t$ and makes an $i$-step prediction for the future value of data $\hat{y}_{t+i}$ at time point $t+i$. Seasonal signals are periodical, thus we can use them for long-sequence forecasting. We adopt a simple method where we buffer the most recent decomposed trend ($\tau_{t-1}$) and one period of seasonal signals ($\vect{v}$) during online decomposition. Then, to make an $i$-step prediction at timestamp $t$, we simply use $\hat{y}_{t+i} = \tau_{t-1}+v_{t \% T + i}$ where $T$ is the period length. The same method applies to OnlineSTL.
\end{enumerate}

\begin{figure*}[t]
	\vspace{-2mm}
	\begin{subfigure}[b]{0.235\textwidth}
		\caption{Syn1 (Trend Abrupt Change)}		
		\vspace{-1mm}		
		\includegraphics[width=\textwidth]{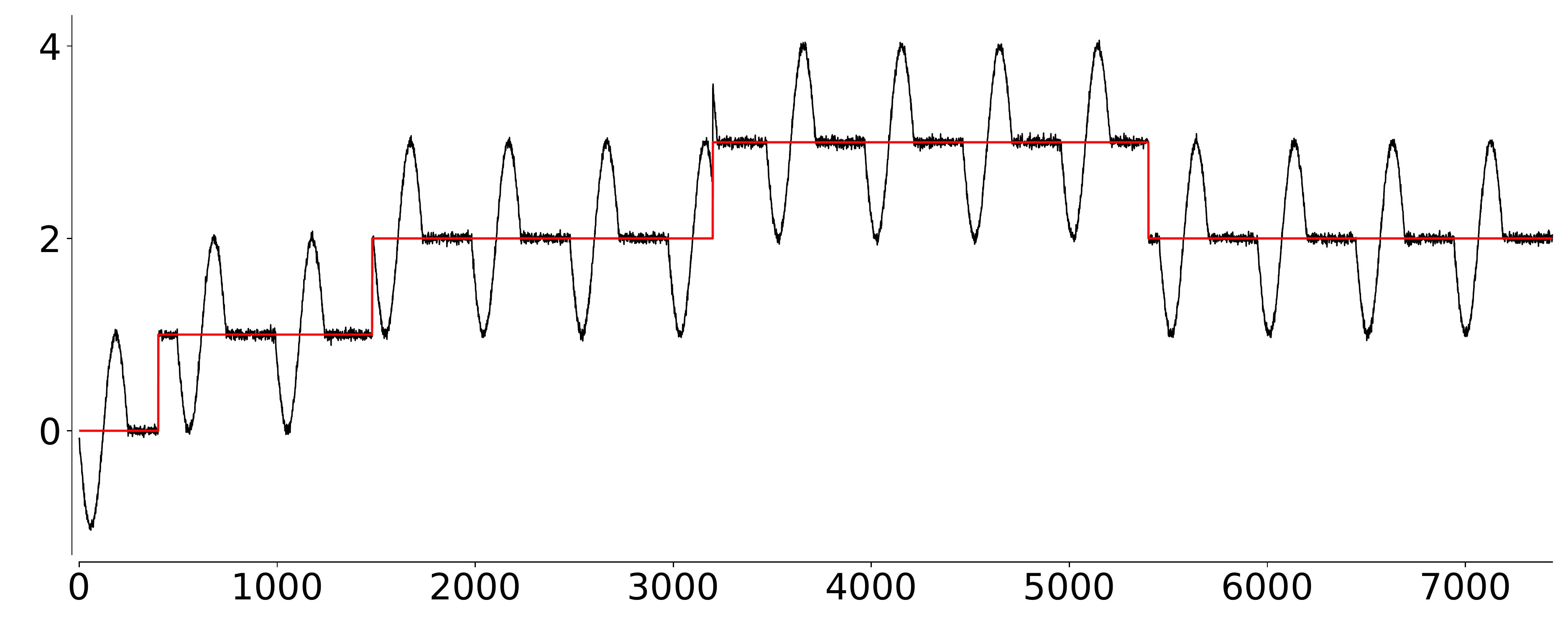}		
		\vspace{-5mm}
	\end{subfigure}	
	\begin{subfigure}[b]{0.235\textwidth}
		\caption{Syn2 (Seasonality Shift)}				
		\vspace{-1mm}				
		\includegraphics[width=\textwidth]{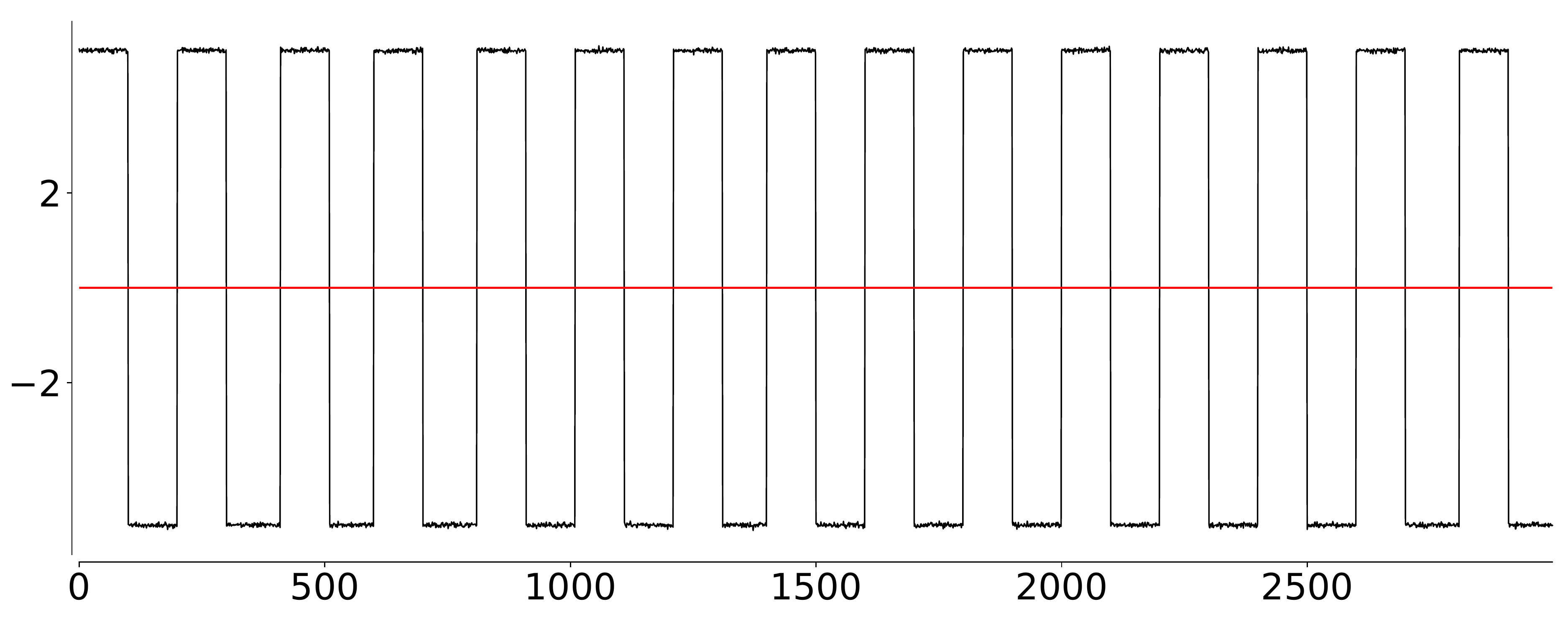}
		\vspace{-5mm}
	\end{subfigure}	
	\begin{subfigure}[b]{0.235\textwidth}
		\caption{Real1 dataset}		
		\vspace{-1mm}				
		\includegraphics[width=\textwidth]{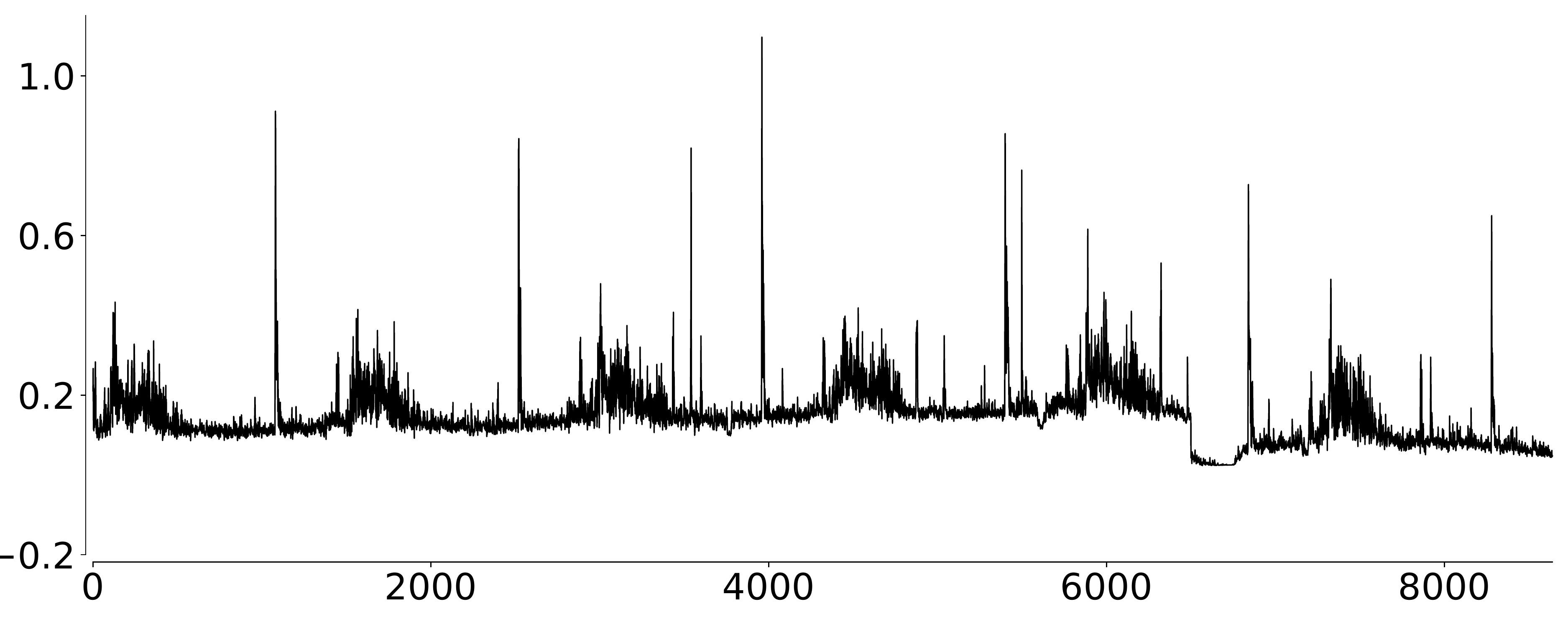}	
		\vspace{-5mm}
	\end{subfigure}	
	\begin{subfigure}[b]{0.235\textwidth}
		\caption{Real2 dataset}				
		\vspace{-1mm}				
		\includegraphics[width=\textwidth]{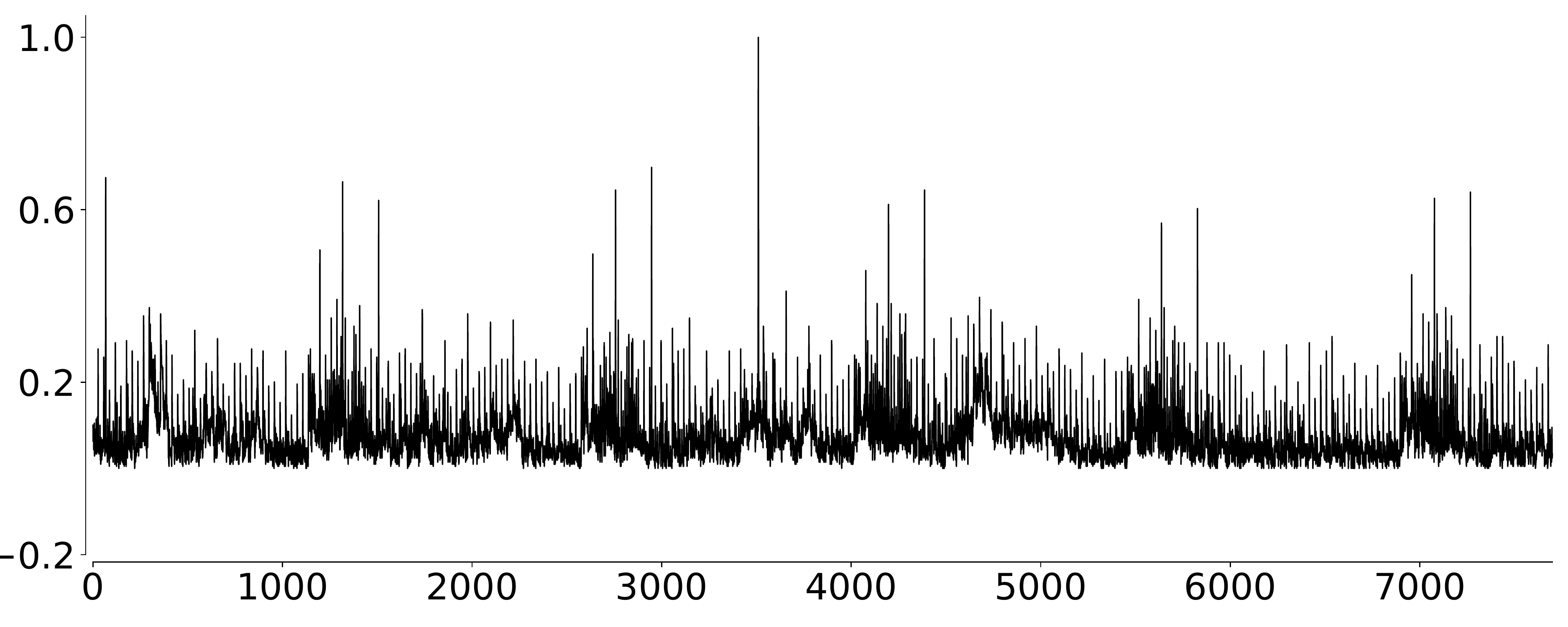}
		\vspace{-5mm}
	\end{subfigure}	
	\vspace{-2mm}
	\caption{Synthetic and Real-world datasets for the evaluation of STD quality.}
	\vspace{-4mm}	
	\label{fig:synthetic}
\end{figure*}

\section{Experiments}
\label{sec:experiment}
We conduct comprehensive experiments to evaluate OneShotSTL
in comparison with other methods. After introducing the experimental setting, we evaluate the decomposition accuracy and the scalability, followed by comparison on downstream TSAD and TSF tasks.

\subsection{Experimental Settings}
\subsubsection{\textbf{Datasets}} Three sets of datasets are used for evaluation.

\textbf{1. Synthetic and Real-world STD datasets.} We generate two synthetic datasets (\textbf{Syn1} and \textbf{Syn2}) and use two real-world datasets 
(\textbf{Real1} and \textbf{Real2}) as shown in Figure \ref{fig:synthetic}. 
These two real-world datasets are on the request rates of some internal APIs on Alibaba Cloud Database systems.
The red lines in Figure \ref{fig:synthetic} (a) and (b) show the ground truth of the trend signals of the synthetic datasets. \textbf{Syn1} and \textbf{Syn2} are designed for cases with abrupt trend changes and seasonality shifts, respectively. In \textbf{Syn2} there are $4$ periods shifted by $10$ data points which are not visually distinguishable. Due to the limited space, we present the details of generating the synthetic datasets in the supplementary material~\cite{supp}. For the two real-world datasets, \textbf{Real1} exhibits an abrupt change of the trend and \textbf{Real2} contains observable noises with weak seasonality.


\textbf{2. Univariate TSAD datasets.} We evaluate the univariate TSAD task using the TSB-UAD benchmark~\cite{DBLP:journals/pvldb/PaparrizosKBTPF22}. There are over $2000$ real-world time series in the TSB-UAD benchmark with anomalies labeled by domain experts. Among them, the most famous one is the KDD CUP 2021 (KDD21) TSAD competition  dataset~\cite{UCRArchive2018} consists of $250$ univariate time series. 
For details of these datasets please refer to~\cite{DBLP:journals/pvldb/PaparrizosKBTPF22}. For KDD21, IOPS, NASA-MSL and NASA-SMAP datasets, we use the train part of the data for initialization. For the rest of the datasets, we use the first $3000$ data points for initialization.


\textbf{3. Univariate TSF datasets.} We evaluate the TSF task on six real-world datasets (ETT, Electricity, Exchange, Traffic, Weather, and Illness) from Informer~\cite{zhou2021informer} which have been extensively used to benchmark long sequence TSF by recent works, e.g., FEDformer~\cite{DBLP:conf/icml/ZhouMWW0022} and FiLM~\cite{DBLP:journals/corr/abs-2205-08897}. We follow the same experimental settings as these methods with forecasting lengths of $\{24, 36, 48, 60\}$ for Illness and $\{96, 192, 336, 720\}$ for the rest. The train, validation, and test splitting details can be found in ~\cite{zhou2021informer} and are publicly available~\cite{tsfdata}.


\begin{figure*}[t]
	\begin{subfigure}[b]{0.235\textwidth}
		\caption{Batch RobustSTL}				
		\vspace{-2mm}				
		\includegraphics[width=\textwidth]{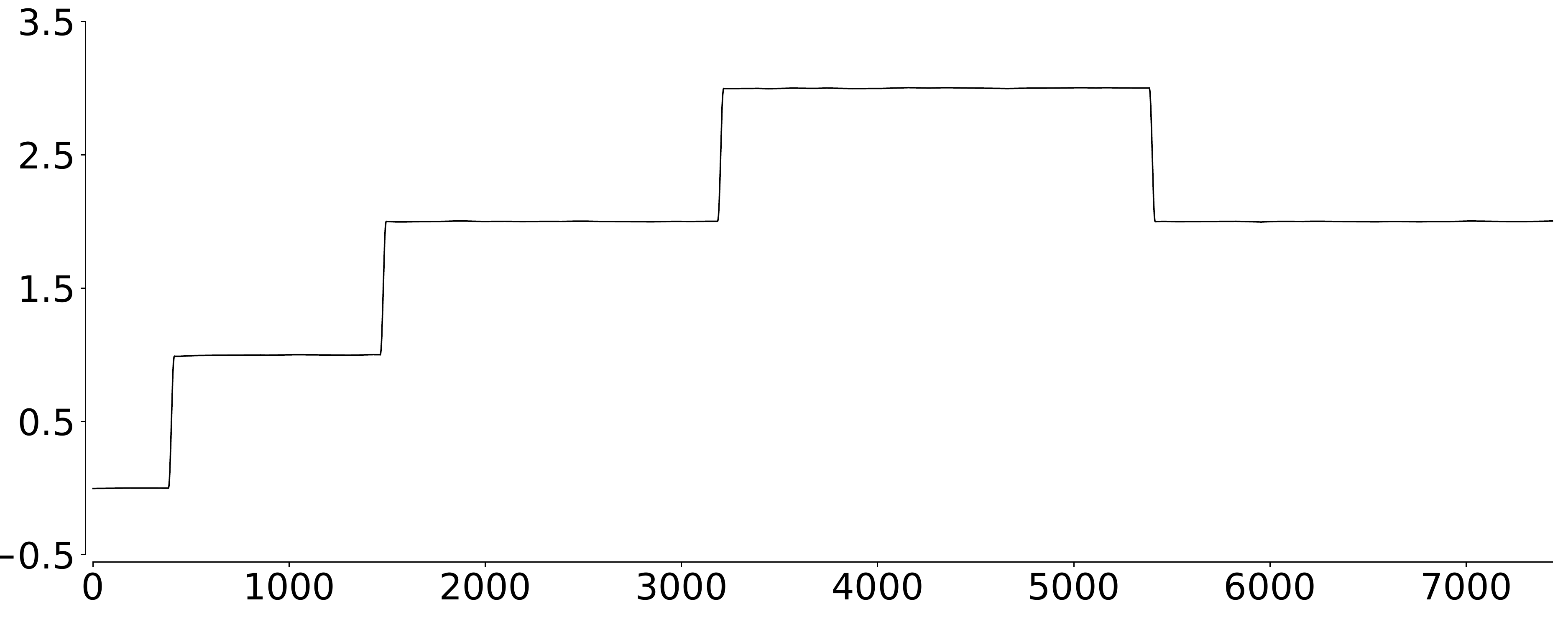}	
		\vspace{-5mm}
	\end{subfigure}	
	\begin{subfigure}[b]{0.235\textwidth}
		\caption{OnlineSTL}				
		\vspace{-2mm}				
		\includegraphics[width=\textwidth]{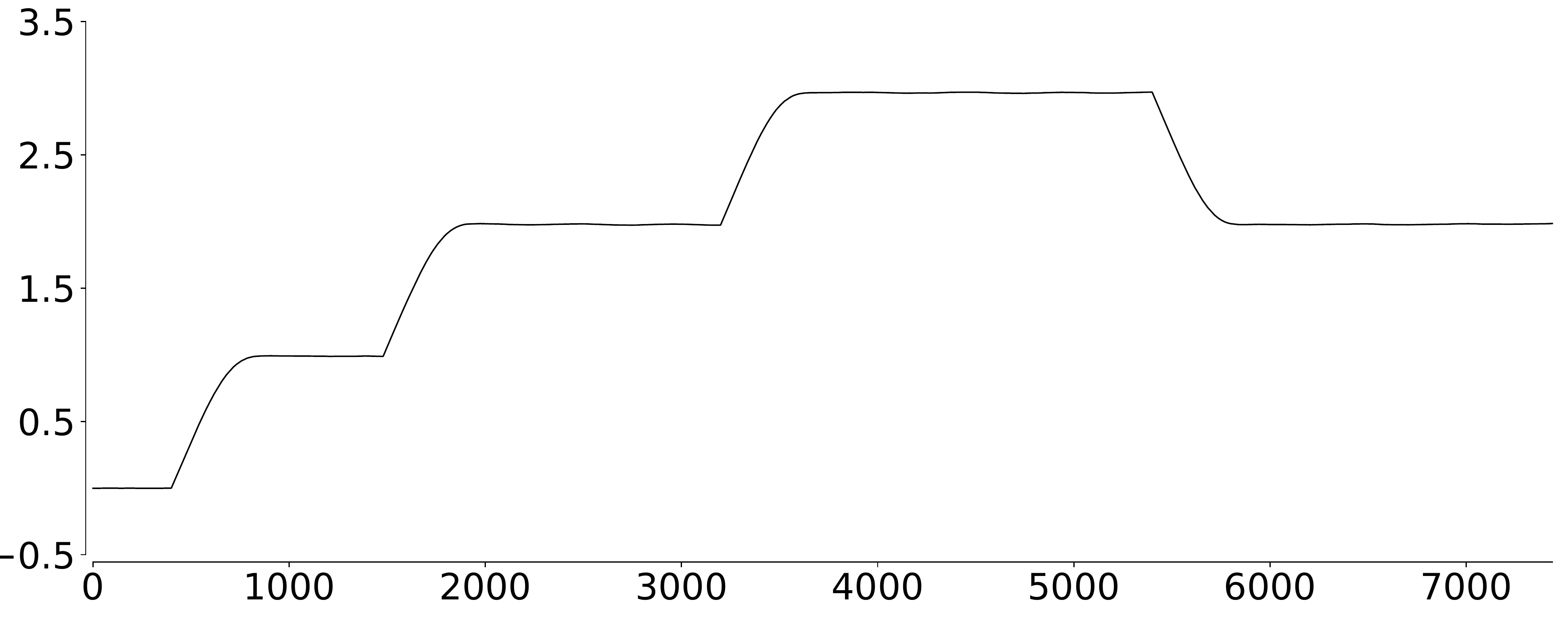}		
		\vspace{-5mm}
	\end{subfigure}	
	\begin{subfigure}[b]{0.235\textwidth}
		\caption{OnlineRobustSTL}				
		\vspace{-2mm}				
		\includegraphics[width=\textwidth]{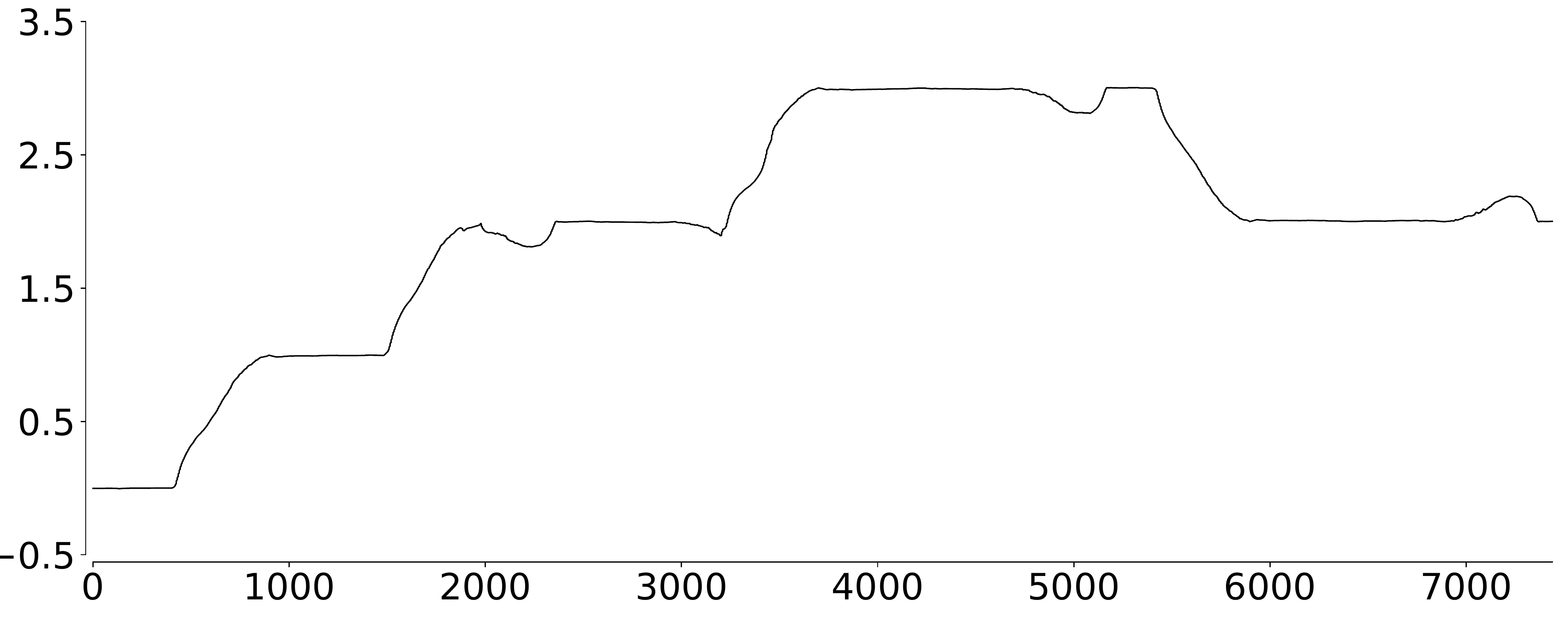}		
		\vspace{-5mm}
	\end{subfigure}	
	\begin{subfigure}[b]{0.235\textwidth}
		\caption{OneShotSTL}				
		\vspace{-2mm}				
		\includegraphics[width=\textwidth]{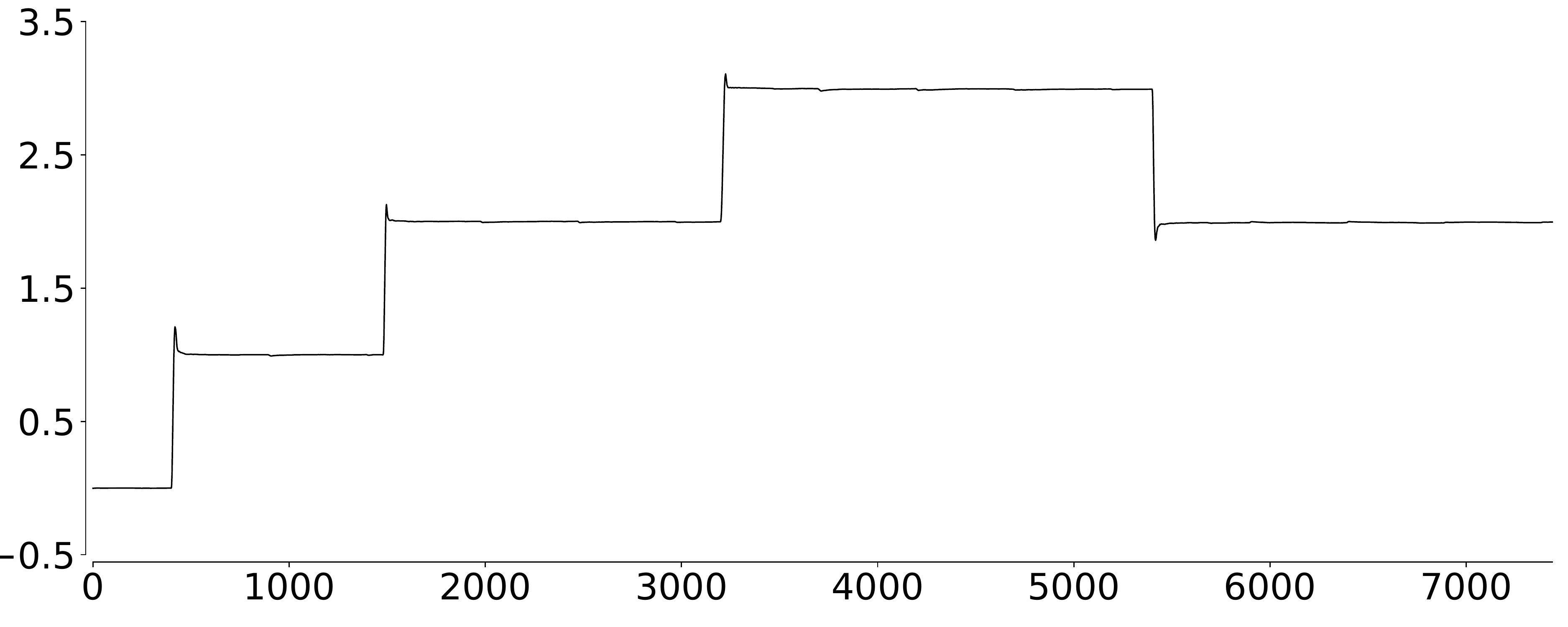}		
		\vspace{-5mm}
	\end{subfigure}	
	
	\begin{subfigure}[b]{0.235\textwidth}
		\includegraphics[width=\textwidth]{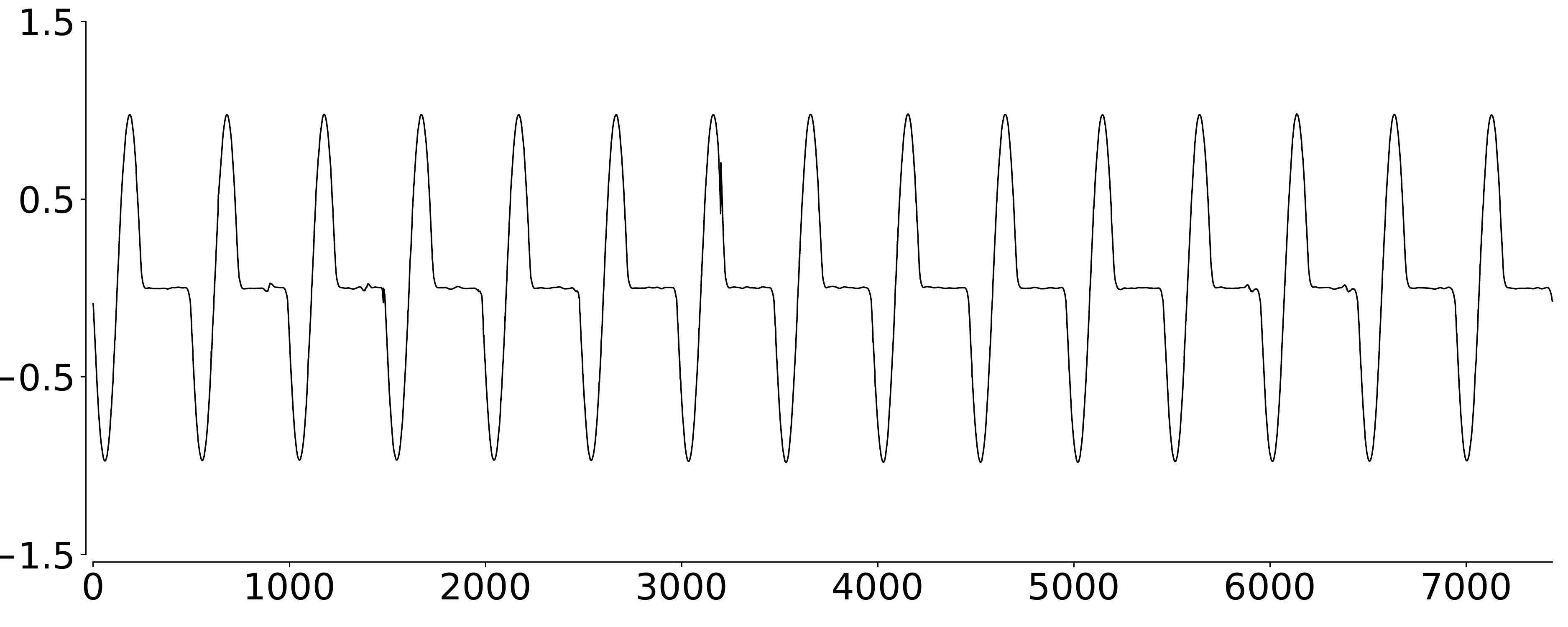}
		\vspace{-5mm}
	\end{subfigure}
	\begin{subfigure}[b]{0.235\textwidth}
		\includegraphics[width=\textwidth]{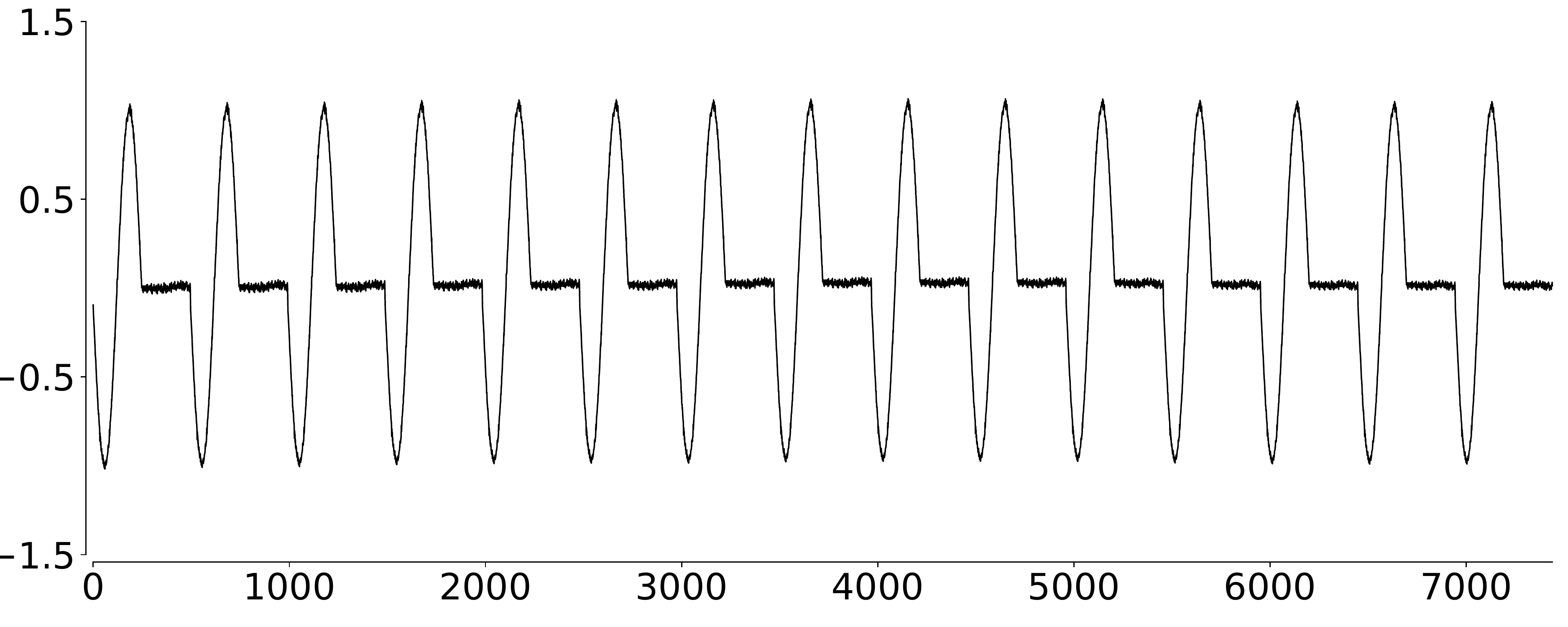}
		\vspace{-5mm}
	\end{subfigure}
	\begin{subfigure}[b]{0.235\textwidth}
		\includegraphics[width=\textwidth]{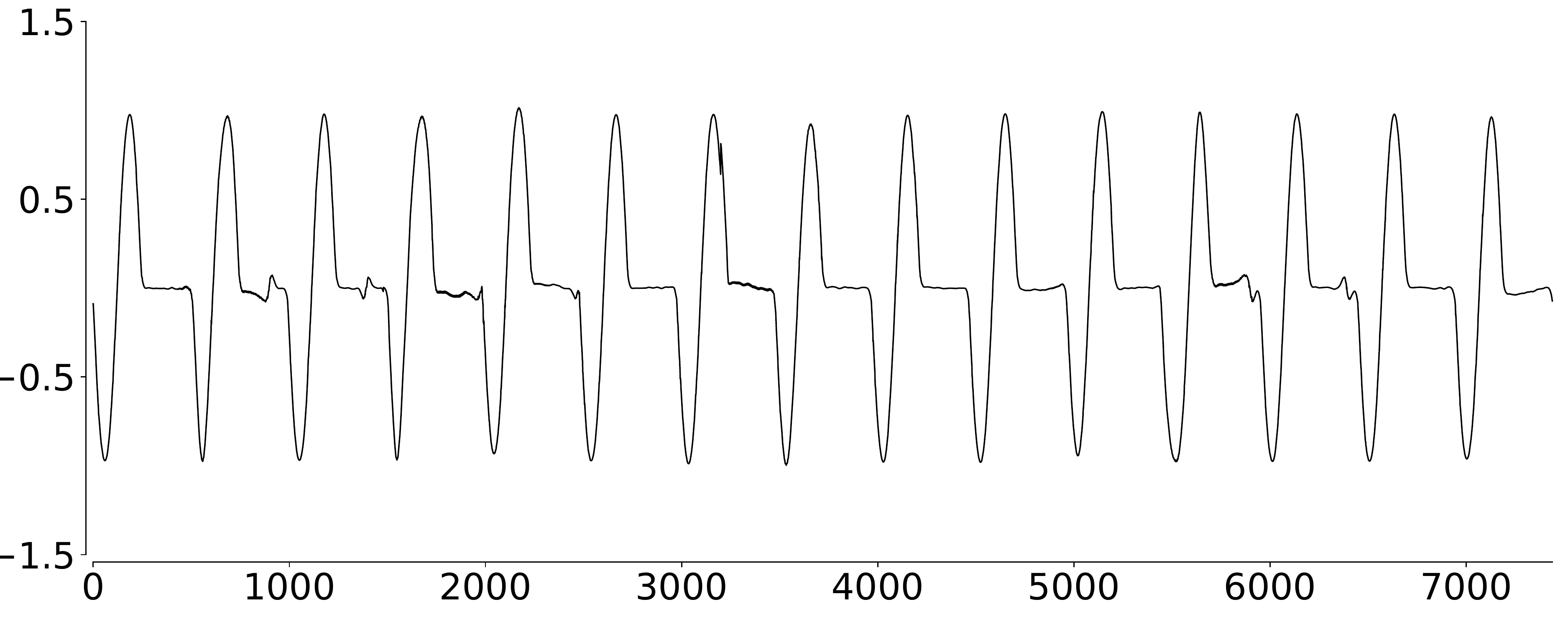}
		\vspace{-5mm}
	\end{subfigure}
	\begin{subfigure}[b]{0.235\textwidth}
		\includegraphics[width=\textwidth]{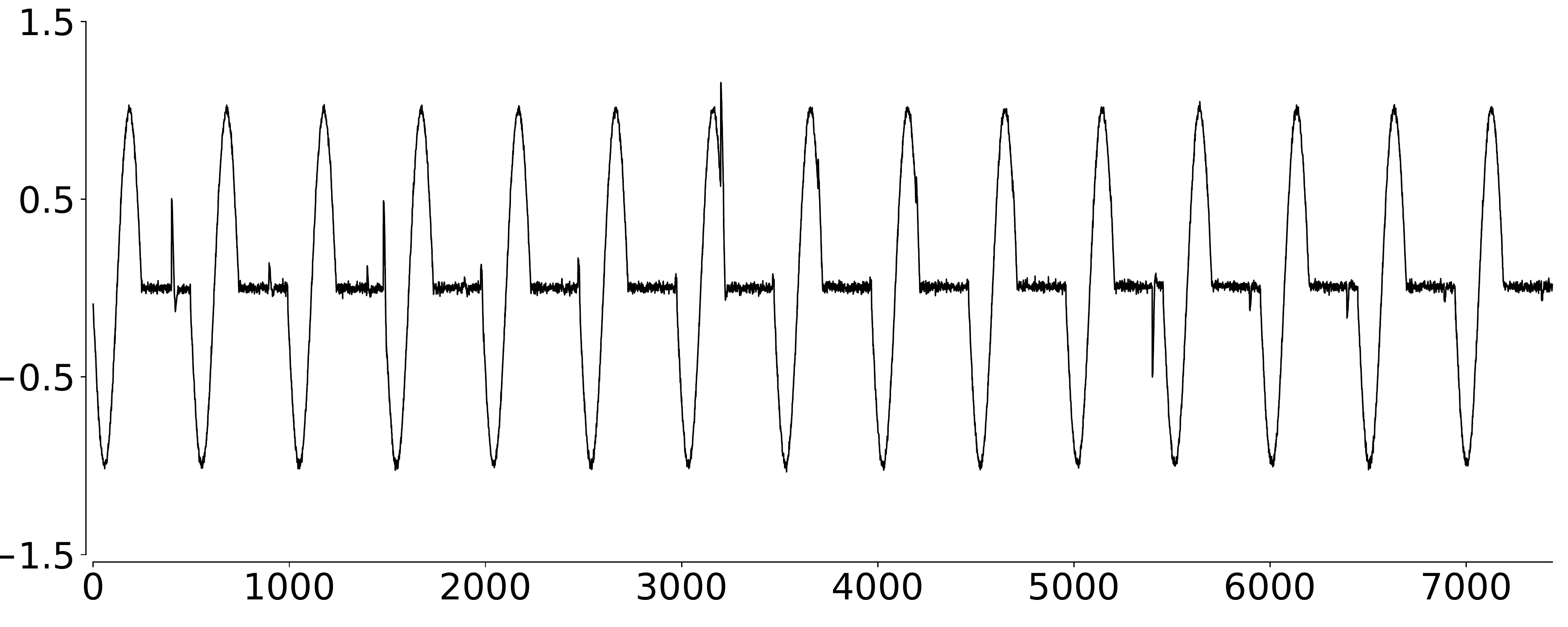}
		\vspace{-5mm}
	\end{subfigure}
	
	\begin{subfigure}[b]{0.235\textwidth}
		\includegraphics[width=\textwidth]{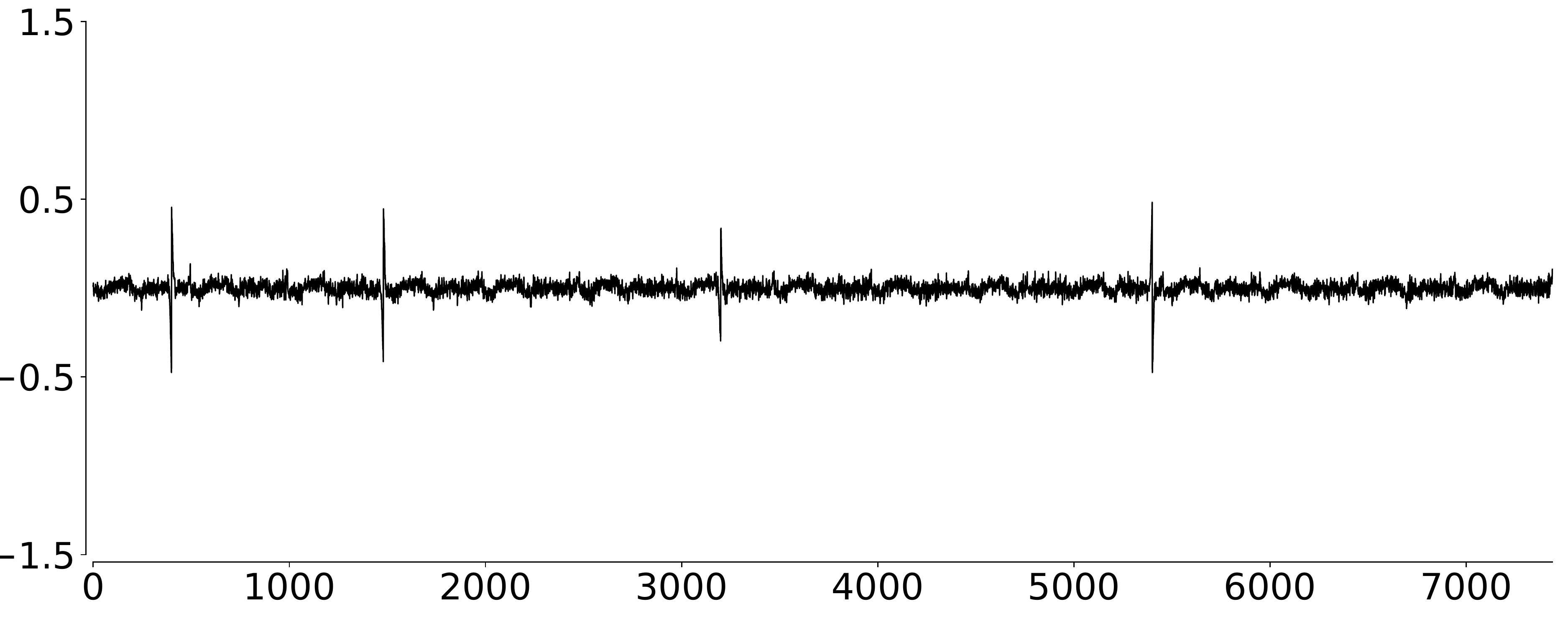}
		\vspace{-5mm}
	\end{subfigure}
	\begin{subfigure}[b]{0.235\textwidth}
		\includegraphics[width=\textwidth]{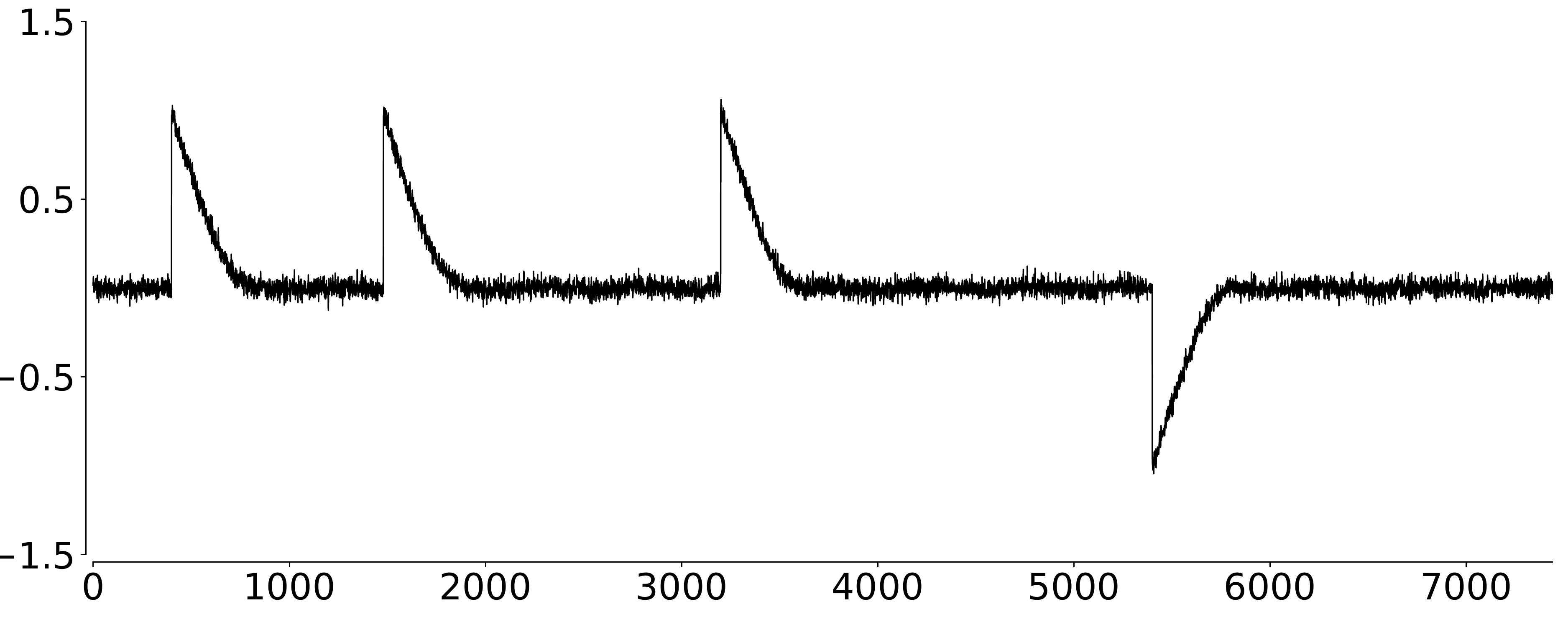}
		\vspace{-5mm}
	\end{subfigure}
	\begin{subfigure}[b]{0.235\textwidth}
		\includegraphics[width=\textwidth]{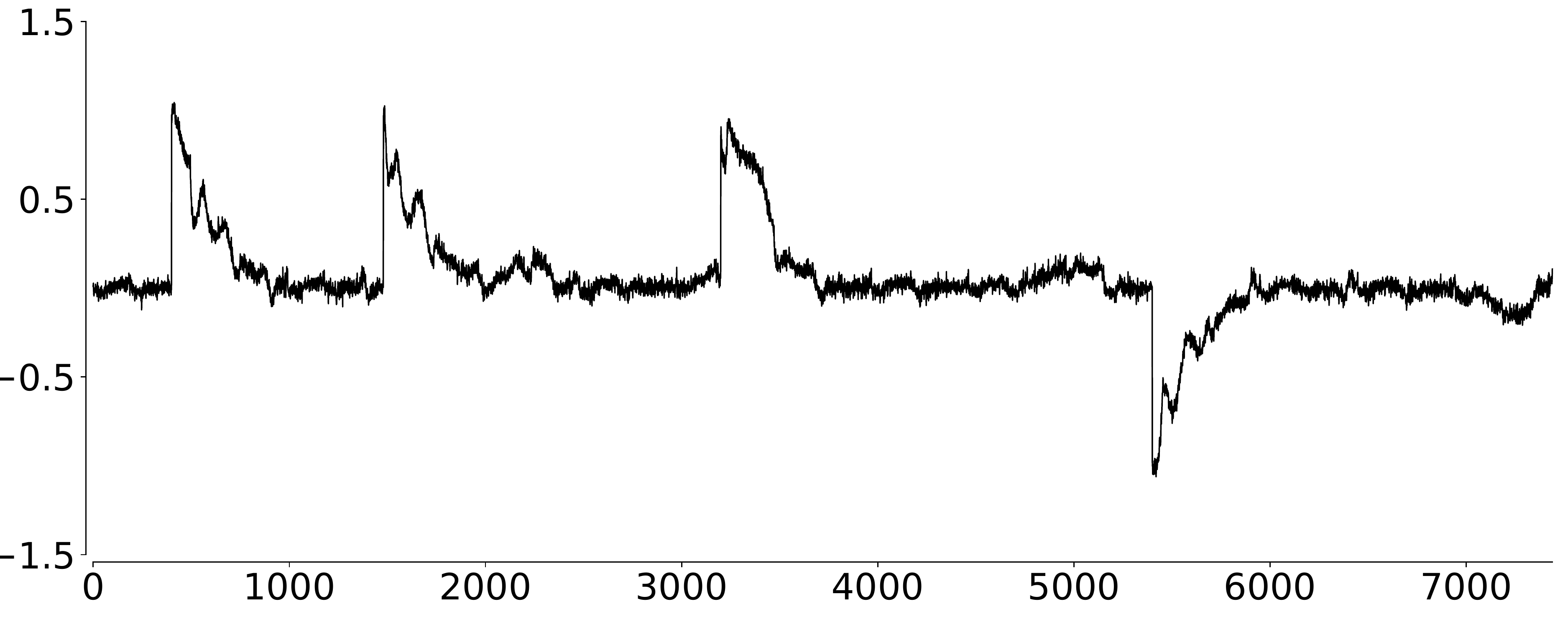}	
		\vspace{-5mm}
	\end{subfigure}
	\begin{subfigure}[b]{0.235\textwidth}
		\includegraphics[width=\textwidth]{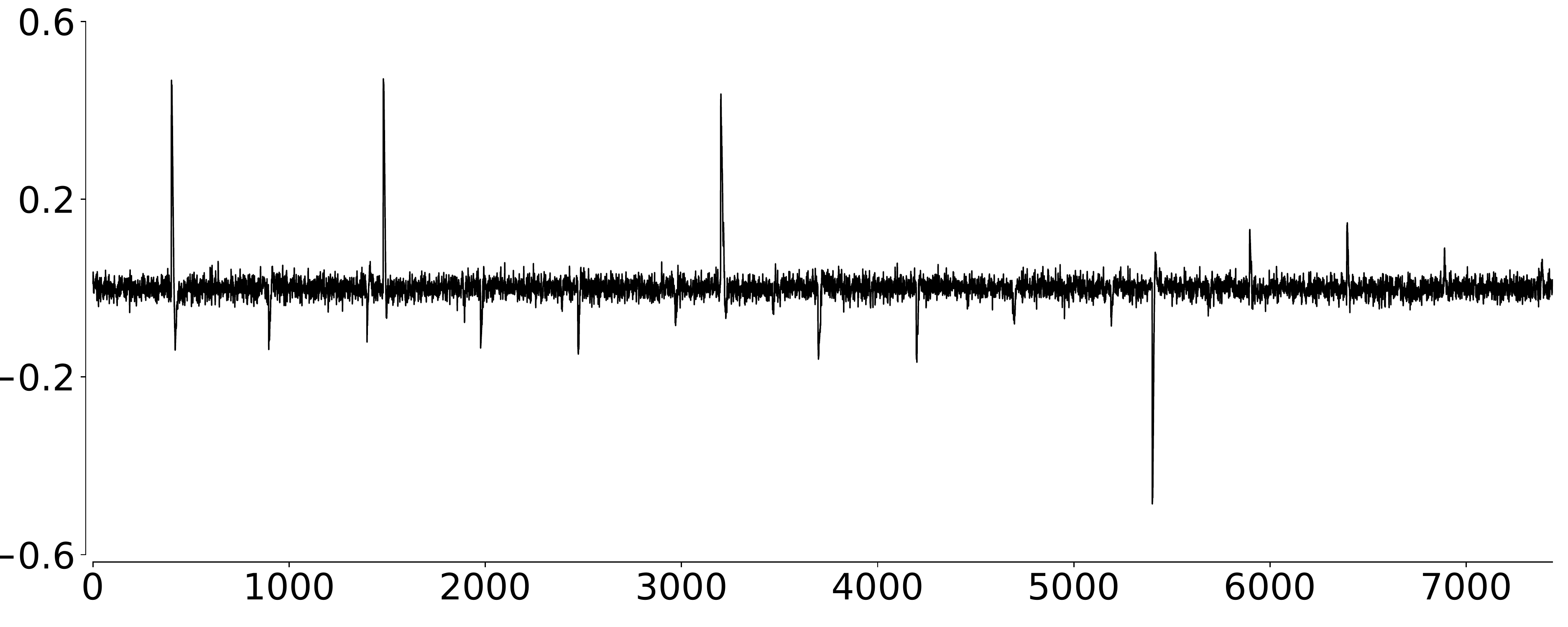}	
		\vspace{-5mm}
	\end{subfigure}
	\begin{subfigure}[b]{0.235\textwidth}
		\caption{Batch RobustSTL}				
		\vspace{-2mm}				
		\includegraphics[width=\textwidth]{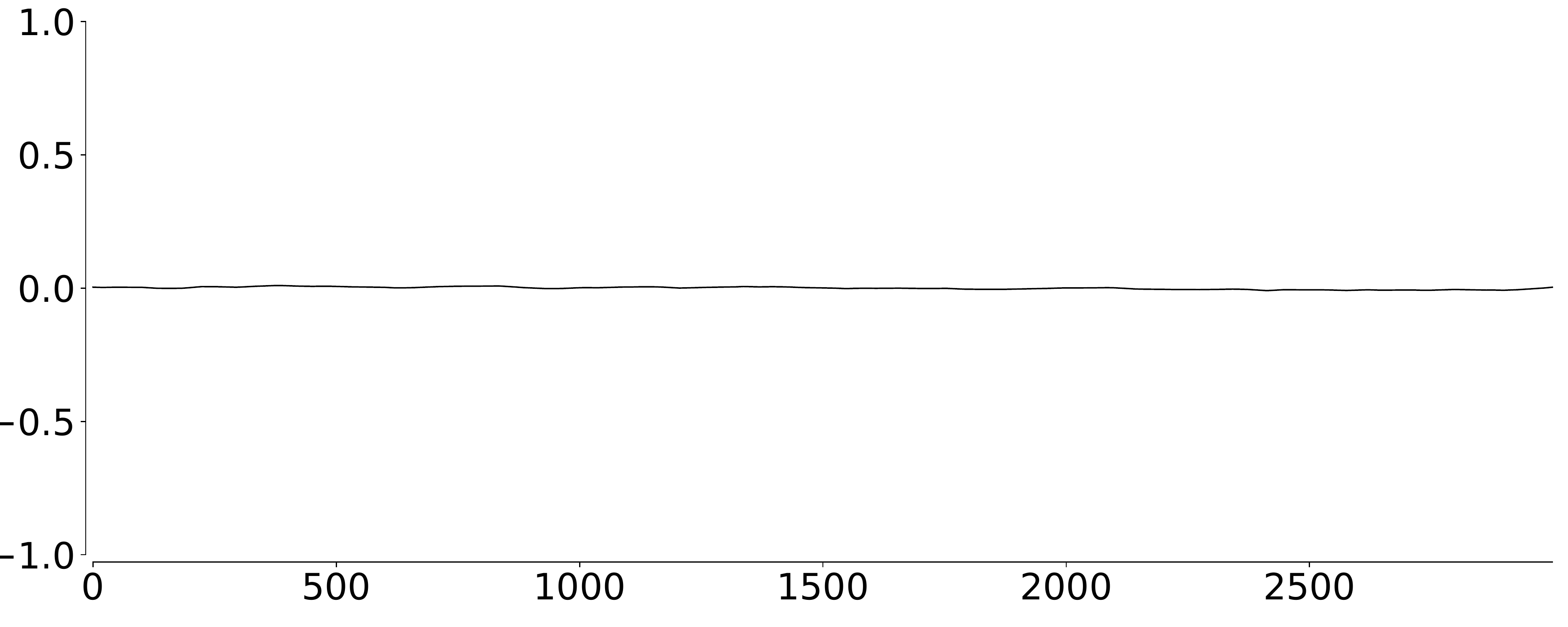}		
		\vspace{-5mm}
	\end{subfigure}	
	\begin{subfigure}[b]{0.235\textwidth}
		\caption{OnlineSTL}				
		\vspace{-2mm}				
		\includegraphics[width=\textwidth]{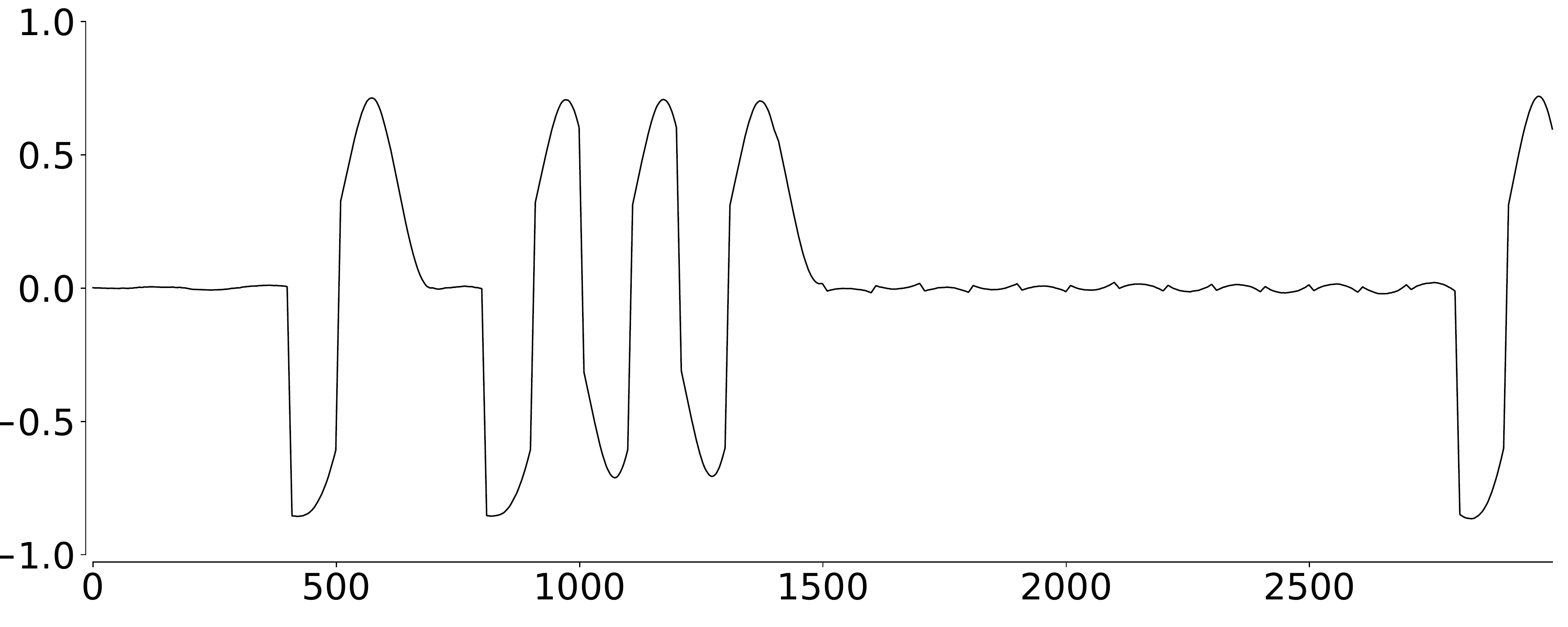}		
		\vspace{-5mm}
	\end{subfigure}	
	\begin{subfigure}[b]{0.235\textwidth}
		\caption{OnlineRobustSTL}				
		\vspace{-2mm}				
		\includegraphics[width=\textwidth]{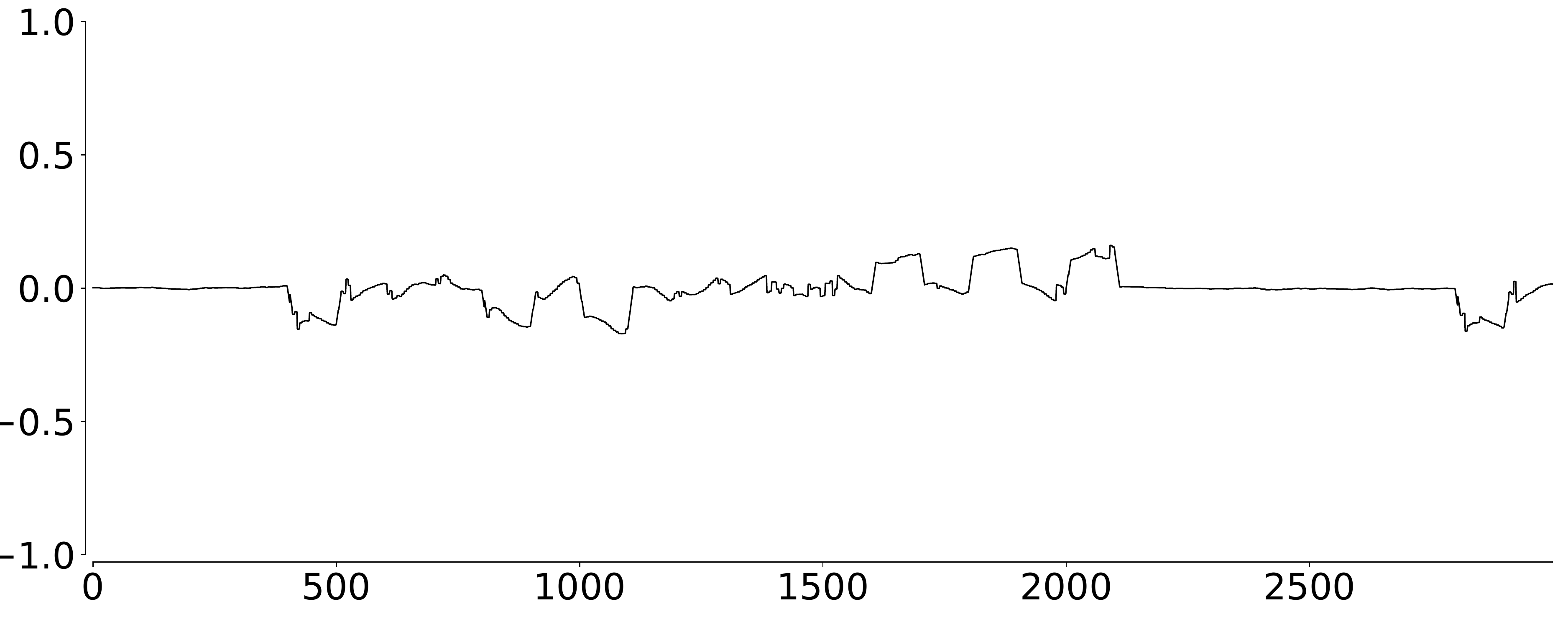}		
		\vspace{-5mm}
	\end{subfigure}	
	\begin{subfigure}[b]{0.235\textwidth}
		\caption{OneShotSTL}				
		\vspace{-2mm}				
		\includegraphics[width=\textwidth]{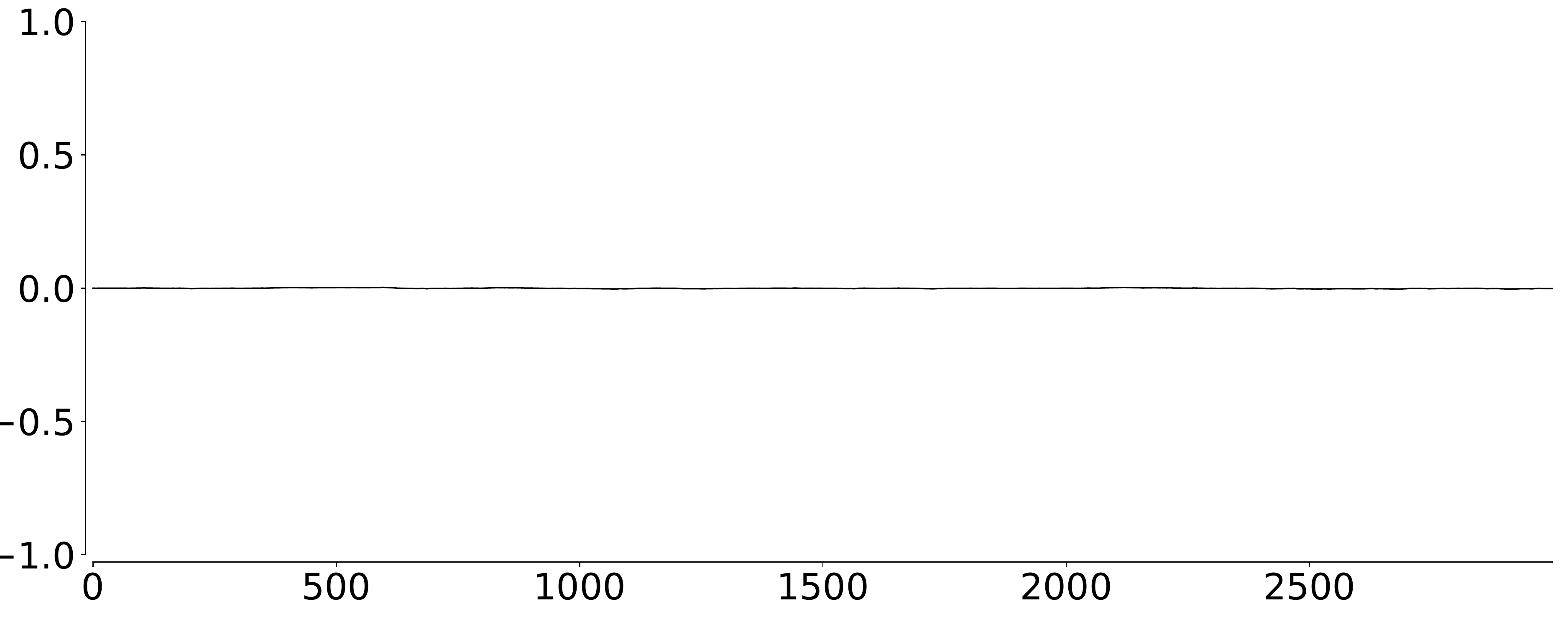}		
		\vspace{-5mm}
	\end{subfigure}	
	
	\begin{subfigure}[b]{0.235\textwidth}
		\includegraphics[width=\textwidth]{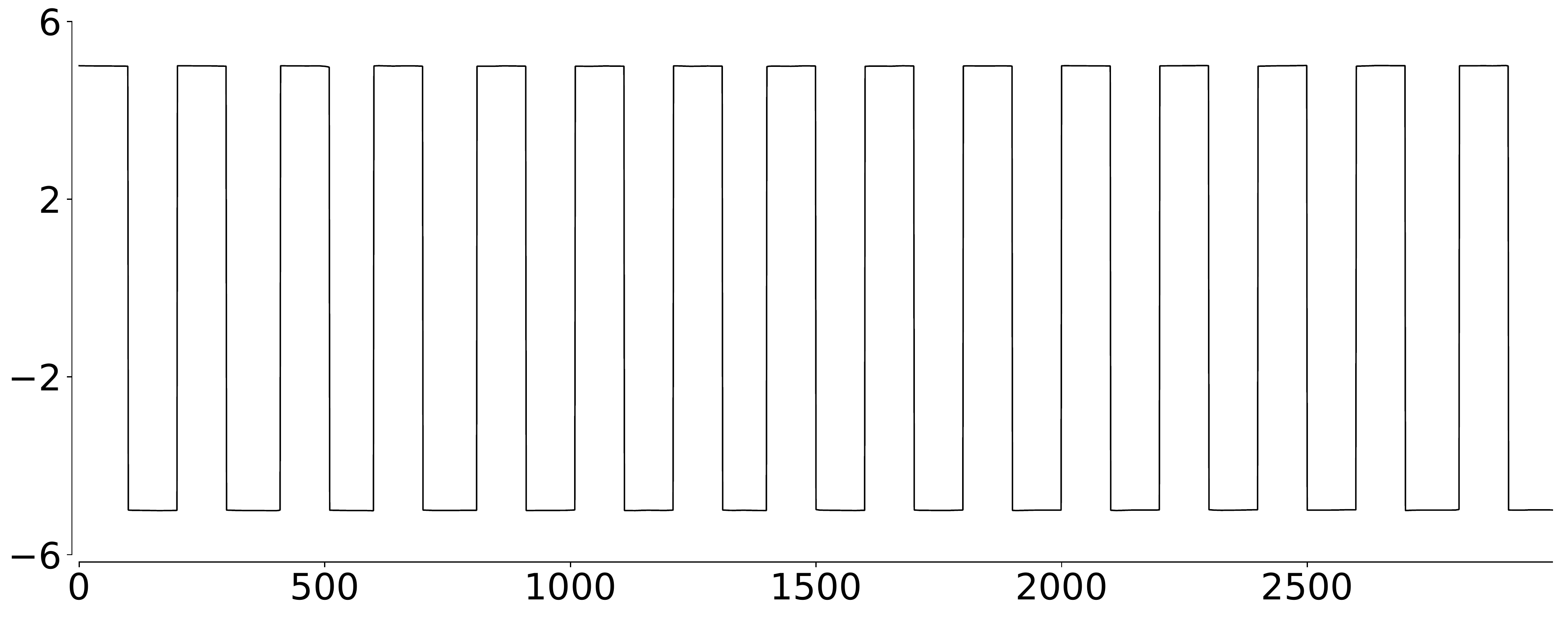}
		\vspace{-5mm}
	\end{subfigure}
	\begin{subfigure}[b]{0.235\textwidth}
		\includegraphics[width=\textwidth]{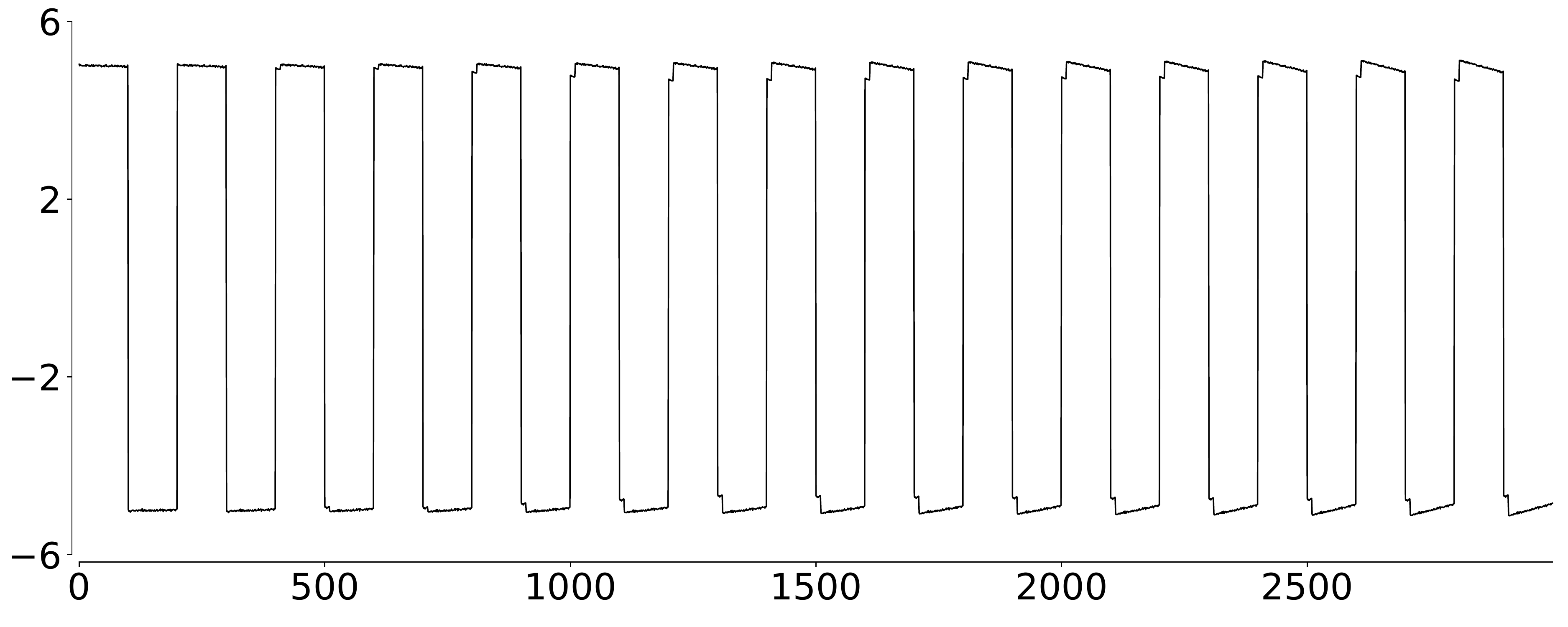}
		\vspace{-5mm}
	\end{subfigure}
	\begin{subfigure}[b]{0.235\textwidth}
		\includegraphics[width=\textwidth]{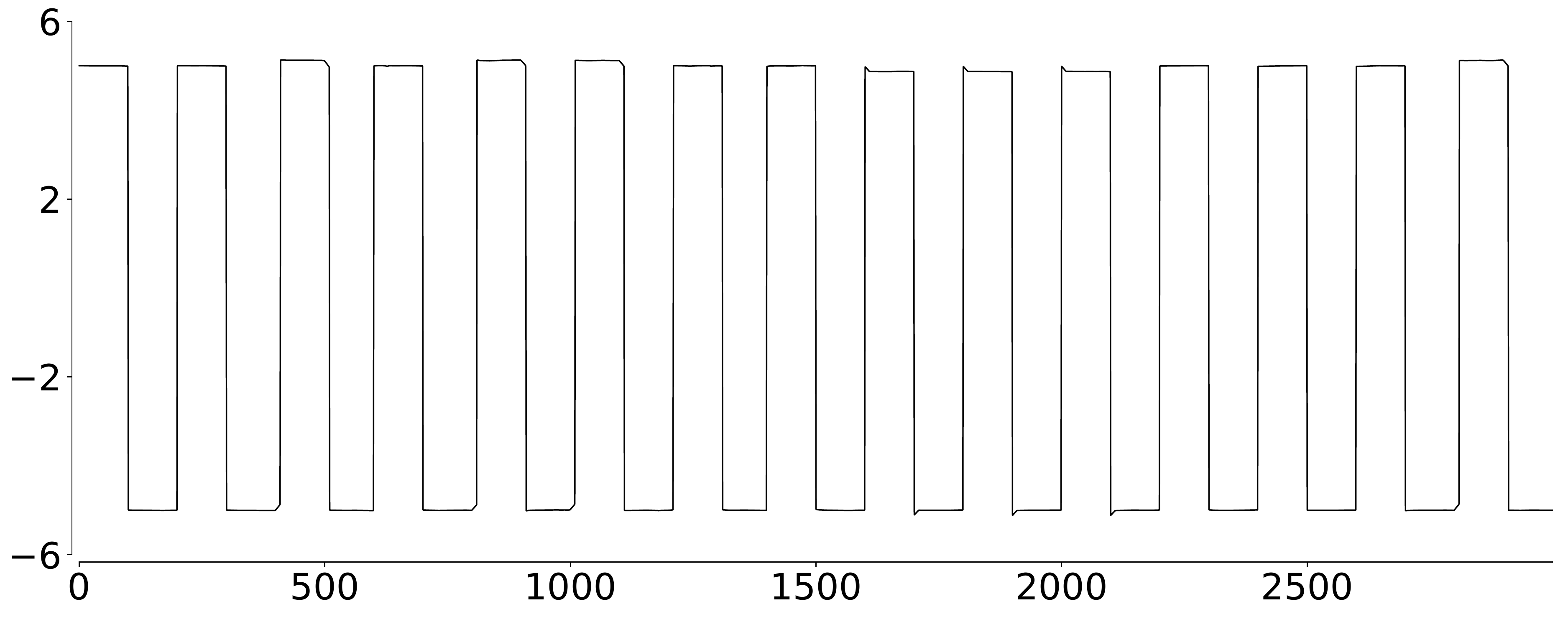}
		\vspace{-5mm}
	\end{subfigure}
	\begin{subfigure}[b]{0.235\textwidth}
		\includegraphics[width=\textwidth]{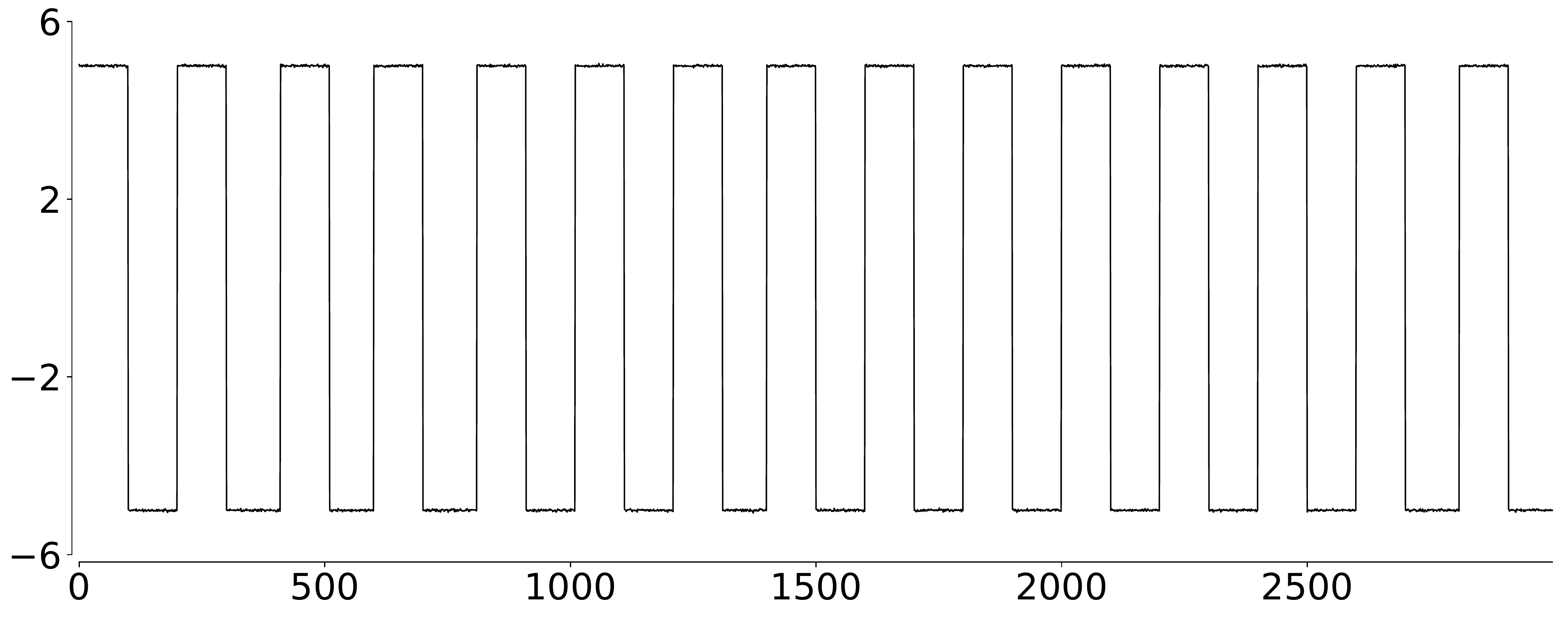}
		\vspace{-5mm}
	\end{subfigure}
	
	\begin{subfigure}[b]{0.235\textwidth}
		\includegraphics[width=\textwidth]{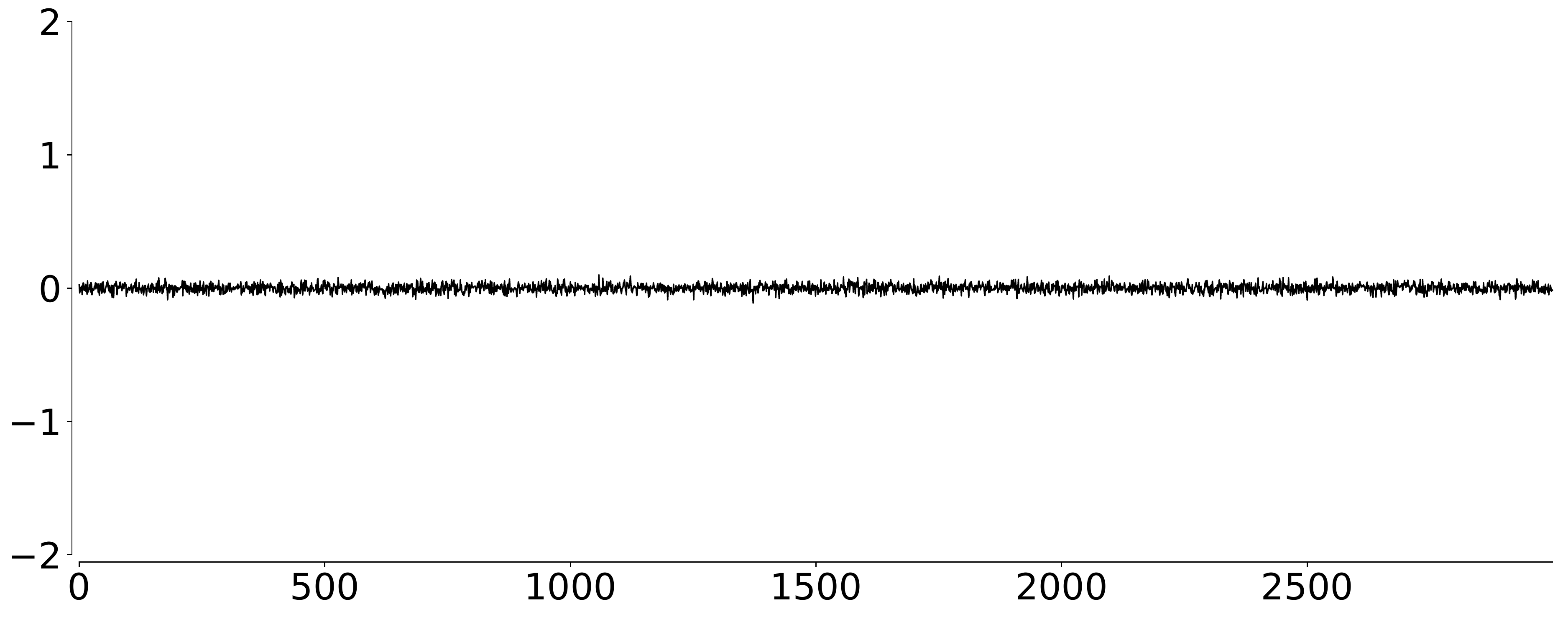}
		\vspace{-5mm}
	\end{subfigure}
	\begin{subfigure}[b]{0.235\textwidth}
		\includegraphics[width=\textwidth]{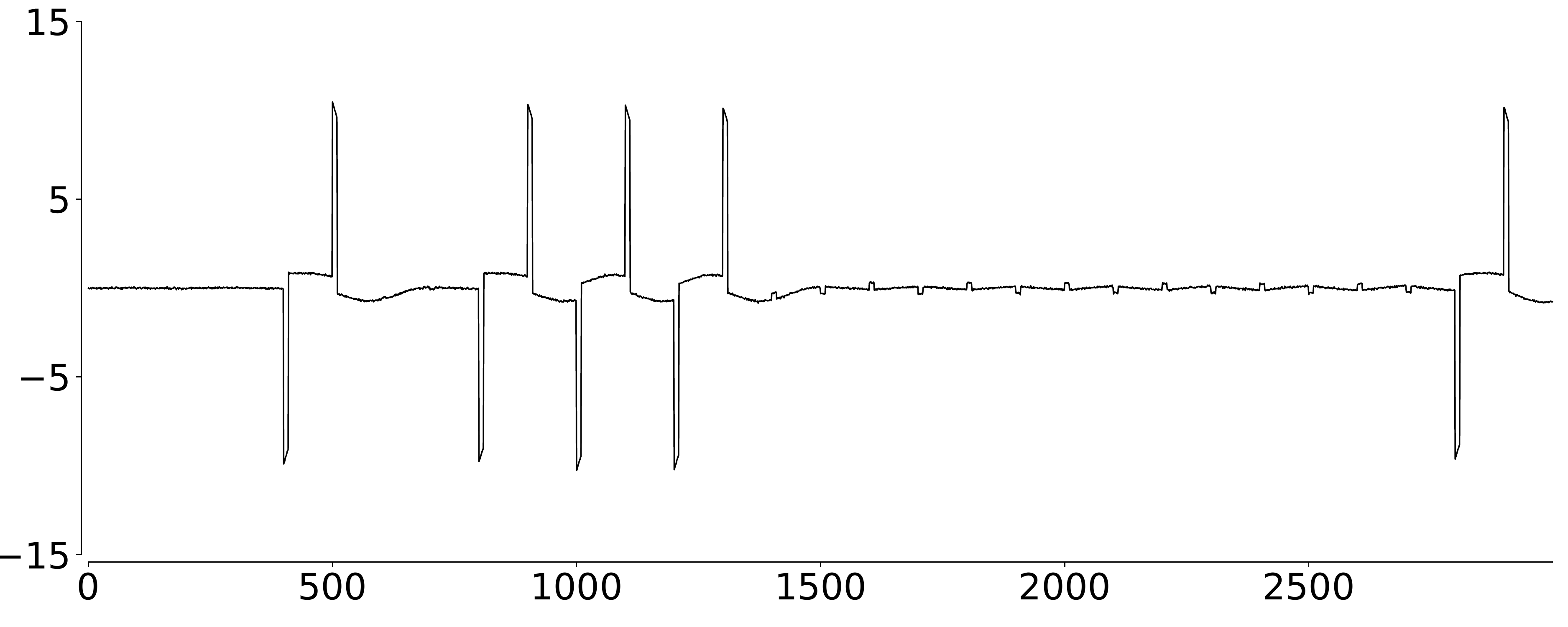}
		\vspace{-5mm}
	\end{subfigure}
	\begin{subfigure}[b]{0.235\textwidth}
		\includegraphics[width=\textwidth]{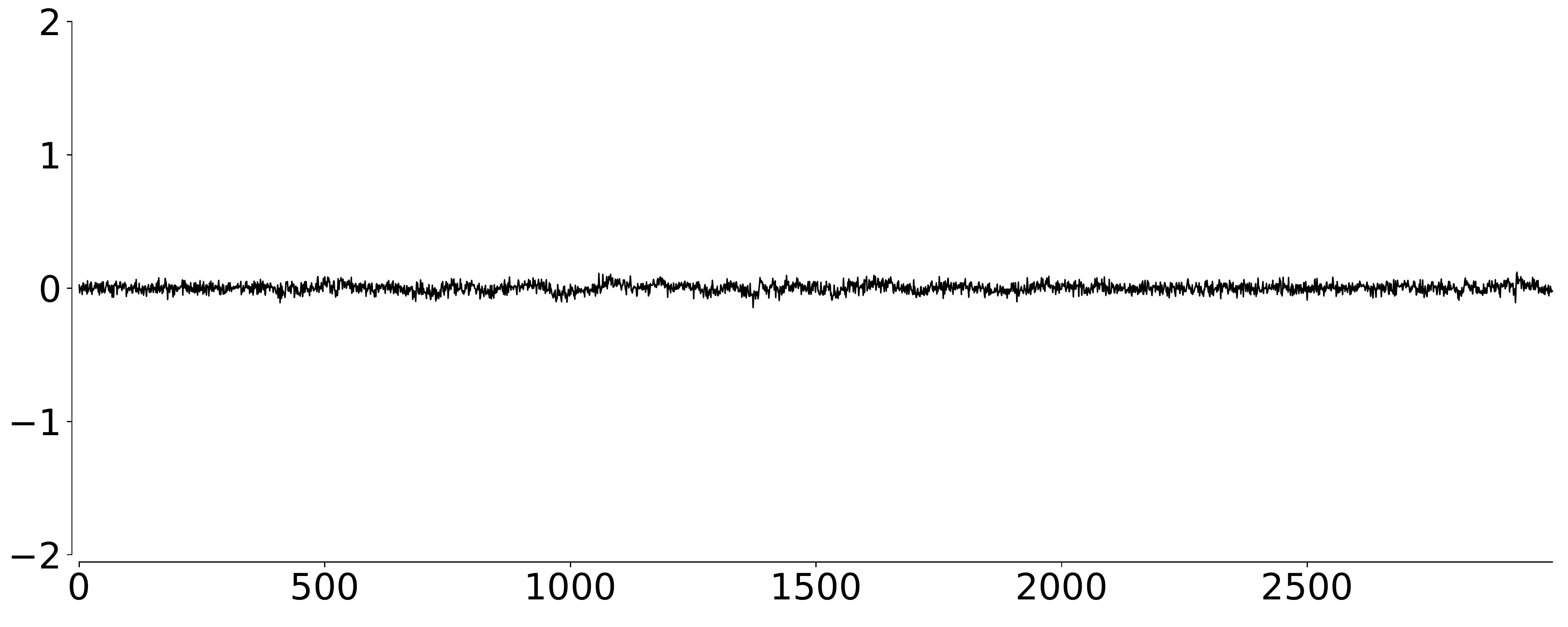}	
		\vspace{-5mm}
	\end{subfigure}
	\begin{subfigure}[b]{0.235\textwidth}
		\includegraphics[width=\textwidth]{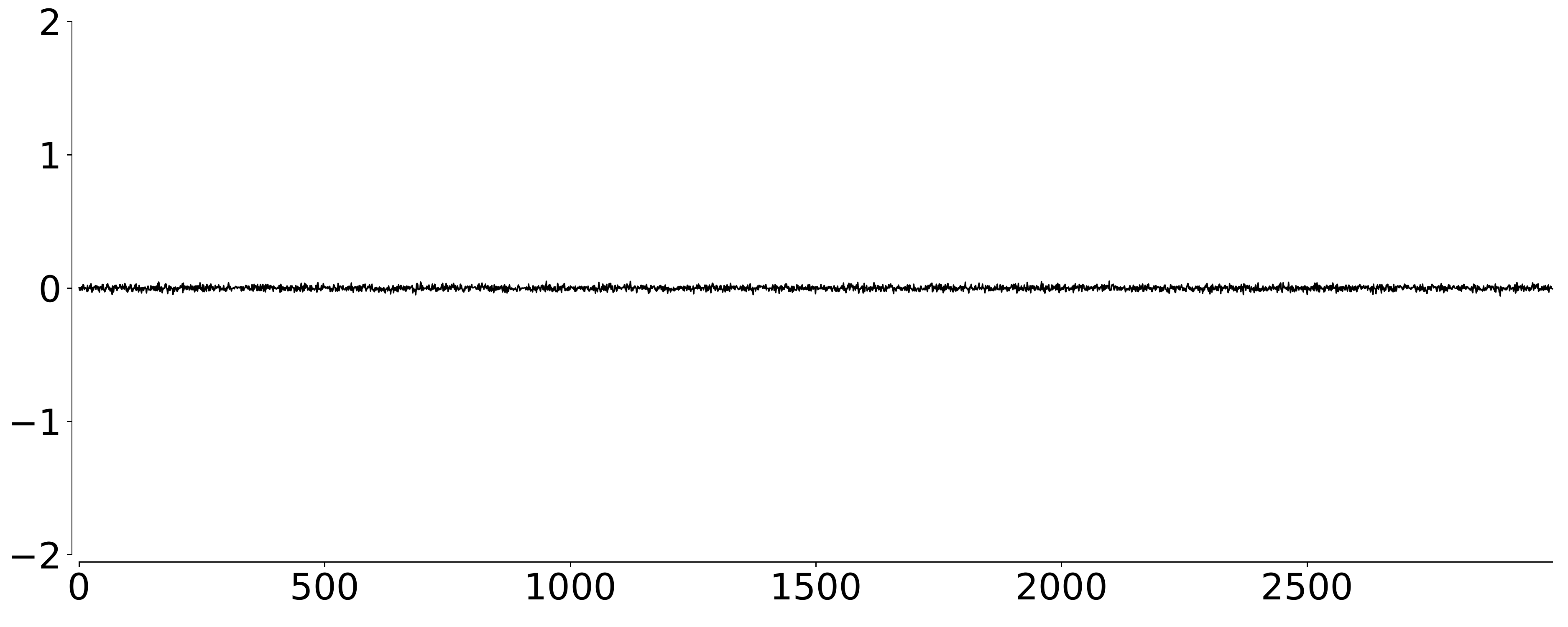}	
		\vspace{-5mm}
	\end{subfigure}
	\vspace{-2mm}
	\caption{Comparison of decomposition results on Syn1 ((a)-(d)) and Syn2 ((e)-(h)).}
	\vspace{-4mm}	
	\label{fig:synthetic2_visual}
\end{figure*}

\subsubsection{\textbf{Comparison Methods}} 
We compare OneShotSTL with the state-of-the-art STD, TSAD, and TSF baseline methods.

\textbf{1. STD Baselines.} We compare OneShotSTL with two batch STD methods (STL~\cite{cleveland90} and RobustSTL~\cite{DBLP:journals/corr/abs-1812-01767,DBLP:conf/kdd/WenZL020}) and two online methods (OnlineSTL~\cite{DBLP:journals/pvldb/MishraSZ22} and OnlineRobustSTL~\cite{DBLP:conf/kdd/WenZL020,sreworks}). For the online setting, STL and RobustSTL can also be used on the sliding window of the most recent data points, which are named Window-STL and Window-RobustSTL. We use the public Java implementation of STL~\cite{javastl} and implement OnlineSTL and OneShotSTL in Java. For RobustSTL we use the Python implementation that is publicly available in package SREWorks~\cite{sreworks}. 

\textbf{2. TSAD Baselines.} We compare with matrix profile-based methods NormA~\cite{DBLP:journals/vldb/BoniolLRPMR21}, STOMPI~\cite{DBLP:journals/datamine/YehZUBDDZSMK18}, SAND~\cite{DBLP:journals/pvldb/BoniolPPF21}, and DAMP~\cite{DBLP:conf/kdd/Lu00ZK22}, which show state-of-the-art performance for univariate time series anomaly detection on the TSB-UAD benchmark. NormA is a batch TSAD method, while STOMPI, SAND, and DAMP are online TSAD methods. A batch method reads all the data of a time series and output an anomaly score for each data point, while an online method uses the training data for initialization and update the model during online detection. Further, we compare with deep learning-based TSAD methods including, LSTM~\cite{DBLP:journals/ral/ParkHK18}, USAD~\cite{DBLP:conf/kdd/AudibertMGMZ20} and TranAD~\cite{DBLP:journals/pvldb/TuliCJ22}. USAD and TranAD are recent methods designed for fast training. We use the Python implementation of LSTM, NormA, and SAND from the TSB-UAD benchmark~\cite{DBLP:journals/pvldb/PaparrizosKBTPF22}. For USAD and TranAD we use the publicly available Python code provided by the authors from \cite{usad} and \cite{tranad} respectively. 
For STOMPI we use the code from {\it stumpy}~\cite{law2019stumpy} Python package. 
Damp is implemented in Matlab and is publicly available~\cite{damp}. 


\textbf{3. TSF Baselines.} We have selected six TSF methods for comparison, including ARIMA~\cite{10.1109/UKSim.2014.67}, DeepAR~\cite{salinas2020deepar}, NBEATS~\cite{DBLP:conf/iclr/OreshkinCCB20}, Informer~\cite{zhou2021informer}, FEDformer~\cite{DBLP:conf/icml/ZhouMWW0022}, and FiLM~\cite{DBLP:journals/corr/abs-2205-08897}. ARIMA is a traditional statistical method and the last five are deep learning-based methods, including the most recent ones ( e.g., FEDformer and FiLM) with state-of-the-art forecasting accuracies on the evaluated datasets. For ARIMA we use the implementation from {\it statsforecast}~\cite{statsforecast} Python package. For DeepAR and NBEATS, we use the code from {\it pytorch-forecasting}~\cite{pytorch-forecast}. For Informer~\cite{zhou2021informer}, FEDformer~\cite{DBLP:conf/icml/ZhouMWW0022} and FiLM~\cite{DBLP:journals/corr/abs-2205-08897}, we use the  Python codes from the authors, which are publicly available.


\subsubsection{\textbf{Evaluation Metrics}} We also have multiple evaluation metrics for different time series analysis tasks.

\textbf{1. Decomposition Metric.} We adopt the Mean Absolute Error (MAE) between the decomposed and the ground truth signals for synthetic datasets to evaluate the decomposition quality. Further, we show the visualization of the decomposed signals for both synthetic and real-world STD datasets.

\textbf{2. Univariate TSAD Metric.} Recently, there are studies arguing that the traditional evaluation metrics (e.g., Precision, Recall, and AUC) of TSAD methods are too sensitive to the noisy labels~\cite{DBLP:conf/aaai/KimCCLY22,paparrizos2022volume}. Therefore, we use a recently proposed evaluation method VUS-ROC~\cite{paparrizos2022volume} (also from the authors of TSB-UAD) for evaluations. VUS-ROC has been shown to be more robust against noisy labels than other TSAD evaluation metrics~\cite{paparrizos2022volume}. Moreover, on KDD21 dataset~\cite{UCRArchive2018} we use the score function from the KDD CUP 2021 TSAD competition for evaluation. Since there is only one anomaly event for each time series in KDD21 datasets, it just checks whether the top-ranked anomaly point is among the neighborhood of the labeled anomaly event. 


\textbf{3. Univariate TSF Metric.} We adopt the MAE between the real and the predicted values in the test set for the evaluation of the forecasting task.

\subsubsection{\textbf{Hyper-parameters setting}} All STD and TSAD methods (except the deep learning methods) need the period length $T$ as the input. For synthetic datasets, we use the ground truth value $T=500$ for \textbf{Syn1} and $T=250$ for \textbf{Syn2}. For real-world datasets, we use the {\it find\_length} function from TSB-UAD package~\cite{findlength} to detect $T$, which is based on auto-correlation.

\textbf{1. STD Methods Setting.} STL does not need any other hyper-parameters. RobustSTL needs a dozen of them and we use the authors-recommended default values~\cite{sreworks}. We set $\alpha=0.7$ for OnlineSTL as the authors suggested~\cite{DBLP:journals/pvldb/MishraSZ22}. The key hyper-parameters for OneShotSTL are the $\lambda_1$ and $\lambda_2$ which control the smoothness of the decomposed trend. We always set them the same with $\lambda=\lambda_1=\lambda_2$ and tune $\lambda$ as follows. On the training data, we perform STL and OneShotSTL with $\lambda \in \{10^0, 10^1, ... , 10^4\}$. Then we pick up the one that is the closest (the smallest MAE) to the results from STL. 
We set the seasonality shift window $H=20$, the maximum iteration $I=8$
and $n=5$ for NSigma, if not otherwise specified. 

\textbf{2. TSAD Methods Setting.} 
We use the hyper-parameters suggested by the authors, which usually are the default ones. 
For methods implemented in TSB-UAD, their default hyper-parameters have already been used for the evaluation in the TSB-UAD benchmark paper~\citep{DBLP:journals/pvldb/PaparrizosKBTPF22}.

\textbf{3. TSF Methods Setting.} For Informer, FEDformer, and FiLM, we use the hyper-parameters that have already been extensively tuned on the evaluated datasets by the authors. For Arima, we use the parameter-free AutoArima from {\it statsforecast} \cite{statsforecast} package.
For DeepAR and NBEATS, hyper-parameters of the initial learning rate and hidden sizes are tuned on the validation sets by following the instructions~\cite{pytorch-forecast}. 
For OnlineSTL and OneShotSTL, we tune $\alpha$ and $\lambda_1$ and $\lambda_2$ using the validation sets respectively.


\subsubsection{\textbf{Hardware and setting}}
All experiments of NSigma, DAMP, OnlineSTL, and OneShotSTL are performed on a Macbook Pro 2018 with $4$ CPU cores (Intel Core i5) and $16$ GiB memories.  Experiments for the other methods are conducted on Alibaba cloud where an ECS (ecs.gn6e, $48$ CPU cores, $368$ GiB memories, NVIDIA V100$\times 4$ GPU cards) is used. For all the deep learning-based methods, an NVIDIA V100 GPU card is used for efficient training and testing.


\begin{figure*}[t]	
	\begin{subfigure}[b]{0.235\textwidth}
		\caption{Batch RobustSTL}				
		\vspace{-2mm}				
		\includegraphics[width=\textwidth]{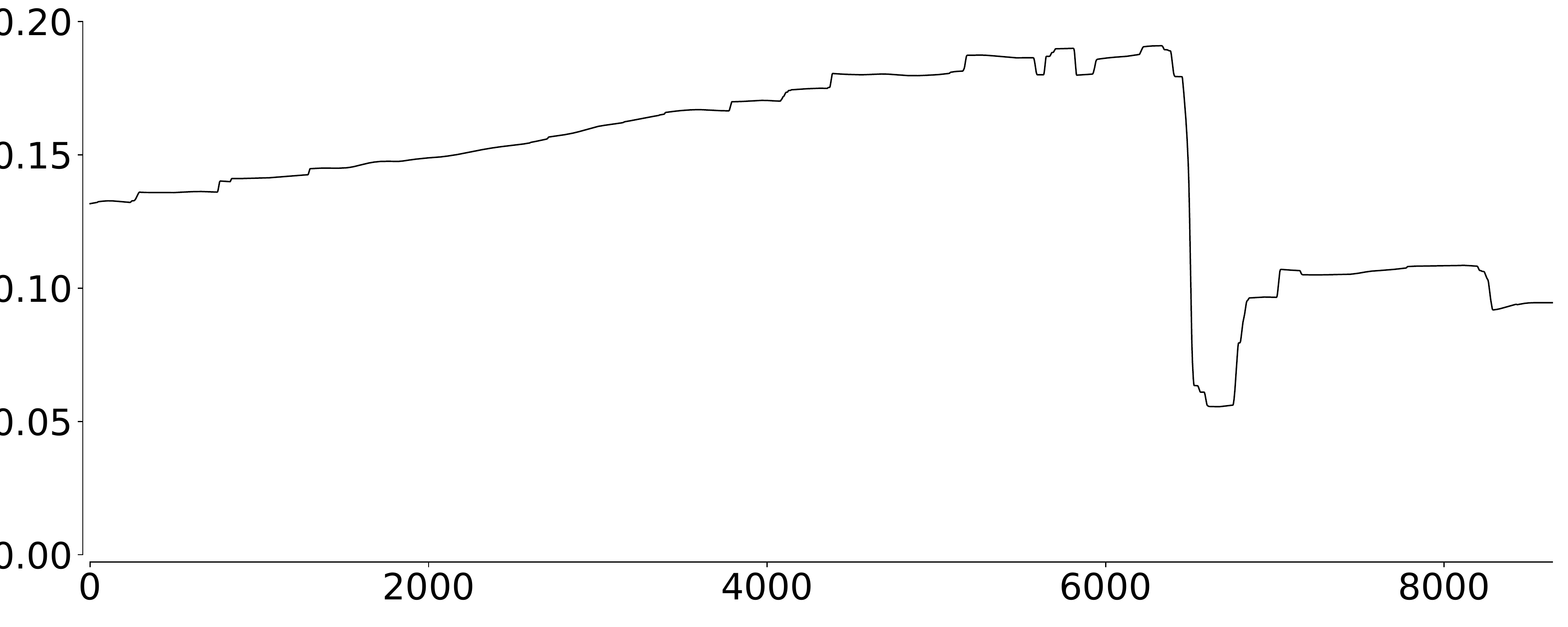}	
		\vspace{-5mm}
	\end{subfigure}	
	\begin{subfigure}[b]{0.235\textwidth}
		\caption{OnlineSTL}				
		\vspace{-2mm}				
		\includegraphics[width=\textwidth]{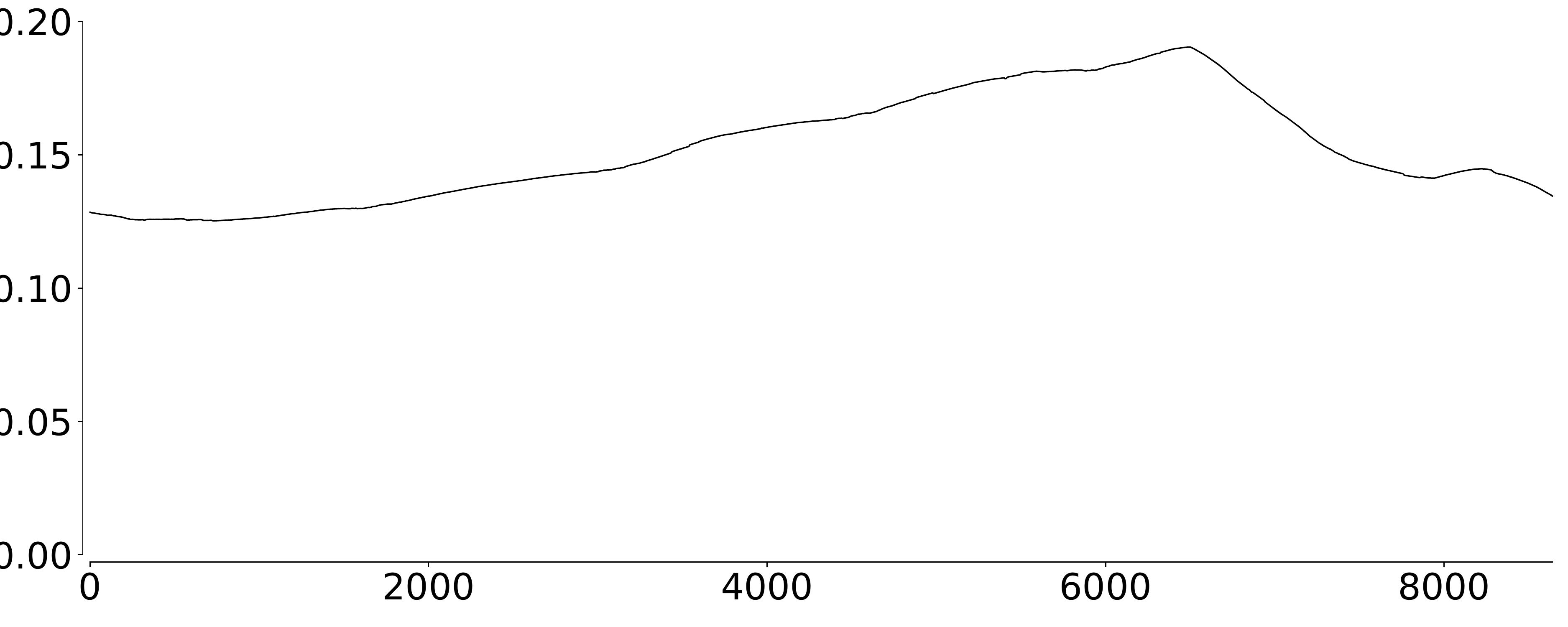}
		\vspace{-5mm}
	\end{subfigure}	
	\begin{subfigure}[b]{0.235\textwidth}
		\caption{OnlineRobustSTL}				
		\vspace{-2mm}				
		\includegraphics[width=\textwidth]{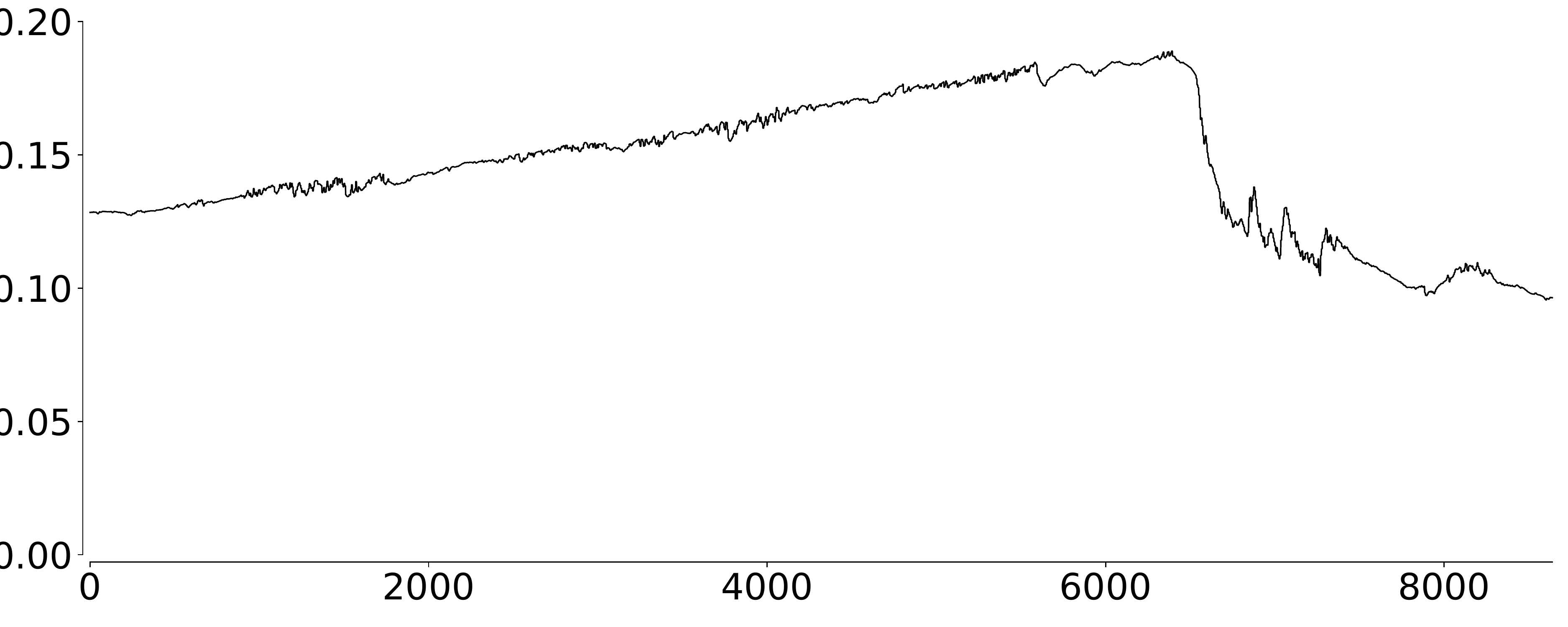}		
		\vspace{-5mm}
	\end{subfigure}	
	\begin{subfigure}[b]{0.235\textwidth}
		\caption{OneShotSTL}				
		\vspace{-2mm}				
		\includegraphics[width=\textwidth]{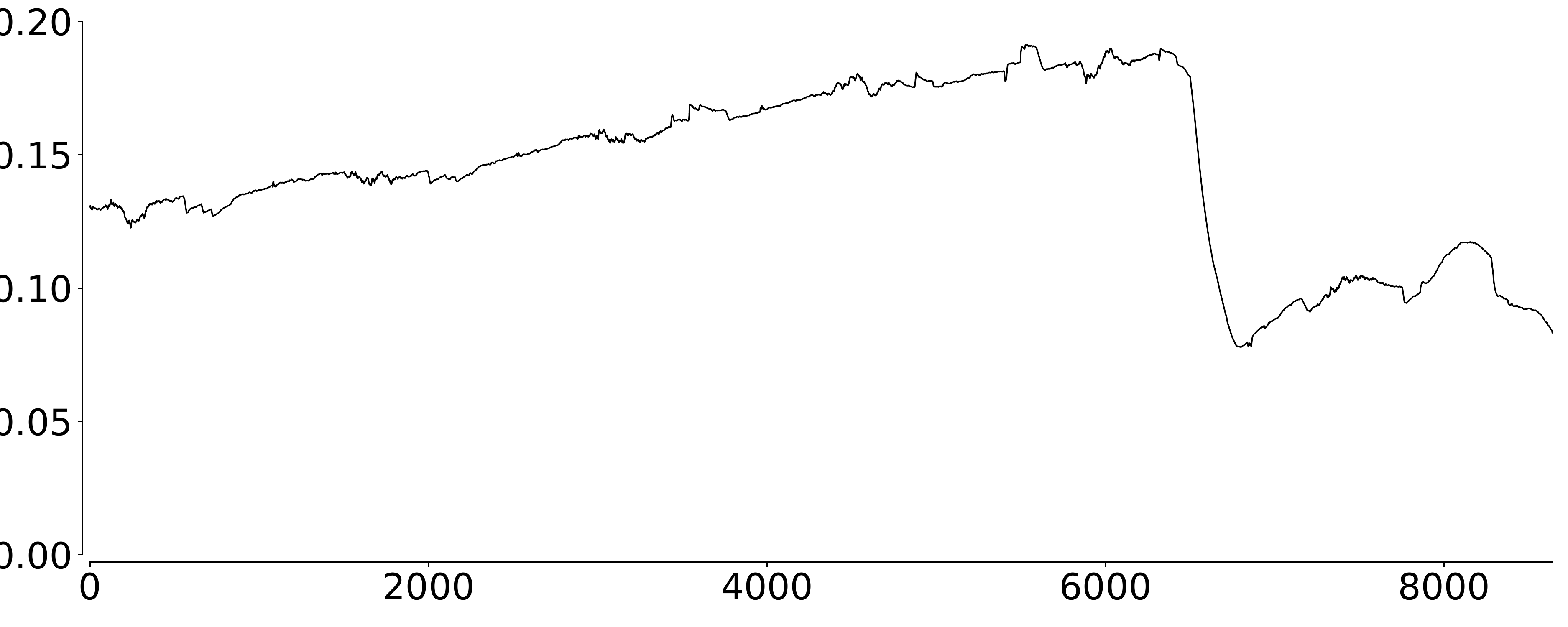}	
		\vspace{-5mm}
	\end{subfigure}	
	
	\begin{subfigure}[b]{0.235\textwidth}
		\includegraphics[width=\textwidth]{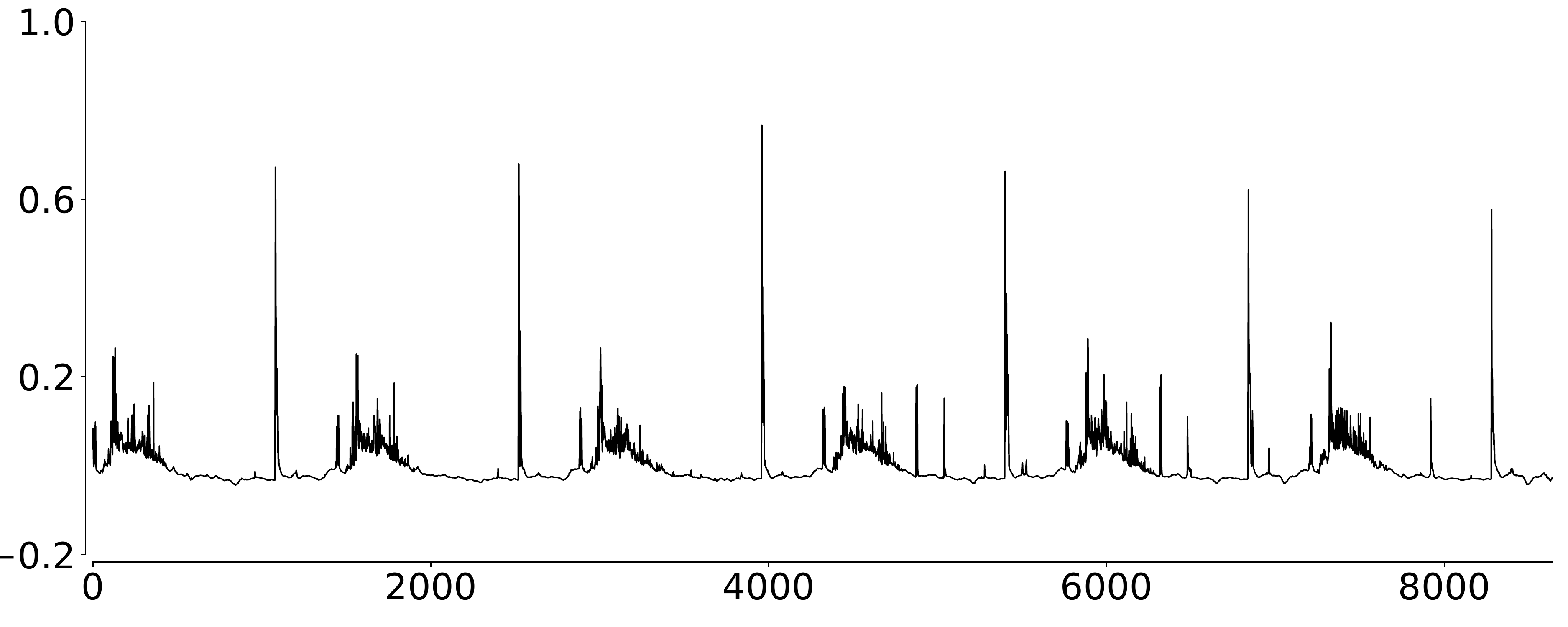}
		\vspace{-5mm}
	\end{subfigure}
	\begin{subfigure}[b]{0.235\textwidth}
		\includegraphics[width=\textwidth]{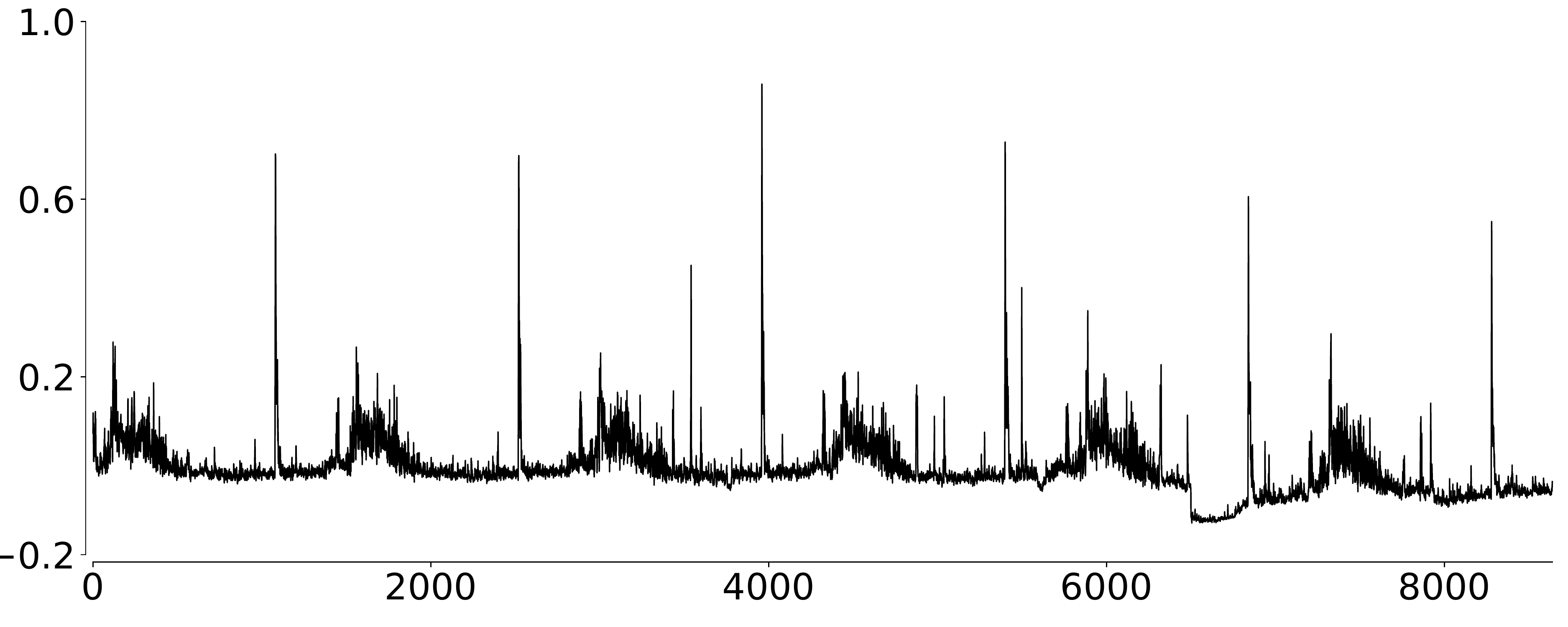}
		\vspace{-5mm}
	\end{subfigure}
	\begin{subfigure}[b]{0.235\textwidth}
		\includegraphics[width=\textwidth]{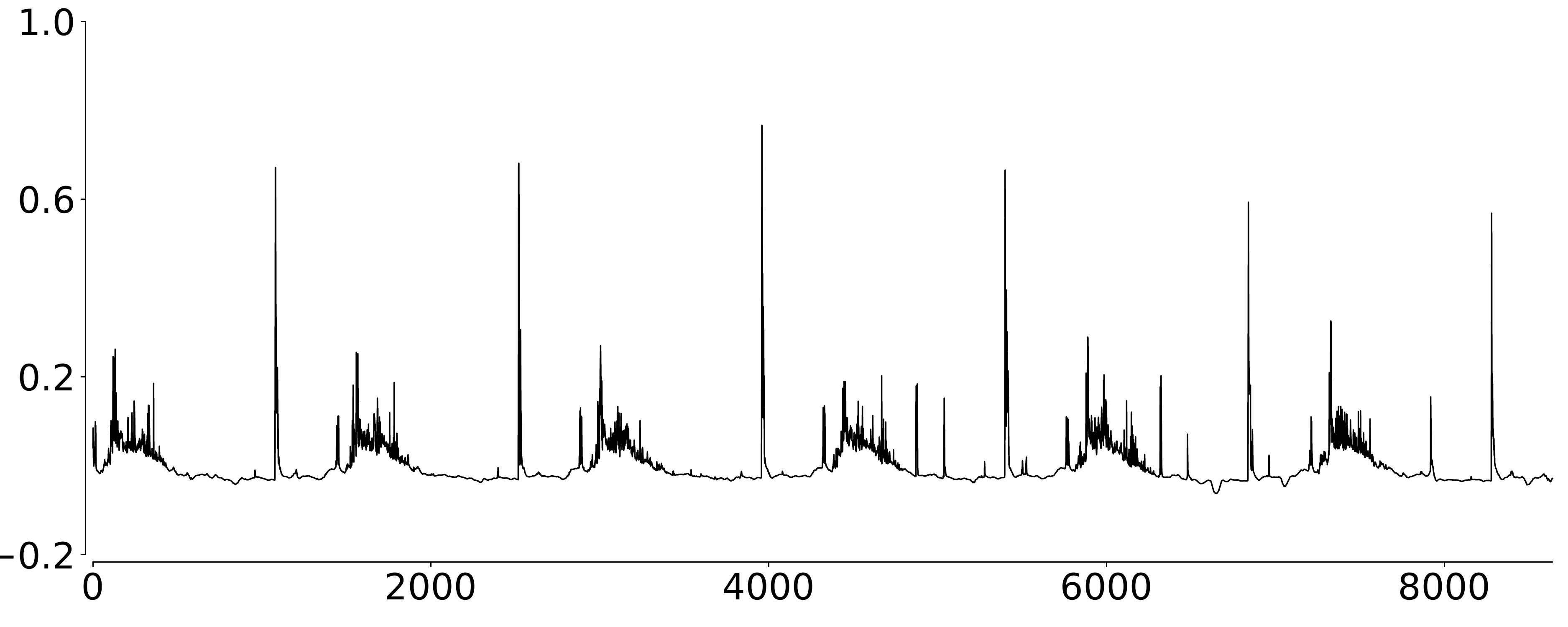}
		\vspace{-5mm}
	\end{subfigure}
	\begin{subfigure}[b]{0.235\textwidth}
		\includegraphics[width=\textwidth]{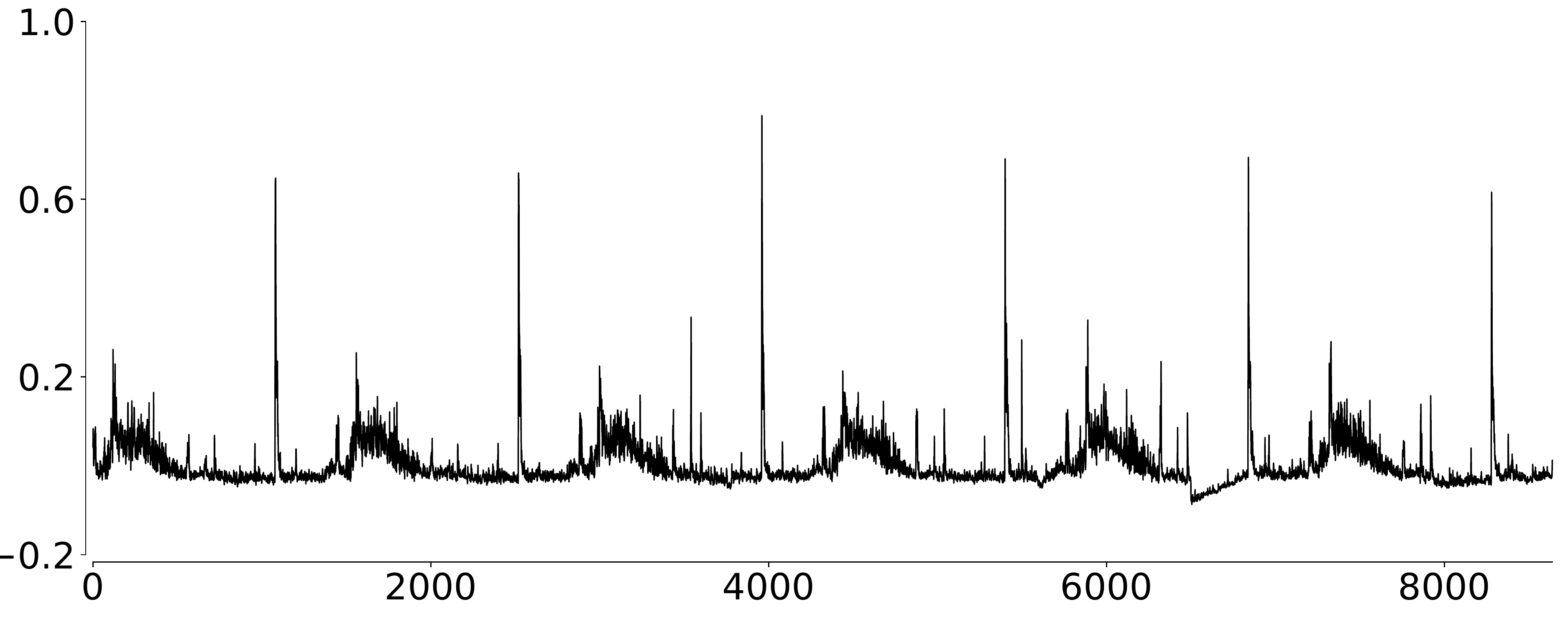}
		\vspace{-5mm}
	\end{subfigure}
	
	\begin{subfigure}[b]{0.235\textwidth}
		\includegraphics[width=\textwidth]{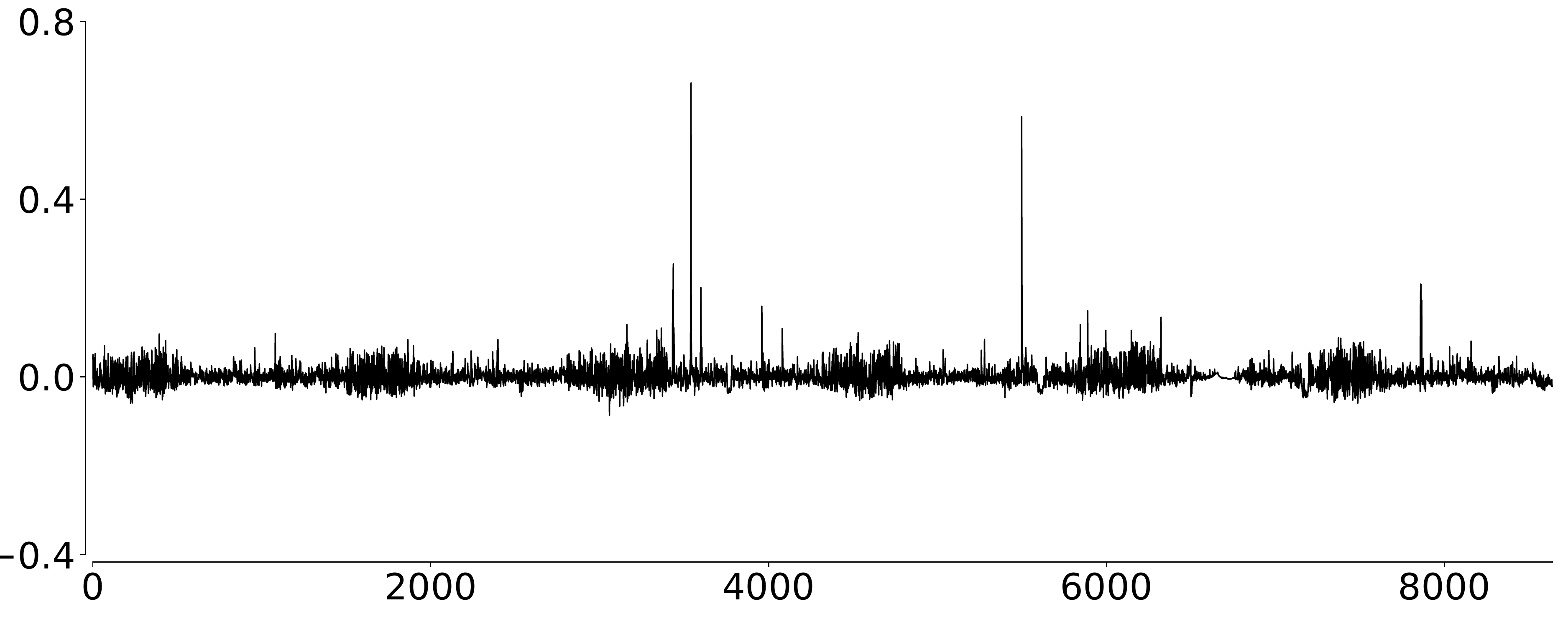}
		\vspace{-5mm}
	\end{subfigure}
	\begin{subfigure}[b]{0.235\textwidth}
		\includegraphics[width=\textwidth]{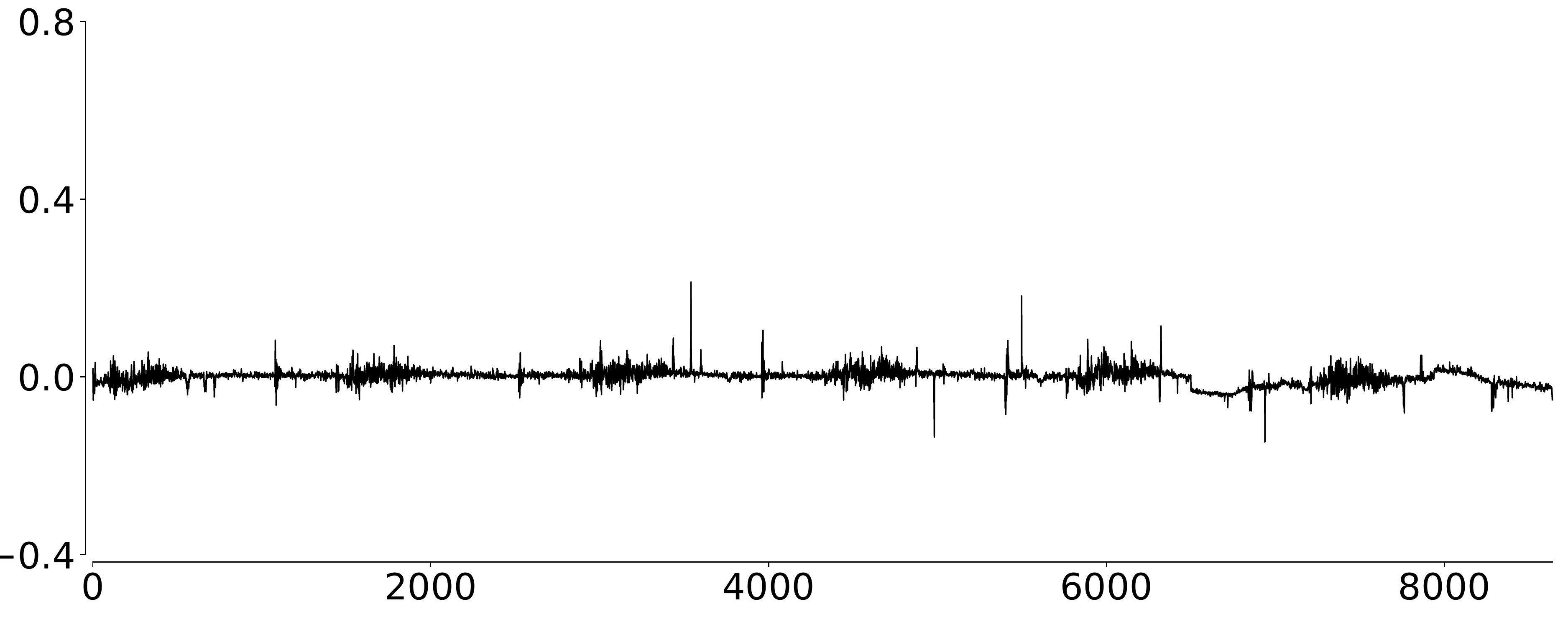}
		\vspace{-5mm}
	\end{subfigure}
	\begin{subfigure}[b]{0.235\textwidth}
		\includegraphics[width=\textwidth]{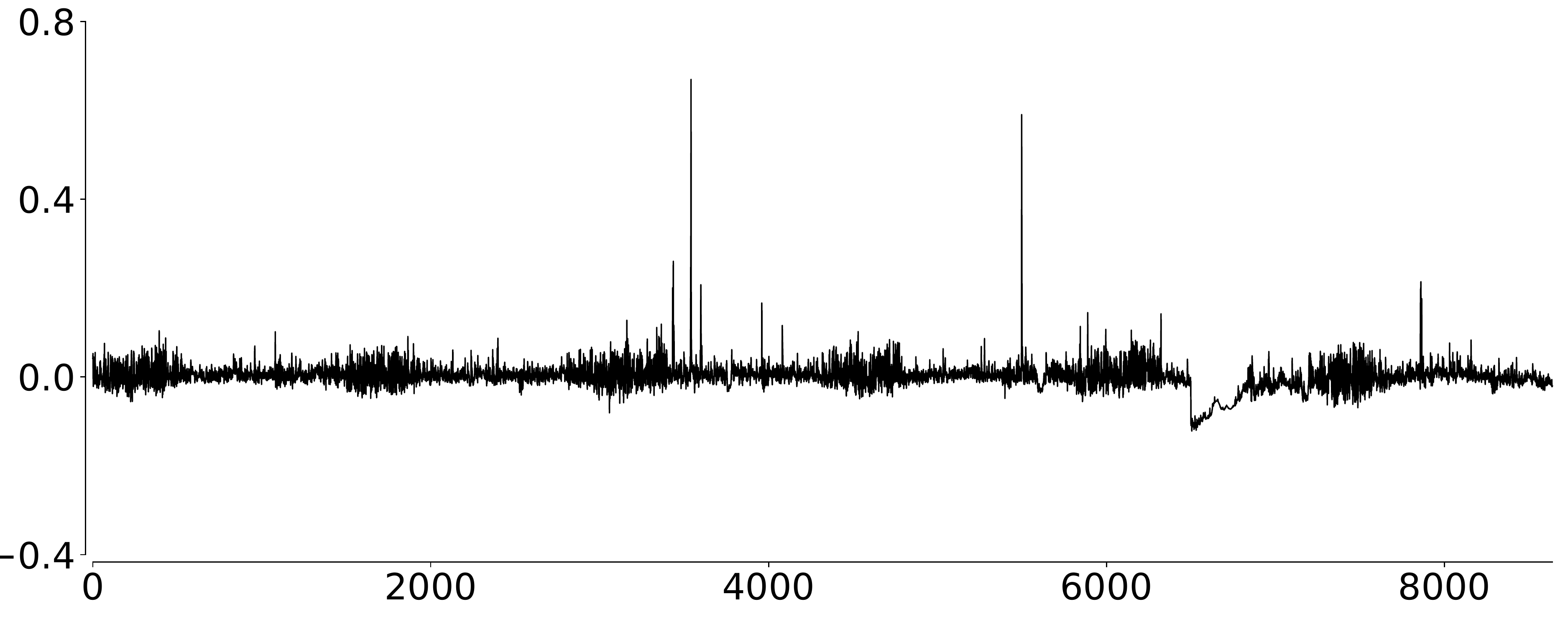}	
		\vspace{-5mm}
	\end{subfigure}
	\begin{subfigure}[b]{0.235\textwidth}
		\includegraphics[width=\textwidth]{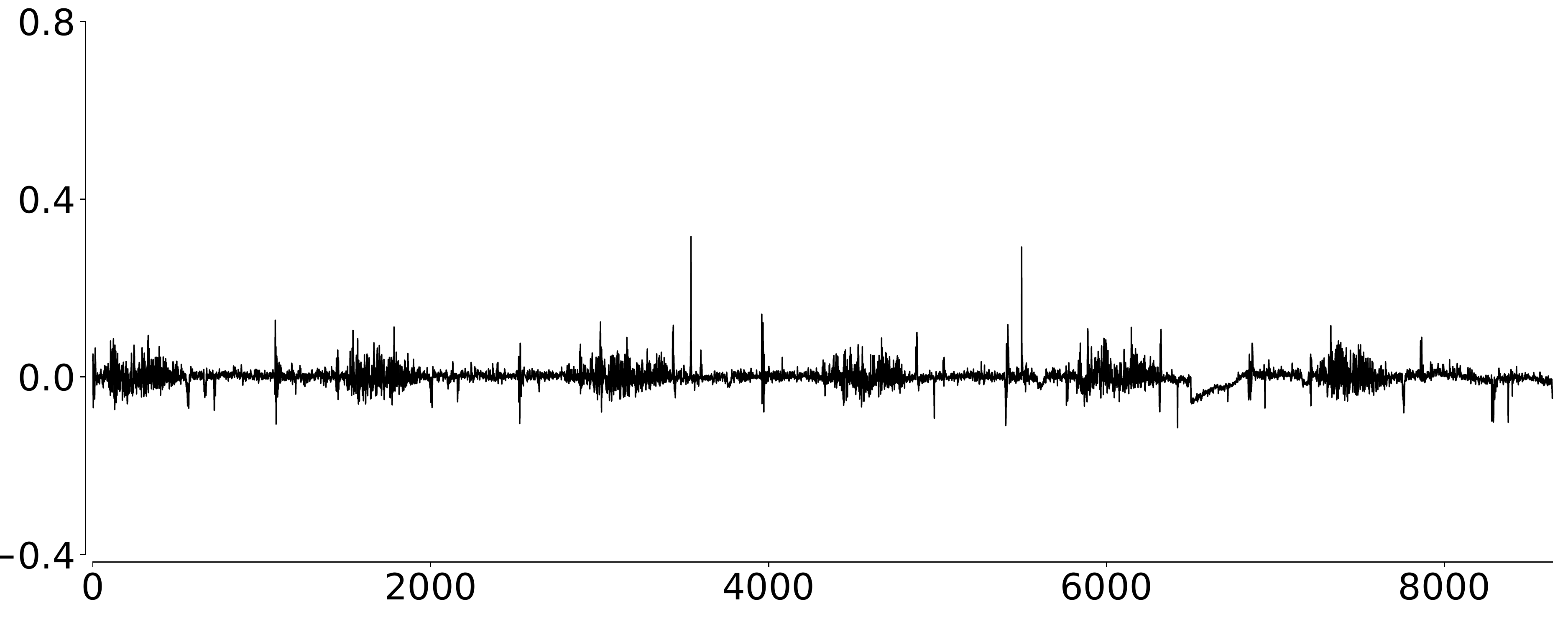}	
		\vspace{-5mm}
	\end{subfigure}
	\begin{subfigure}[b]{0.235\textwidth}
		\caption{Batch RobustSTL}				
		\vspace{-2mm}		
		\includegraphics[width=\textwidth]{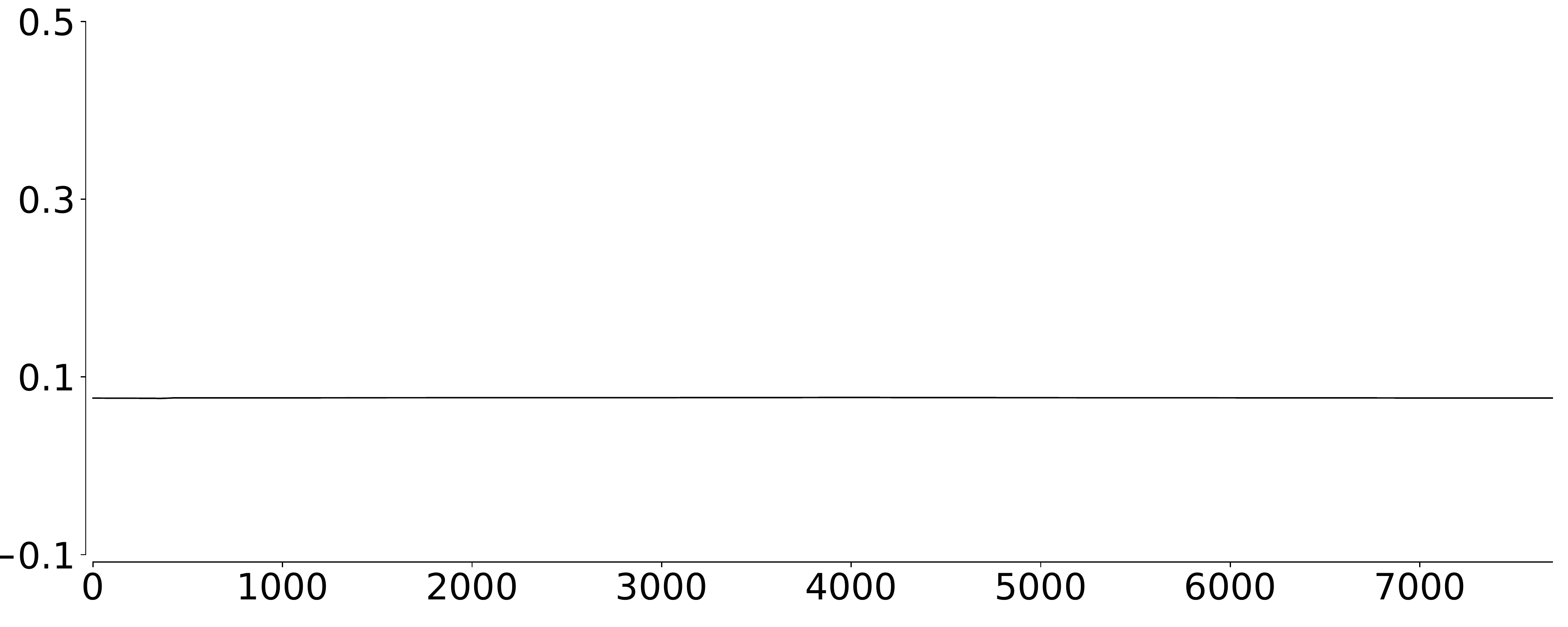}	
		\vspace{-5mm}
	\end{subfigure}	
	\begin{subfigure}[b]{0.235\textwidth}
		\caption{OnlineSTL}				
		\vspace{-2mm}				
		\includegraphics[width=\textwidth]{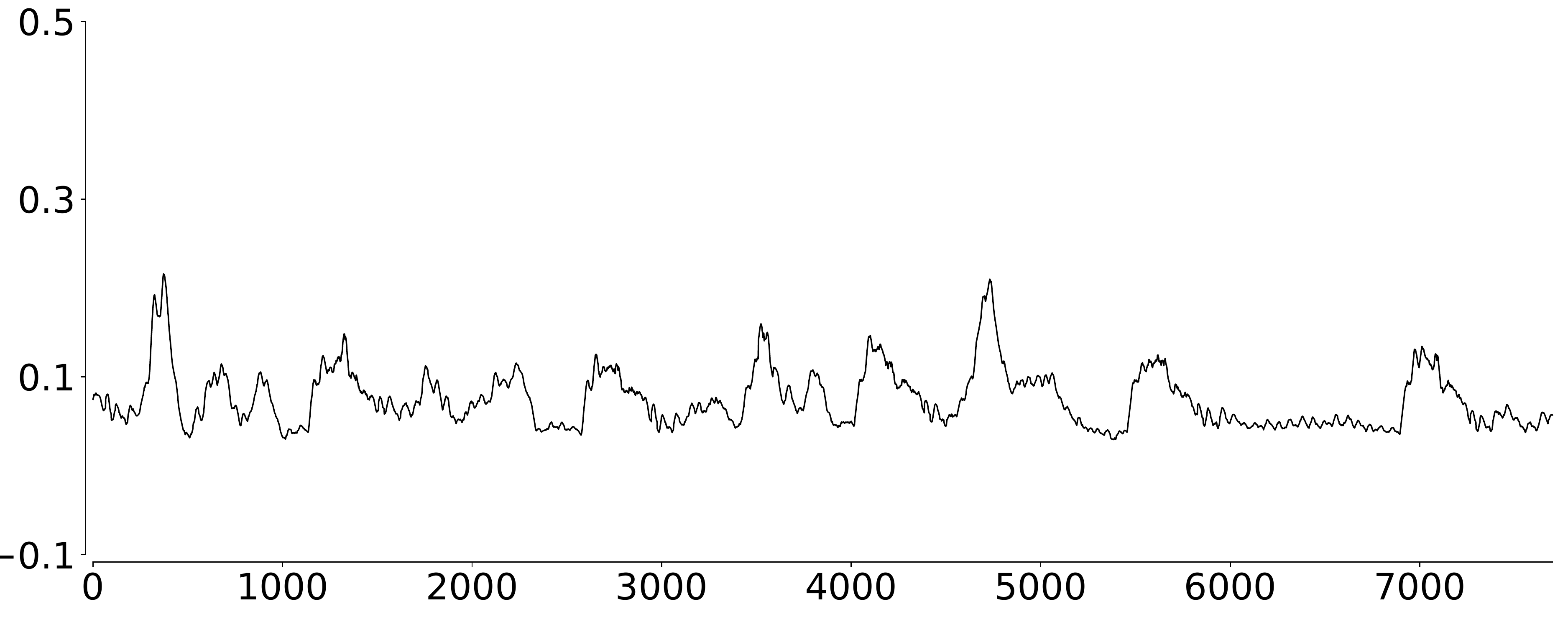}	
		\vspace{-5mm}
	\end{subfigure}	
	\begin{subfigure}[b]{0.235\textwidth}
		\caption{OnlineRobustSTL}		
		\vspace{-2mm}						
		\includegraphics[width=\textwidth]{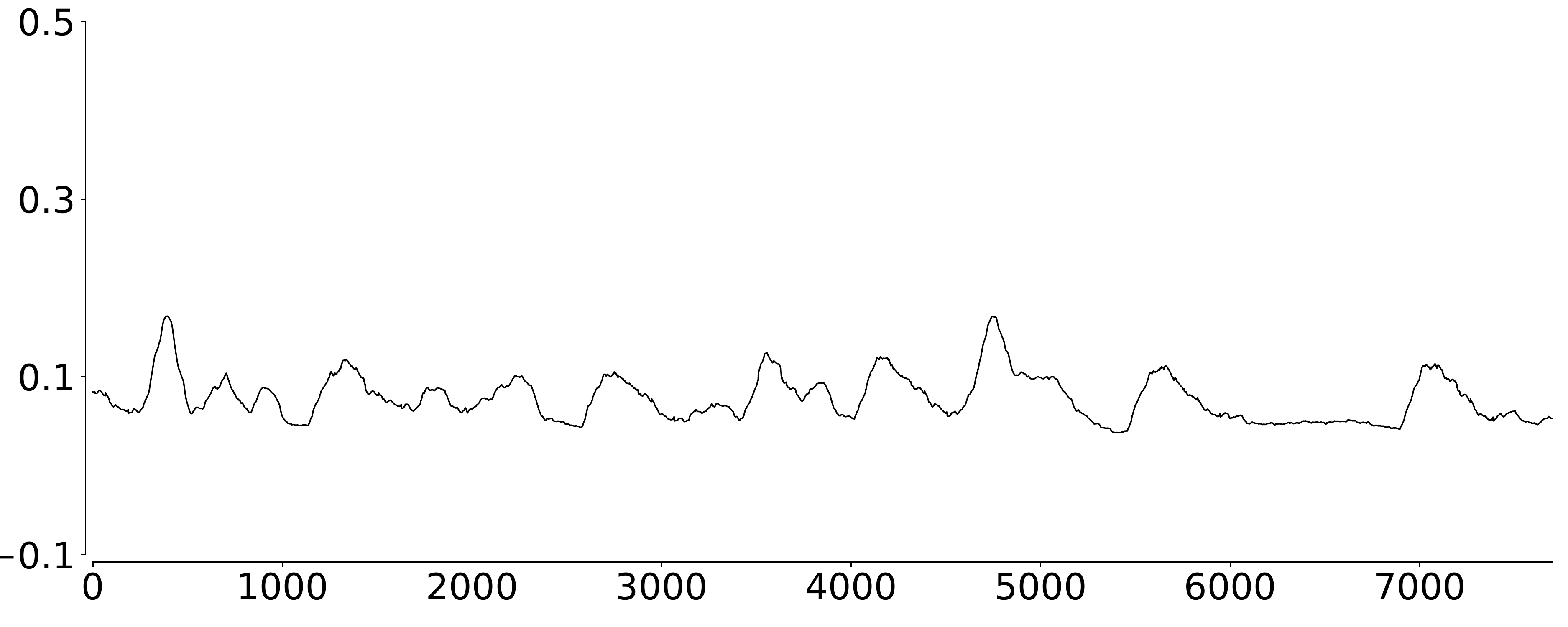}		
		\vspace{-5mm}
	\end{subfigure}	
	\begin{subfigure}[b]{0.235\textwidth}
		\caption{OneShotSTL}		
		\vspace{-2mm}						
		\includegraphics[width=\textwidth]{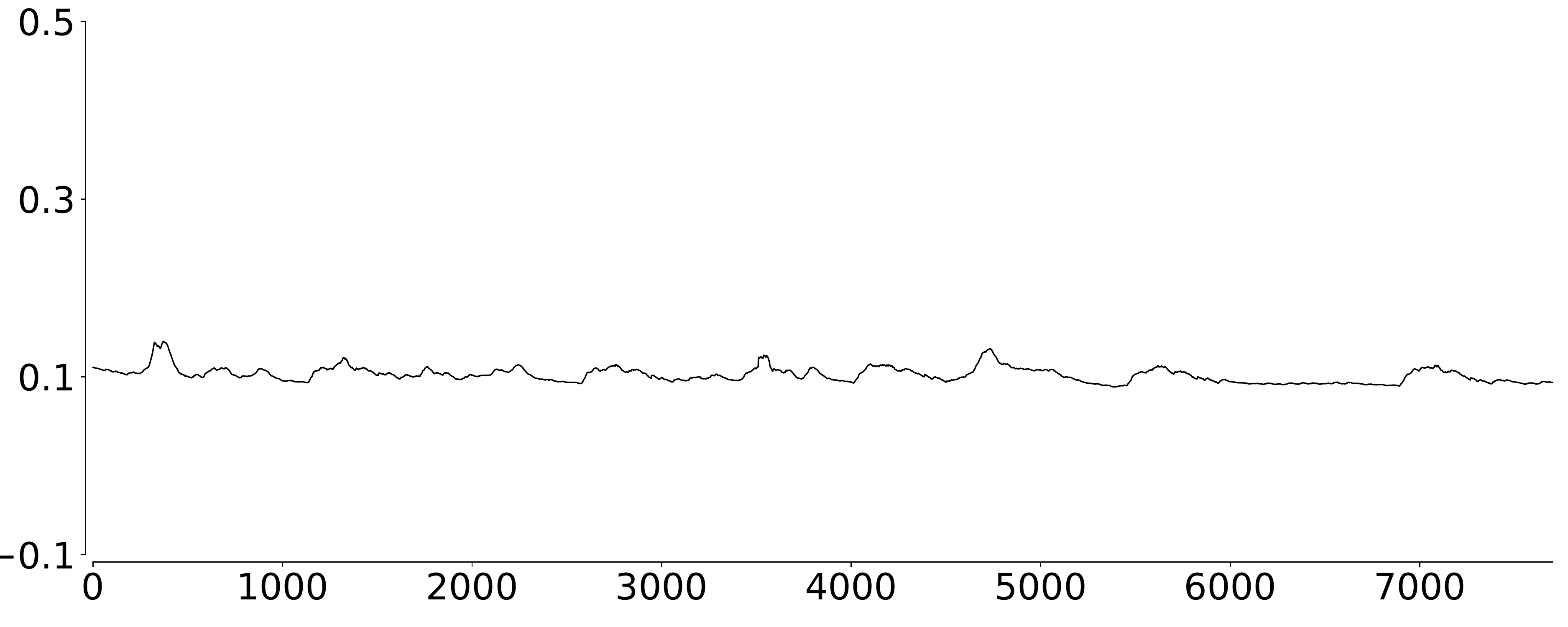}	
		\vspace{-5mm}
	\end{subfigure}	
	
	\begin{subfigure}[b]{0.235\textwidth}
		\includegraphics[width=\textwidth]{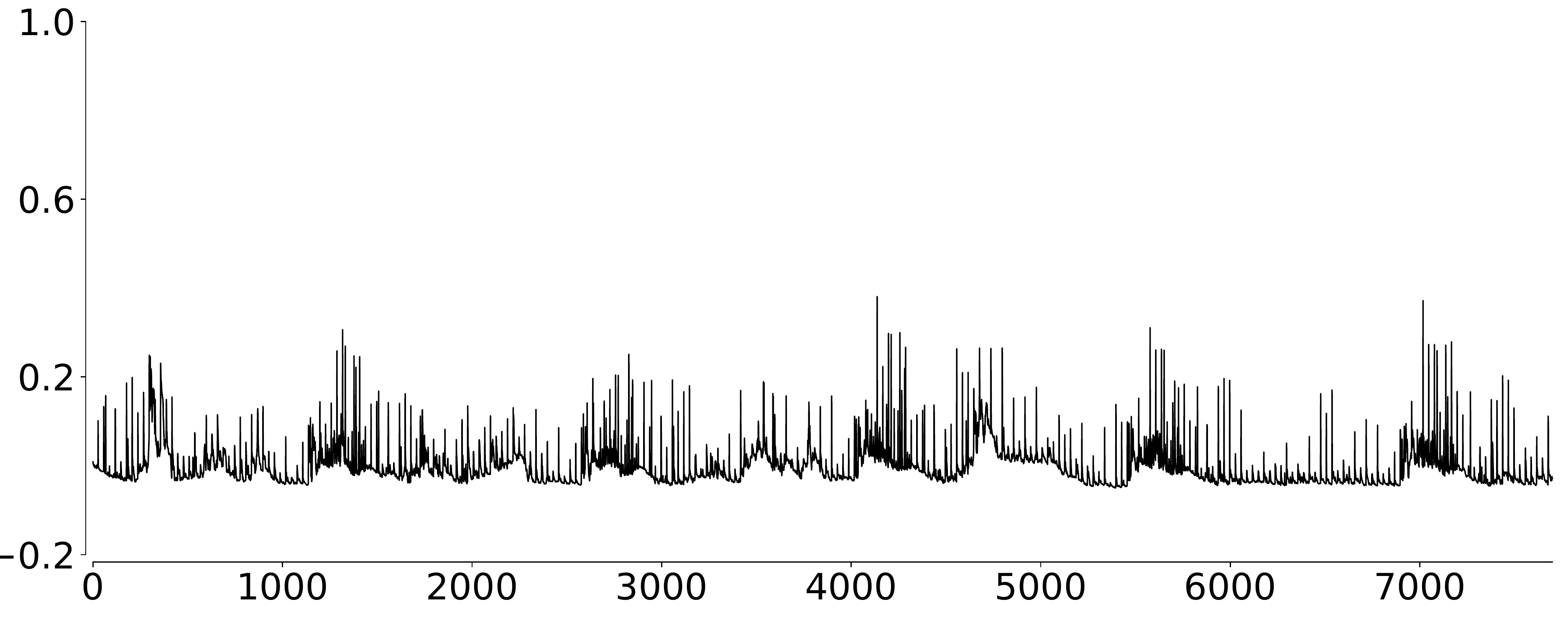}
		\vspace{-5mm}
	\end{subfigure}
	\begin{subfigure}[b]{0.235\textwidth}
		\includegraphics[width=\textwidth]{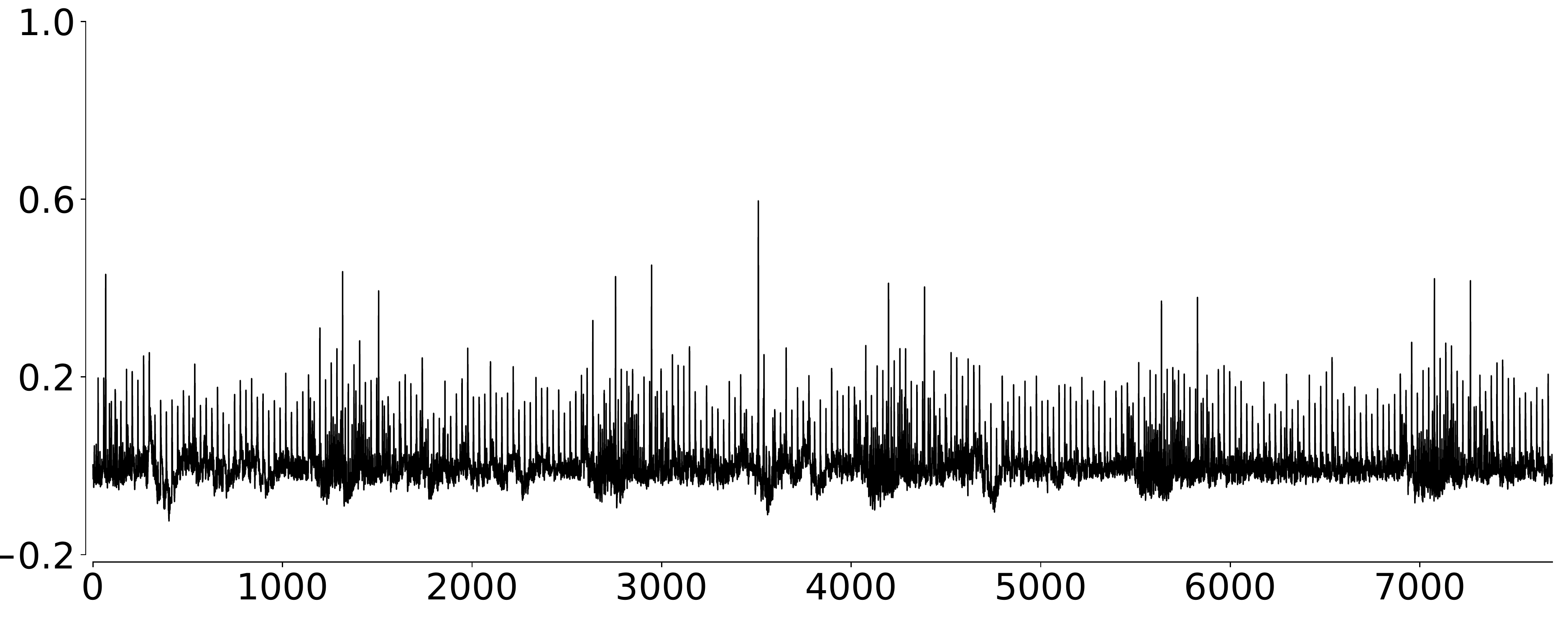}
		\vspace{-5mm}
	\end{subfigure}
	\begin{subfigure}[b]{0.235\textwidth}
		\includegraphics[width=\textwidth]{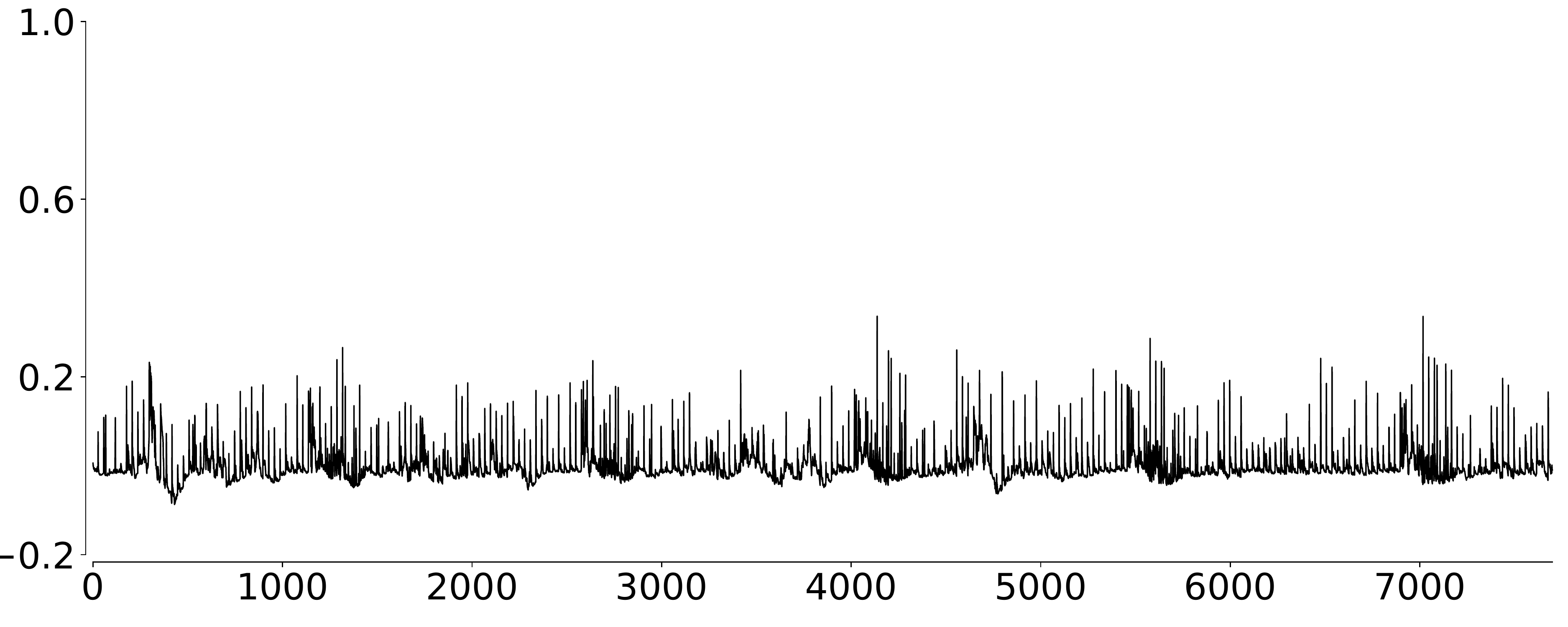}
		\vspace{-5mm}
	\end{subfigure}
	\begin{subfigure}[b]{0.235\textwidth}
		\includegraphics[width=\textwidth]{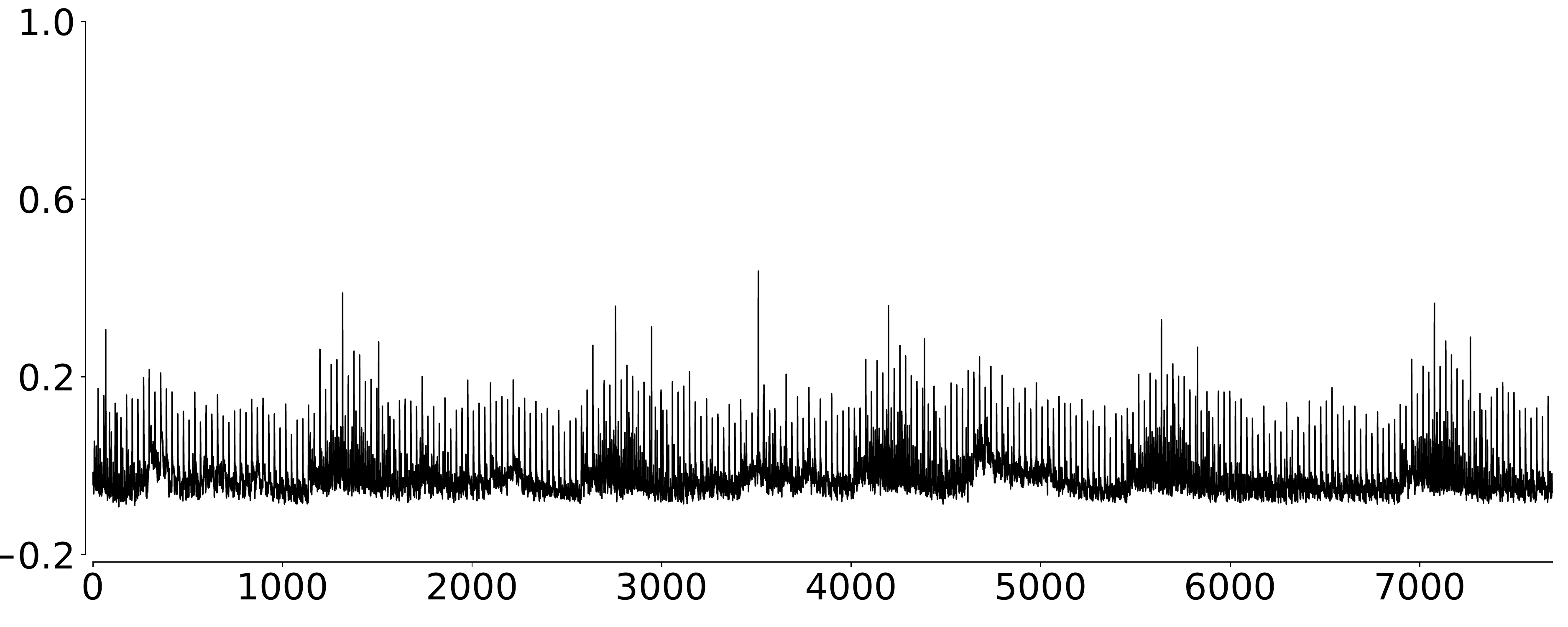}
		\vspace{-5mm}
	\end{subfigure}
	
	\begin{subfigure}[b]{0.235\textwidth}
		\includegraphics[width=\textwidth]{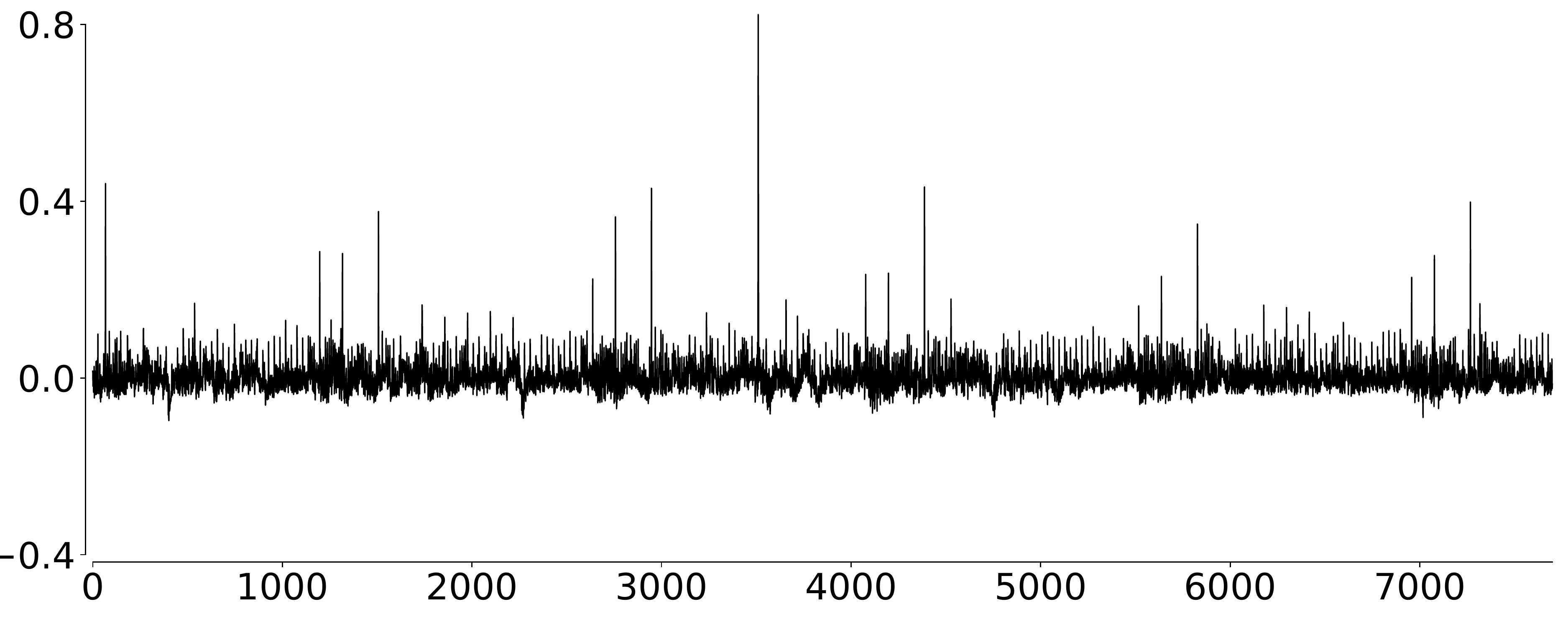}
		\vspace{-5mm}
	\end{subfigure}
	\begin{subfigure}[b]{0.235\textwidth}
		\includegraphics[width=\textwidth]{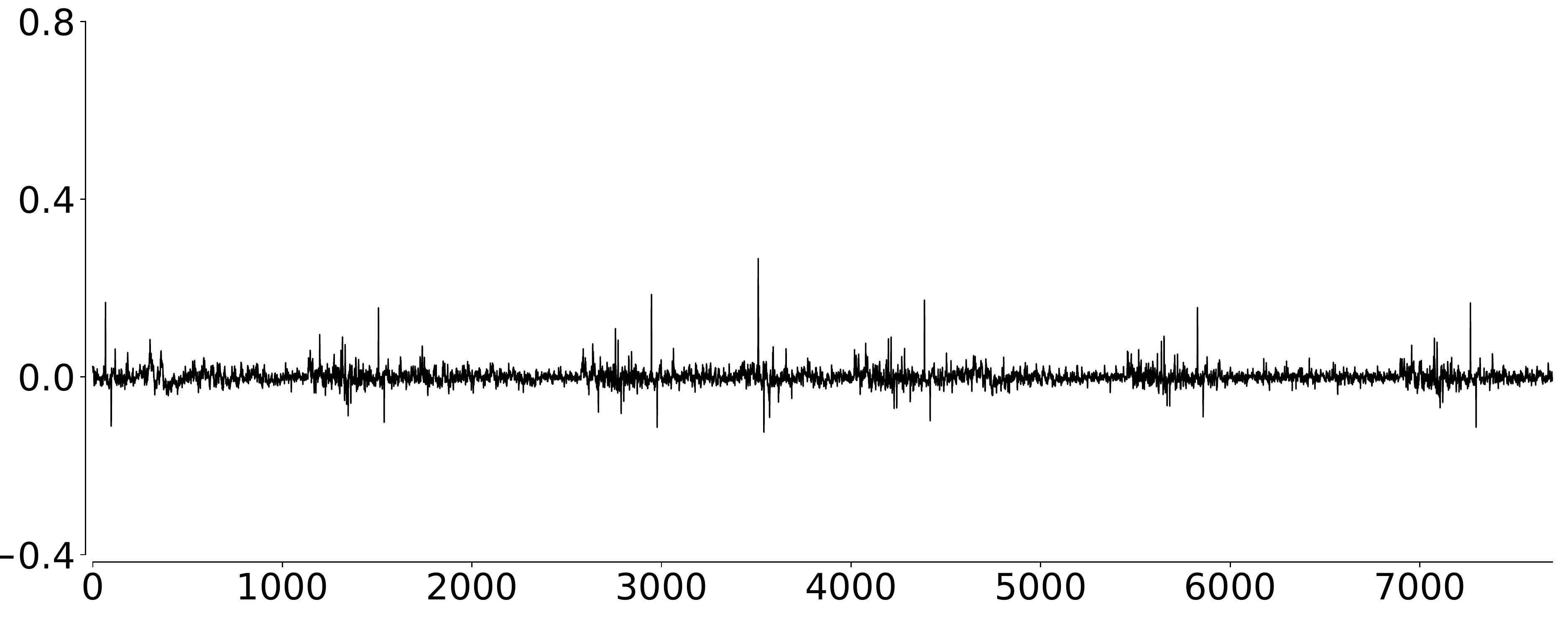}
		\vspace{-5mm}
	\end{subfigure}
	\begin{subfigure}[b]{0.235\textwidth}
		\includegraphics[width=\textwidth]{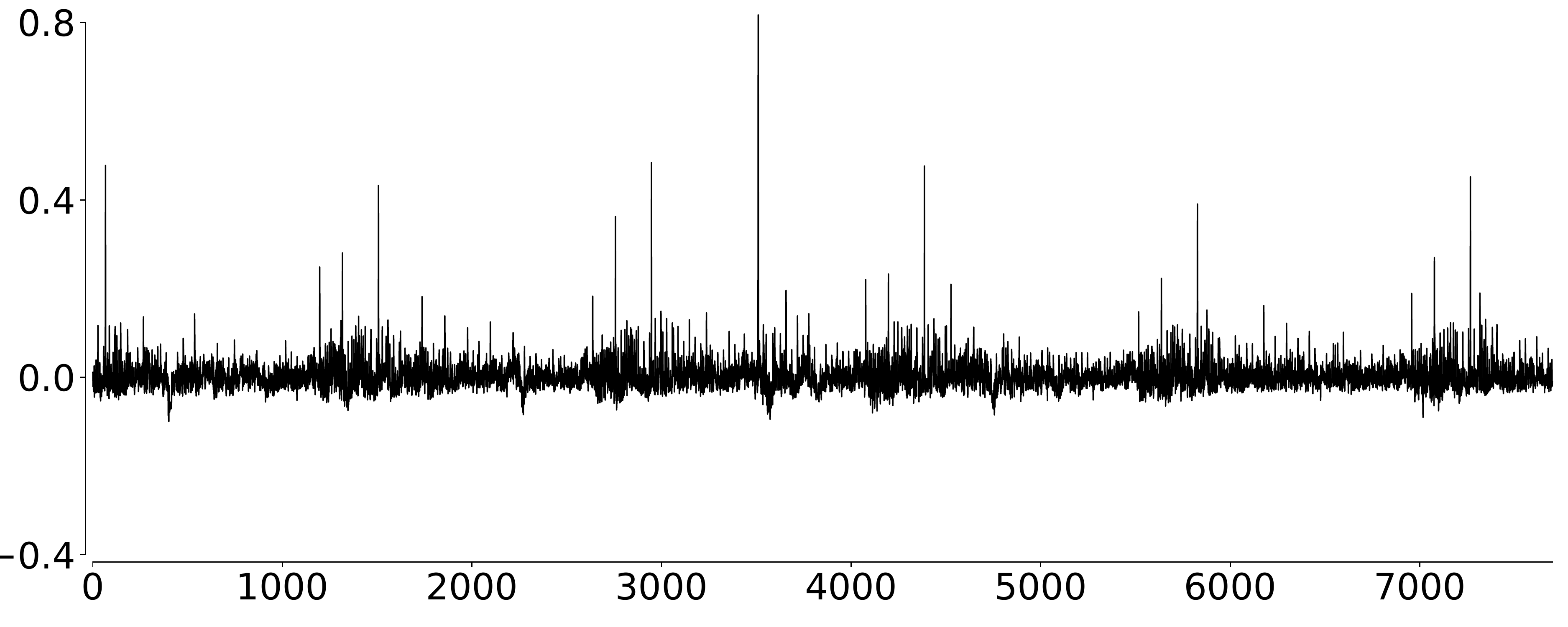}	
		\vspace{-5mm}
	\end{subfigure}
	\begin{subfigure}[b]{0.235\textwidth}
		\includegraphics[width=\textwidth]{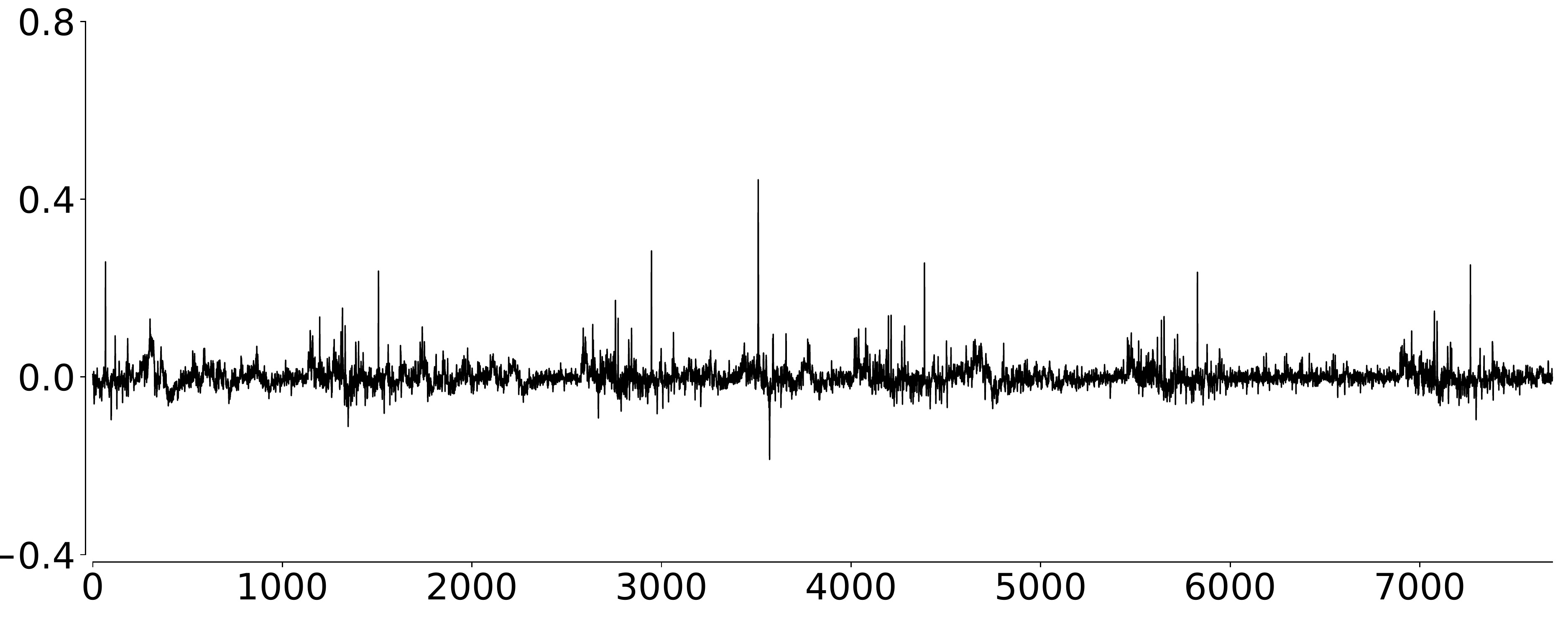}	
		\vspace{-5mm}
	\end{subfigure}
	\vspace{-2mm}
	\caption{Visualization of decomposition results on Real1 ((a)-(d)) and Real2 ((e)-(h)) datasets.}
	\vspace{-2mm}	
	\label{fig:real2_visual}
\end{figure*}

\subsection{How accurate is OneShotSTL?}
We first compare the quality of the decomposition from OneShotSTL with the previous STD methods. Figure \ref{fig:synthetic2_visual} depicts the decomposition results on \textbf{Syn1} (from (a) to (d)) and \textbf{Syn2} (from (e) to (h)) respectively. As can be seen from sub-figures (a) to (d), OnlineSTL and OnlineRobustSTL output too smooth trend on \textbf{Syn1} and fail to recover the ground truth signal of trend, while OneShotSTL successfully handles such cases. Further, OnlineSTL fails to handle seasonality shift on \textbf{Syn2} and outputs significantly different trend and residual signals. In comparison, OneShotSTL can deal with such cases perfectly as the batch method RobustSTL. Moreover, Table \ref{tab:result_syn} depicts the MAE between the ground truth signals and the decomposed signals from both batch and online methods. As can be seen, RobustSTL and OneShotSTL (shown in bold) are the best-performing batch and online methods respectively on both synthetic datasets. Their MAE scores are quite close, which are significantly better than all the other comparisons. 

\begin{table}[t]
	\caption{Decomposition comparison on synthetic datasets}
	\vspace{-4mm}
	\begin{center}
		\small
		\begin{tabular}{|l|l|l|ccc|}
			\hline
			\multirow{2}{*}{Data} & \multirow{2}{*}{Type} & \multirow{2}{*}{Algorithm} & Trend & Seasonal & Residual\\
			& & & (MAE) & (MAE) & (MAE))\\
			\hline
			\multirow{7}{*}{Syn1}			
			& \multirow{2}{*}{Batch}
			&STL&0.134&0.015&0.144\\
			& &\textbf{RobustSTL}&\textbf{0.004}&\textbf{0.013}&\textbf{0.016}\\
			\cline{2-6}
			& \multirow{5}{*}{Online}
			&Window-STL&0.134&0.092&0.174\\
			& &OnlineSTL&0.104&0.023&0.093\\
			& &Window-RobustSTL&0.045&0.018&0.046\\			
			& &OnlineRobustSTL&0.131&0.033&0.123\\			
			& &\textbf{OneShotSTL}&\textbf{0.007}&\textbf{0.014}&\textbf{0.019}\\				
			\hline		
			\multirow{7}{*}{Syn2}				
			& \multirow{2}{*}{Batch}
			&STL&0.084&0.433&0.505\\
			& &\textbf{RobustSTL}&\textbf{0.004}&\textbf{0.004}&\textbf{0.004}\\
			\cline{2-6}
			& \multirow{5}{*}{Online}
			&Window-STL&0.084&0.313&0.313\\
			& &OnlineSTL&0.225&0.374&0.571\\
			& &Window-RobustSTL&0.032&0.031&\textbf{0.006}\\
			& &OnlineRobustSTL&0.037&0.031&0.013\\
			& &\textbf{OneShotSTL}&\textbf{0.004}&\textbf{0.013}&0.013\\
			\hline			
		\end{tabular}
	\end{center}
	\vspace{-4mm}
	\label{tab:result_syn}
\end{table}

Next, we visualize the decomposed signals of real-world datasets \textbf{Real1} and \textbf{Real2} in Figure \ref{fig:real2_visual} (a) to (d) and (e) to (h) respectively. As can be seen from the sub-figures (a) to (d) of Figure \ref{fig:real2_visual}, the decomposed trend signal of \textbf{Real1} from OneShotSTL is closer to that from RobustSTL, which can better discover the abrupt change of the trend than OnlineSTL and OnlineRobustSTL. Further, from the results of \textbf{Real2} in sub-figures (e) to (h) of Figure \ref{fig:real2_visual} we can see that there are strong variations of the decomposed trend signals from OnlineSTL and OnlineRobustSTL. These may increase the possibility of false alarms for anomaly detection on the decomposed trend signal. In comparison, OneShotSTL can handle it better.

\begin{table*}
	\caption{Univariate TSAD Results (VUS-ROC) on sixteen datasets from TSB-UAD benchmark. A higher VUS-ROC indicates better performance. The ranking of the method is shown in the bracket and the best results are shown in bold.}	
	\vspace{-2mm}
	\label{tab:tsad}	
	\begin{center}
		\small		
		\begin{tabular}{|c|c|c|c|c|c|c|c|c|c|c|c|c|c|c|c|c|c|c|c|c|c|c|}
			\hline 	
			Method & LSTM & USAD & TranAD & NormA & SAND & STOMPI & DAMP & NSigma & OnlineSTL & OneShotSTL\\
			\hline Daphnet  
			& 0.476 (7) & 0.653 (3) & 0.548 (6) & 0.462 (10) & 0.594 (5) & 0.465 (7) & 0.475 (8) & 0.605 (4) & 0.696 (2) & \textbf{0.735 (1)}\\					
			\hline Dodgers  
			& 0.743 (9) & \textbf{0.872 (1)} & 0.768 (7) & 0.815 (4) & 0.788 (6) & 0.794 (5) & 0.630 (10) & 0.825 (2) & 0.756 (8) & 0.819 (3)\\			
			\hline ECG  
			& 0.745 (9) & \textbf{0.936 (1)} & 0.759 (7) & 0.935 (2) & 0.884 (4) & 0.891 (3) & 0.864 (5) & 0.715 (10) & 0.754 (8) & 0.767 (6)\\			
			\hline Genesis  
			& 0.507 (10) & \textbf{0.827 (1)} & 0.742 (4) & 0.669 (5) & 0.507 (9) & 0.530 (8) & 0.644 (7) & 0.823 (2) & 0.666 (6) & 0.762 (3)\\					
			\hline GHL  
			& 0.627 (8) & \textbf{0.913 (1)} &  0.831 (4) & 0.759 (5) & 0.727 (7) & 0.603 (9) & 0.553 (10) & 0.732 (6) & 0.866 (3) & 0.872 (2)\\					
			\hline IOPS
			& 0.743 (3) & 0.740 (5) & 0.690 (6) & 0.503 (9) & 0.487 (10) & 0.550 (8) & 0.616 (7) & \textbf{0.774 (1)} & 0.741 (4) & 0.749 (2) \\		
			\hline	MGAB  
			& 0.683 (2) & 0.628 (10) & 0.647 (4) & 0.628 (9) & 0.637 (5) & 0.628 (8) & \textbf{0.731 (1)} & 0.665 (3) & 0.634 (6) & 0.628 (7) \\		
			\hline MITDB  
			& 0.613 (10) & 0.745 (5)  & 0.667 (8) & \textbf{0.824 (1)} & 0.792 (2) & 0.746 (4) & 0.773 (3) & 0.671 (7) & 0.664 (9) & 0.685 (6) \\		
			\hline NAB  
			& 0.510 (10) & \textbf{0.631 (1}) & 0.551 (6) & 0.580 (4) & 0.519 (9) & 0.520 (8) & 0.540 (7) & 0.570 (5) & 0.584 (3) & 0.590 (2)\\		
			\hline NASA-MSL  
			& 0.613 (8) & \textbf{0.699 (1)} & 0.650 (6) & 0.645 (7) & 0.695 (3) & 0.606 (10) & 0.609 (9) & 0.694 (5) & 0.696 (2) & 0.688 (5)\\		
			\hline NASA-SMAP  
			& 0.642 (8) & 0.711 (4) & 0.551 (10) & \textbf{0.818 (1)} & 0.800 (2) & 0.711 (3) & 0.665 (6) & 0.642 (7) & 0.601 (9) &  0.666 (5)\\		
			\hline Occupancy  
			& 0.532 (6) & 0.421 (7) & 0.845 (2) & 0.626 (5) & 0.254 (10) & 0.399 (9) & 0.408 (8) & 0.770 (4) & \textbf{0.855 (1)} & 0.840 (3) \\		
			\hline Opportunity
			& 0.564 (6) & 0.481 (7) & 0.454 (8) & \textbf{0.808 (1)} & 0.777 (2) & 0.767 (3) & 0.707 (4) & 0.586 (5) & 0.418 (9) & 0.404 (10) \\					
			\hline SensorScope  
			& 0.546 (10) & 0.555 (8) & 0.573 (7) & 0.632 (4) & 0.551 (9) & 0.587 (6) & 0.609 (5) & 0.668 (2) & 0.646 (3) & \textbf{0.672 (1)} \\		
			\hline SMD  
			& 0.606 (6) & 0.697 (4) & 0.699 (3) &0.612 (9) & 0.580 (10) & 0.622 (8) & 0.635 (7) &0.680 (5) & 0.732 (2) & \textbf{0.749 (1)}\\
			\hline SVDB  
			& 0.650 (7) & 0.731 (5) & 0.647 (10) & \textbf{0.924 (1)} & 0.896 (2) & 0.830 (3) & 0.815 (4) & 0.649 (8) & 0.647 (9) & 0.690 (6) \\
			\hline YAHOO  
			& 0.768 (6) & 0.626 (9) & 0.677 (8) & \textbf{0.915 (1)} & 0.895 (2) & 0.544 (10) & 0.810 (4) & 0.761 (7) & 0.833 (3) & 0.796 (5) \\
			\hline \hline Avg. VUS-ROC  
			& 0.624 (10) & 0.698 (3) & 0.664 (7) & \textbf{0.713 (1)} & 0.669 (6)  & 0.634 (9) & 0.652 (8) & 0.695 (4) & 0.693 (5) & \textbf{0.713 (1)}\\
			\hline Avg. Rank  
			& 7.35 (10) & 4.29 (2) & 6.23 (8)  & 4.58 (3) & 5.70 (6) & 6.70 (9) & 6.17 (7) & 4.82 (4) & 5.11 (5) & \textbf{4.00 (1)} \\
			\hline Avg. Time (sec) 
			& 13995 s (9)  & 7465 s (8)  & 545 s (4) & 1164 s (5) & 23601 s (10) & 3689 s (6) & 5298 s (7) & \textbf{2 s (1)} & 61 s (2) & 160 s (3)\\
			\hline											
		\end{tabular}		
	\end{center}	
	\vspace{-4mm}		
\end{table*}

\subsection{How Fast is OneShotSTL?}
Next, we compare the efficiency of different STD methods. Remember that OneShotSTL is the only online method with $O(1)$ update complexity. All the other methods have an update complexity of $O(T)$. Specifically, we build a new synthetic dataset of $200,000$ points by repeating \textbf{Syn1}. Please note that the runtime of all the STD methods only depends on the size and period length $T$ of the input time series. Then we evaluate the decomposition methods with different input $T \in \{100,200,400,800,1600,3200,6400,12800\}$ and report the average latency of processing one data point. We evaluate this range of $T$ because, in the context of real-time metric monitoring scenarios in AIOps, there is usually a daily or weekly seasonality pattern. Then for minute data, the period length is $1,440$ and $10,080$ for the daily and weekly seasonality respectively.  

\begin{figure}[t]
	\includegraphics[width=0.4\textwidth]{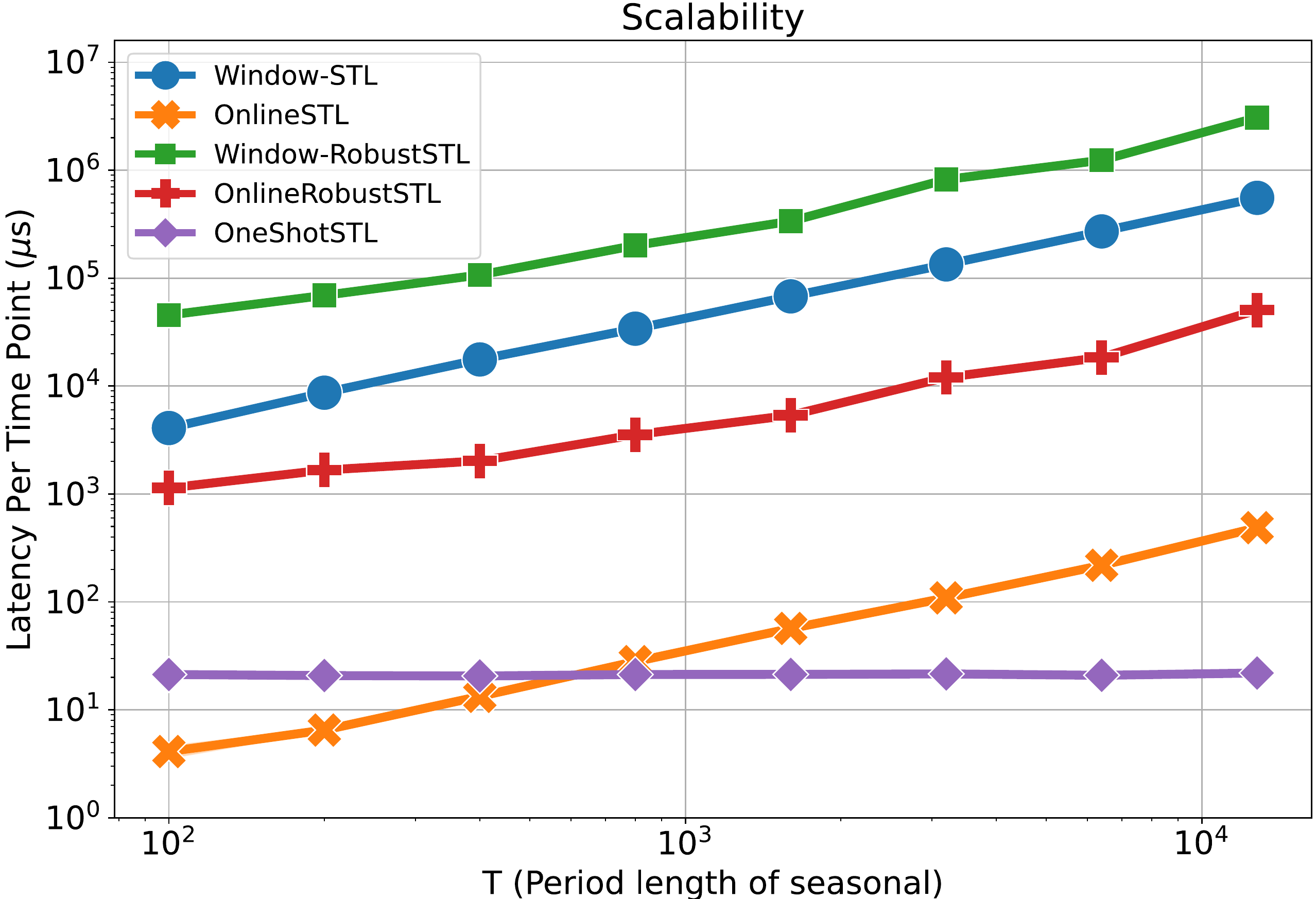}		
	\vspace{-4mm}
	\caption{Comparison of average latency per data point with different input of $T \in \{100, 200,400,800,1600,3200,6400,12800\}$.}
	\vspace{-4mm}	
	\label{fig:scalability}
\end{figure}

The results are shown in Figure \ref{fig:scalability}. All methods except OneShotSTL scale linearly with the increase of $T$. OneShotSTL is the only method with a constant latency (around $20 \mu s$) to process one data point, which agrees with the theoretical analysis in Section \ref{sec:model}. Approximately, it can process up to $50,000$ time series per second using a single CPU core. OnlineSTL performs better when $T < 800$, while OneShotSTL is faster when $T > 800$. When $T=12,800$, OnlineSTL needs about $450 \mu s$ to decompose one data point, which is more than $20$ times slower than OneShotSTL. The other three methods Window-STL, Window-RobustSTL, and OnlineRobustSTL are at least two orders of magnitude slower than OnlineSTL and OneShotSTL. 
Therefore, in the following evaluation of the TSAD and TSF tasks, we only include OnlineSTL and OneShotSTL.

\subsection{How do the methods compare in TSAD?}
In this subsection, we apply STD methods to the TSAD task and compare them with the corresponding state-of-the-art methods.

Firstly, we evaluate seventeen datasets from the TSB-UAD benchmark with results shown in Table \ref{tab:tsad}. The average VUS-ROC score from a dataset is displayed in each row of the table, along with the method's ranking, which is indicated in brackets. The average VUS-ROC, ranking, and execution time for all seventeen datasets are shown in the table's last three rows. 
The table shows that OneShotSTL surpasses majority of the comparison methods with an average VUS-ROC of $0.713$ and an average ranking of $4.00$.
The table also demonstrates that no particular approach consistently outperforms the others. Specifically, USAD, NormA and OneShotSTL perform the best on $6$, $5$, and $3$ datasets, respectively. Notably, we also observe that these three approaches are low-ranked algorithms on other datasets. OneShotSTL is higher in both the average ranking and the VUS-ROC score than the other methods because it performs consistently across various datasets.
Surprisingly, simple NSigma provides competitive results with an average VUS-ROC score of $0.695$, which outperforms many sophisticated methods like SAND ($0.669$) and DAMP ($0.652$). Furthermore, NSigma is also the fastest method with an average runtime of only $2$ seconds followed by OnlineSTL ($61$ seconds) and OneShotSTL ($160$ seconds). The other methods take thousands or even tens of thousands of seconds to complete.

\begin{table}[t]
	\caption{Univariate TSAD Results on KDD21 datasets}
	\vspace{-4mm}
	\begin{center}
		\small
		\begin{tabular}{|l|l|l|c|c|}
			\hline
			Data & Type & Method & Score & Time\\			
			\hline					
			\multirow{13}{*}{KDD21}
			& \multirow{3}{*}{Deep} 
			& LSTM & 0.460 & 1842 min\\
			& & USAD & 0.168 & 54 min\\			
			& & TranAD & 0.196 & 54 min \\
			\cline{2-5}
			& \multirow{4}{*}{MP}
			& NormA & \textbf{0.500} & 24 min \\
			& & STOMPI & 0.360 & 68 min \\
			& & SAND & 0.388 & 13534 min \\						
			& & DAMP & \textbf{0.512} & 384 min \\			
			\cline{2-5}
			& \multirow{3}{*}{STD}
			&NSigma & 0.132 & < 0.1 min \\			
			& &OnlineSTL & 0.268 & 4.3 min \\
			& &OneShotSTL & 0.288 & 7.5 min  \\					
			\cline{2-5}
			& \multirow{3}{*}{STD+MP}			
			&NSigma+DAMP& 0.324 & 4.7 min \\			
			& &OnlineSTL+DAMP & 0.408 & 7.2 min \\
			& &OneShotSTL+DAMP& \textbf{0.508} & 9.5 min  \\						
			\hline				
		\end{tabular}
	\end{center}
	\vspace{-4mm}
	\label{tab:KDD21}
\end{table}

\begin{table*}
	\caption{Univariate TSF results (MAE) on six datasets with forecasting lengths $\{96, 192, 336, 720\}$ ($\{24, 36, 48, 60\}$ for Illness). The ranking of the method is shown in the bracket and the best results are highlighted in bold.}
	\vspace{-2mm}
	\label{tab:tsf}
	\begin{center}
		\small
		\begin{tabular}{|c|c|c|c|c|c|c|c|c|c|c|c|c|c|c|c|c|c|c|c|c|c|}
			\hline 	
			\multicolumn{2}{|c|}{Method} & FiLM &  FEDformer & Informer & NBEATS & DeepAR  & AutoArima & OnlineSTL & OneShotSTL \\			
			\hline					
			\multirow{4}{*}{ETTm2} 
			& 96   & \textbf{0.189 (1)} & \textbf{0.189 (1)} & 0.225 (5)  & 0.201 (3)  & 0.942 (8) & 0.379 (7) & 0.375 (6) & 0.211 (4)\\
			& 192 & \textbf{0.233 (1)} & 0.245 (3) & 0.283 (5) & 0.248 (4) & 0.993 (8) & 0.419 (7) & 0.391 (6) & 0.244 (2)\\		
			& 336 & 0.274 (2) & 0.279 (4) & 0.336 (5) & 0.274 (2) & 0.442 (8) & 0.437 (7) & 0.419 (6) & \textbf{0.273 (1)} \\		
			& 720 & 0.323 (2) & 0.325 (3) & 0.435 (4) & 0.468 (6) & 1.028 (8) & 0.475 (7) & 0.452 (5) & \textbf{0.321 (1)} \\		
			\hline			
			\multirow{4}{*}{Electricity} 
			& 96   & \textbf{0.247 (1) }& 0.370 (3) & 0.538 (5) & 0.387 (4) & 0.641 (6) & 0.898 (7) & 1.031 (8) & 0.331 (2)\\
			& 192 & \textbf{0.258 (1) }& 0.386 (3) & 0.558 (5) & 0.458 (4) & 0.698 (6) & 0.925 (7) & 1.013 (8) & 0.355 (2)\\		
			& 336 & \textbf{0.283 (1) }& 0.431 (3) & 0.613 (5) & 0.494 (4) & 0.747 (6) & 1.013 (7) & 1.037 (8) & 0.389 (2)\\		
			& 720 & \textbf{0.341 (1)} & 0.484 (4) & 0.682 (5) & 0.483 (3) & 0.796 (6) & 1.096 (8) & 1.070 (7) & 0.444 (2)\\		
			\hline			
			\multirow{4}{*}{Exchange} 
			& 96   & 0.259 (4) & 0.284 (5) & 0.615 (8) & 0.358 (6) & 0.597 (7) & \textbf{0.241 (1)} & \textbf{0.241 (1)} & 0.252 (3) \\
			& 192 & 0.352 (4) & 0.420 (7) & 0.912 (8) & 0.387 (5) & 0.419 (6) & \textbf{0.321 (1)} & 0.327 (2) & 0.334 (3) \\		
			& 336 & 0.461 (4) & 0.511 (5) & 0.984 (8) & 0.652 (6) & 0.920 (7) & \textbf{0.432 (1)} & 0.433 (2)& 0.436 (3)\\		
			& 720 & 0.708 (4) & 0.832 (6) & 1.072 (7) & 1.114 (8) & 0.710 (5) & 0.681 (3) & 0.626 (2) & \textbf{0.624 (1)}\\		
			\hline			
			\multirow{4}{*}{Traffic} 
			& 96   & 0.215 (3) & 0.263 (4) & 0.353 (5) & 0.206 (2) & 0.614 (6) & 1.036 (7)  & 1.465 (8) & \textbf{0.181 (1)} \\
			& 192 & 0.199 (2) & 0.265 (4) & 0.376 (5) & 0.211 (3) & 0.992 (6) & 1.101 (7)  & 1.461 (8) & \textbf{0.181 (1)} \\		
			& 336 & 0.212 (3) & 0.266 (4) & 0.387 (5) & 0.209 (2) & 1.105 (6) & 1.138 (7)  & 1.470 (8) & \textbf{0.182 (1)} \\		
			& 720 & 0.252 (3) & 0.286 (4) & 0.436 (5) & 0.243 (2) & 1.166 (6) & 1.213 (7)  & 1.461 (8) & \textbf{0.199 (1)} \\		
			\hline			
			\multirow{4}{*}{Weather} 
			& 96   & 0.026 (4) & 0.046 (7) & 0.044 (6) & \textbf{0.022 (1)}& 0.059 (8) & 0.025 (3)  & 0.031 (5) & 0.024 (2)\\
			& 192 & 0.029 (4) & 0.059 (8) & 0.040 (6) & 0.026 (2) & 0.047 (7) & 0.028 (3)  & 0.032 (5) & \textbf{0.025 (1)}\\		
			& 336 & 0.030 (2) & 0.050 (8) & 0.049 (7) & 0.031 (3) & 0.035 (6) & 0.031 (3) & 0.033 (5) & \textbf{0.026 (1)}\\		
			& 720 & 0.037 (5) & 0.091 (8) & 0.042 (7) & 0.033 (2) & 0.033 (2) & 0.035 (4)  & 0.037 (6) & \textbf{0.030 (1)}\\						
			\hline
			\multirow{4}{*}{Illness} 
			& 24 & 0.538 (2) & 0.629 (3) & 2.050 (8) & \textbf{0.467 (1)} & 0.726 (5) & 0.821 (6)  & 0.917 (7) & 0.682 (4)\\
			& 36 & \textbf{0.481 (1)} & 0.604 (3) & 1.916 (8) & 0.519 (2) & 0.810 (5) & 0.928 (7)  & 0.840 (6) & 0.757 (4)\\		
			& 48 & \textbf{0.584 (1)} & 0.699 (3) & 1.846 (8) & 0.678 (2) & 0.861 (6) & 0.941 (7)  & 0.843 (5) & 0.773 (4) \\		
			& 60 & \textbf{0.644 (1)} & 0.828 (4) & 2.057 (8) & 0.773 (2) & 0.873 (5) & 0.913 (6) & 0.963 (7) & 0.826 (3) \\						
			\hline\hline
			\multicolumn{2}{|c|}{Avg. MAE } & \textbf{0.308 (1)} & 0.368 (3)  & 0.702 (7) & 0.373 (4) & 0.677 (6) & 0.647 (5) & 0.707 (8) & 0.337 (2)\\
			\hline			
			\multicolumn{2}{|c|}{Avg. Ranking} & 2.37 (2) & 4.45 (4) & 6.16 (7) & 3.29 (3) & 6.20 (8) & 5.41 (5) & 5.79 (6) & \textbf{2.08 (1)}\\						
			\hline					
			\multicolumn{2}{|c|}{Avg. Time (sec)} & 7860 s (7) & 2110 s (5) & 2003 s (4) & 790 s (3) & 4709 s (6) & 8779 s (8) & \textbf{0.3 s (1)}& \textbf{0.3 s (1)}\\						
			\hline								
		\end{tabular}			
	\end{center}
	\vspace{-4mm}
\end{table*}

Next, we evaluate KDD21 dataset. The average accuracy across $250$ time series and total time for training/testing are shown in Table \ref{tab:KDD21}. LSTM provides good accuracy of $0.460$, but its runtime ($1842$ minutes) is significantly longer than the other methods except SAND ($13534$ minutes).
The other two deep learning-based methods are faster, but they do not perform well on KDD21 datasets in terms of accuracy. Online TSAD method DAMP performs the best with an accuracy of $0.512$. However, DAMP takes $384$ minutes (around $6$ hours) to finish detecting all the datasets. 
Further, OnlineSTL and OneShotSTL can significantly improve the results of NSigma ($0.132$) to $0.268$ and $0.288$ respectively.
However, their accuracies are still significantly worse than DAMP. This is because STD methods can only handle seasonal patterns but the KDD21 dataset contains non-seasonal patterns. But OnlineSTL and OneShotSTL are much faster than DAMP.
Therefore, we further integrate them with DAMP by using them as a pre-filtering step. For each time series, we filter the top-ranked $1\%$ of testing points by the STD methods and give them to DAMP for detection. By doing this, OneShotSTL can reduce the computation time of DAMP from $384$ minutes to $9.5$ minutes (more than $40$ times faster) with negligible loss of accuracy (from $0.512$ to $0.508$). Meanwhile, we can observe that OneShotSTL outperforms OnlineSTL ($0.408$) while being used as filtering for DAMP.

Based on the above results, we conclude that OneShotSTL outperforms OnlineSTL for TSAD tasks by providing better detection accuracies in most cases. Moreover, OneShotSTL and matrix profile-based methods perform well on different datasets of the TSB-UAD benchmark, but OneShotSTL is from $10$ to $100$ times faster.

\begin{figure*}[t]	
	\vspace{-1mm}
	\begin{subfigure}[b]{0.245\textwidth}
		\includegraphics[width=\textwidth]{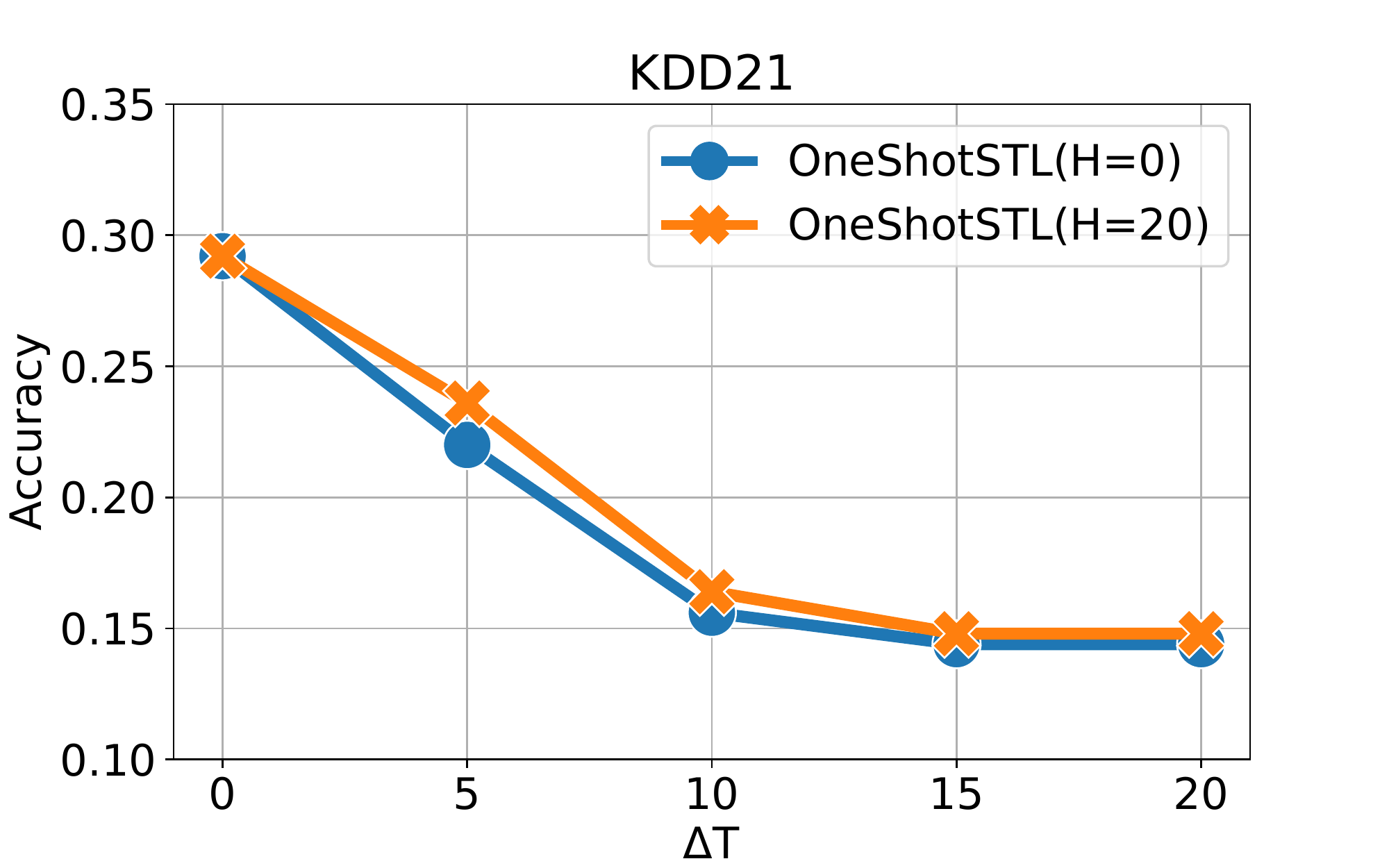}	
		\vspace{-5mm}
	\end{subfigure}	
	\begin{subfigure}[b]{0.245\textwidth}
		\includegraphics[width=\textwidth]{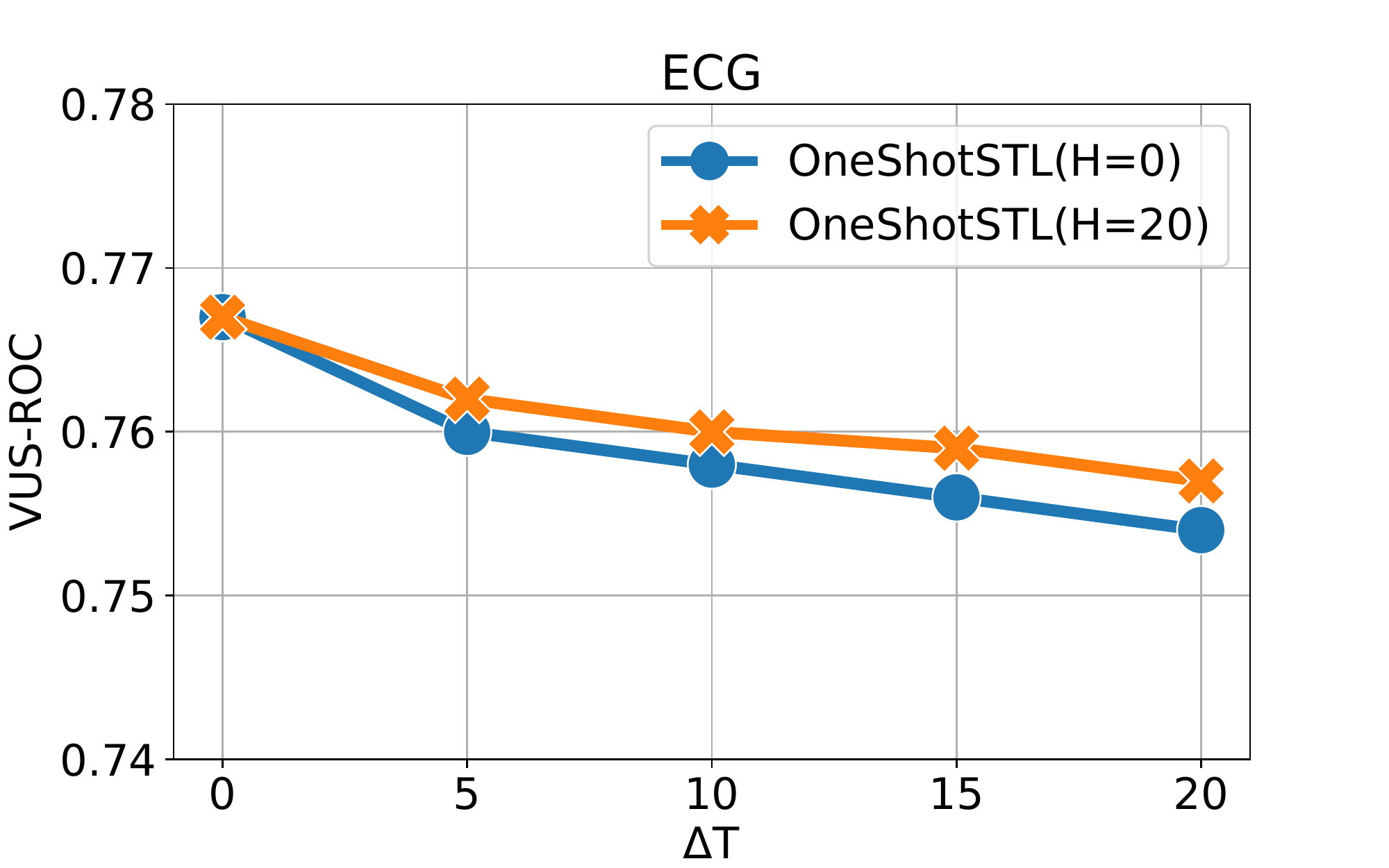}
		\vspace{-5mm}
	\end{subfigure}
	\begin{subfigure}[b]{0.245\textwidth}
		\includegraphics[width=\textwidth]{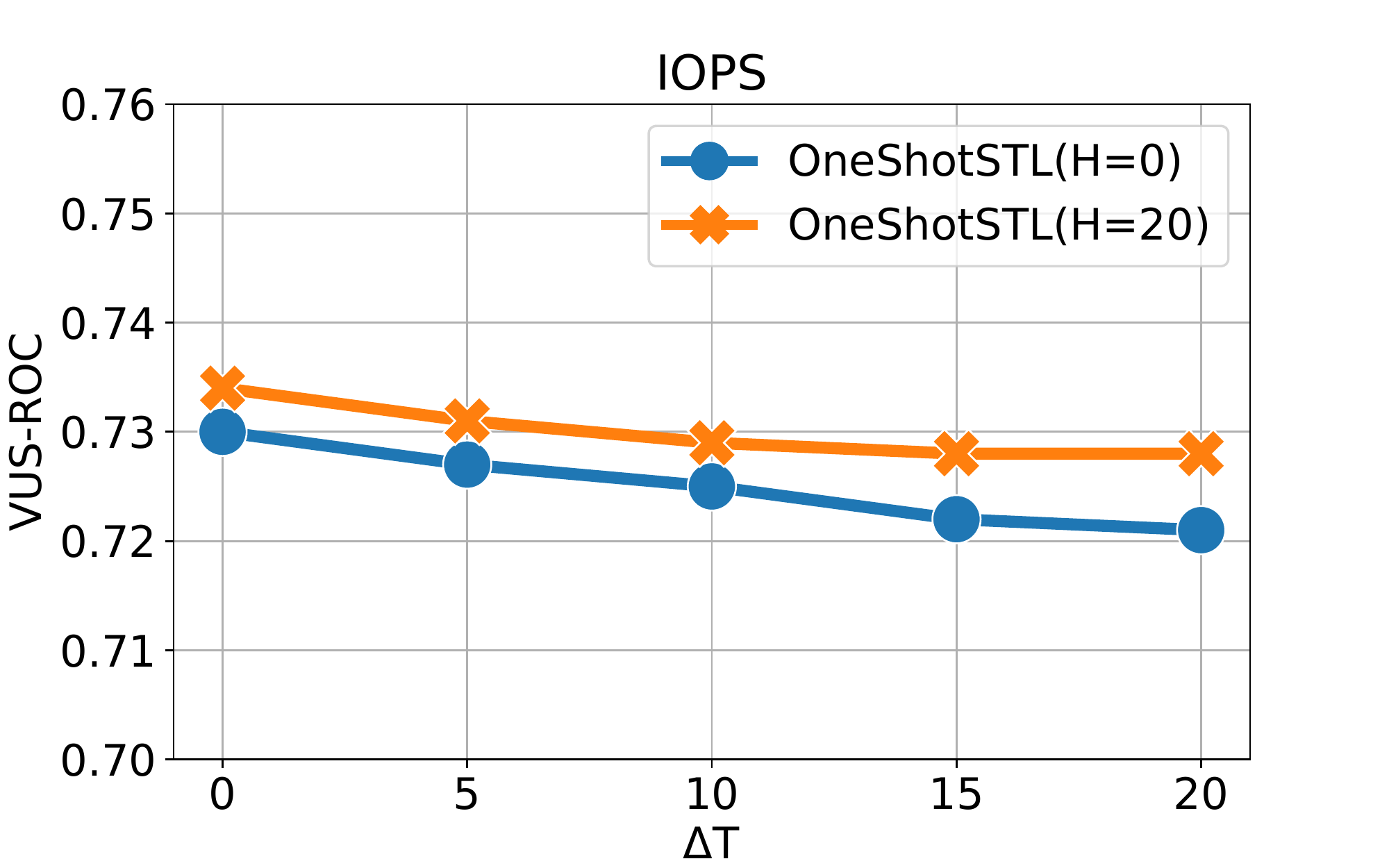}	
		\vspace{-5mm}
	\end{subfigure}	
	\begin{subfigure}[b]{0.245\textwidth}
		\includegraphics[width=\textwidth]{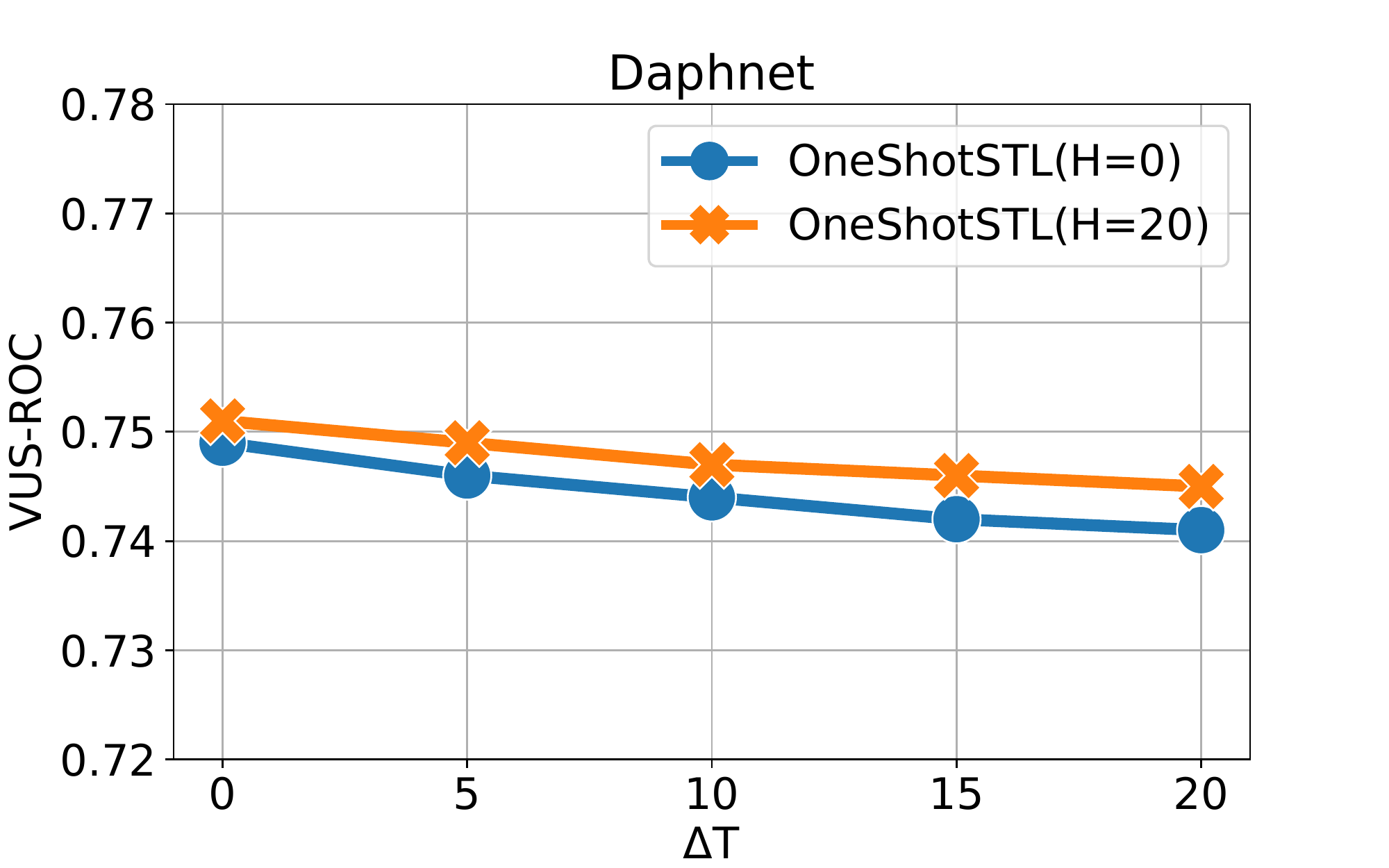}
		\vspace{-5mm}
	\end{subfigure}
	\vspace{-2mm}
	\caption{Ablation Study of hyper-parameters $T$ and $H$ of OneShotSTL on TSAD task. We add noise to $T$ by $\Delta T \in \{0, 5, 10, 15, 20\}$. Y-axis represents the accuracy and VUS-ROC scores (higher the better).}
	\vspace{-3mm}	
	\label{fig:ablation_TSAD_T}
\end{figure*}

\begin{figure*}[t]	
	\begin{subfigure}[b]{0.245\textwidth}
		\includegraphics[width=\textwidth]{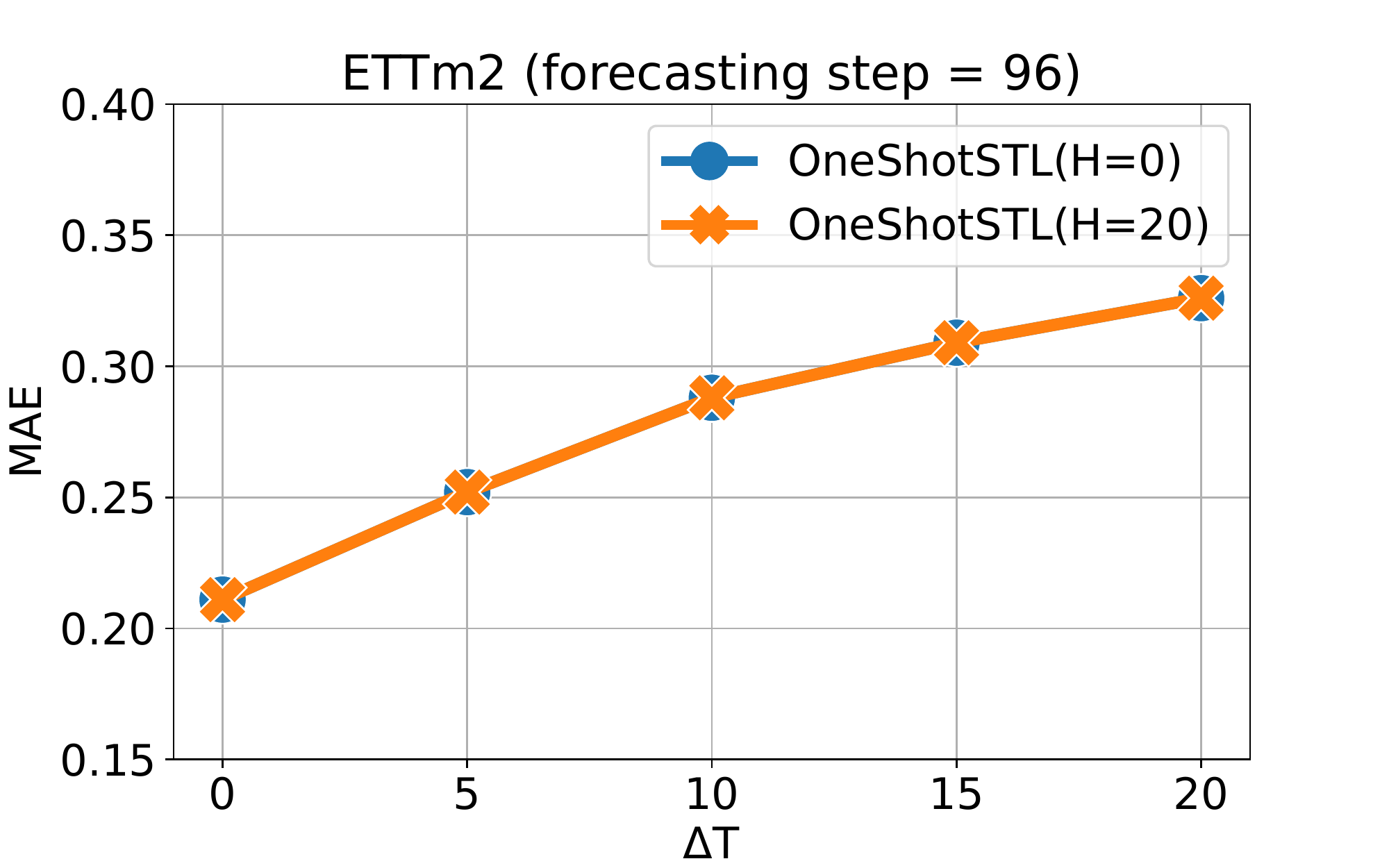}	
		\vspace{-5mm}
	\end{subfigure}	
	\begin{subfigure}[b]{0.245\textwidth}
		\includegraphics[width=\textwidth]{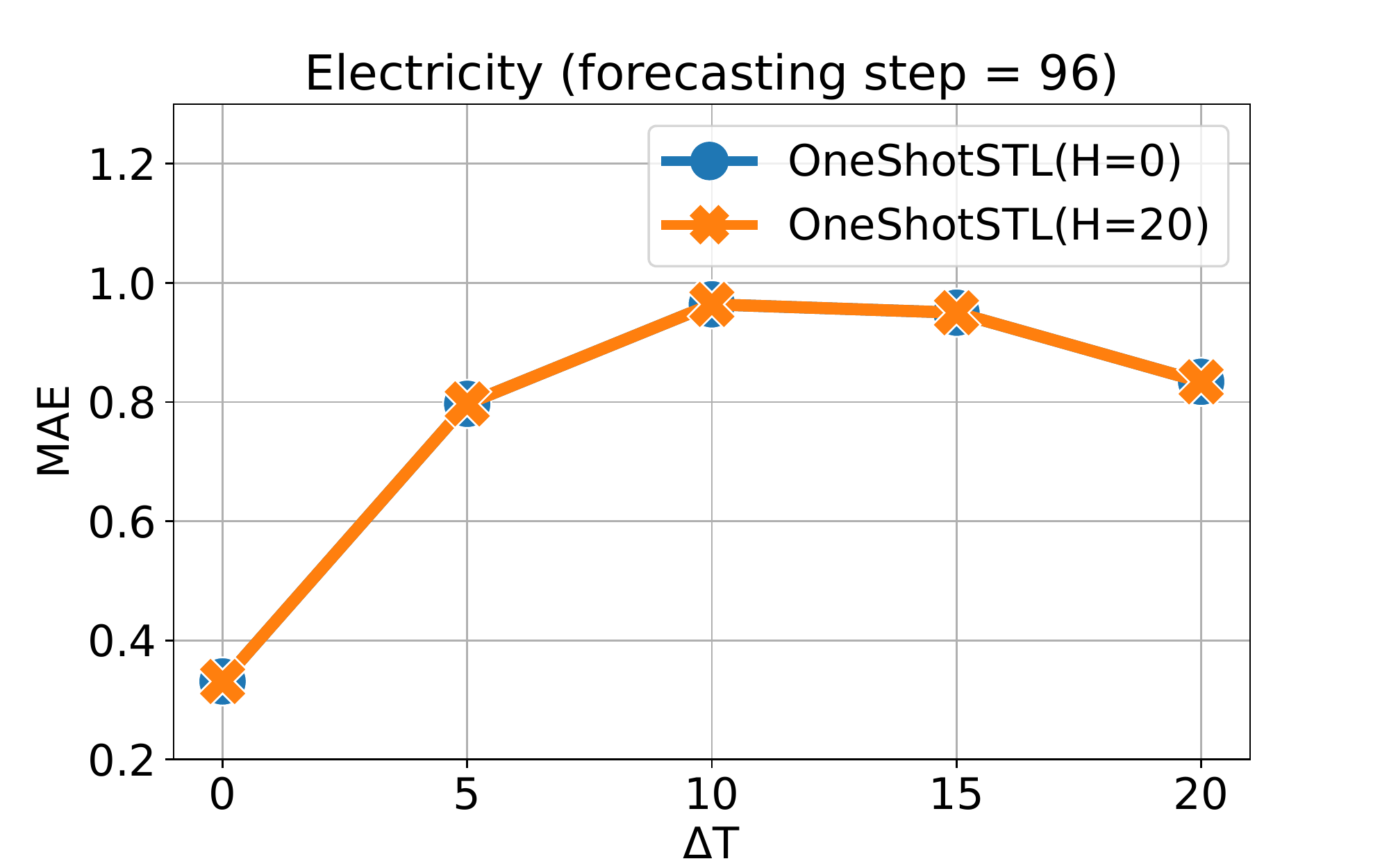}
		\vspace{-5mm}
	\end{subfigure}
	\begin{subfigure}[b]{0.245\textwidth}
		\includegraphics[width=\textwidth]{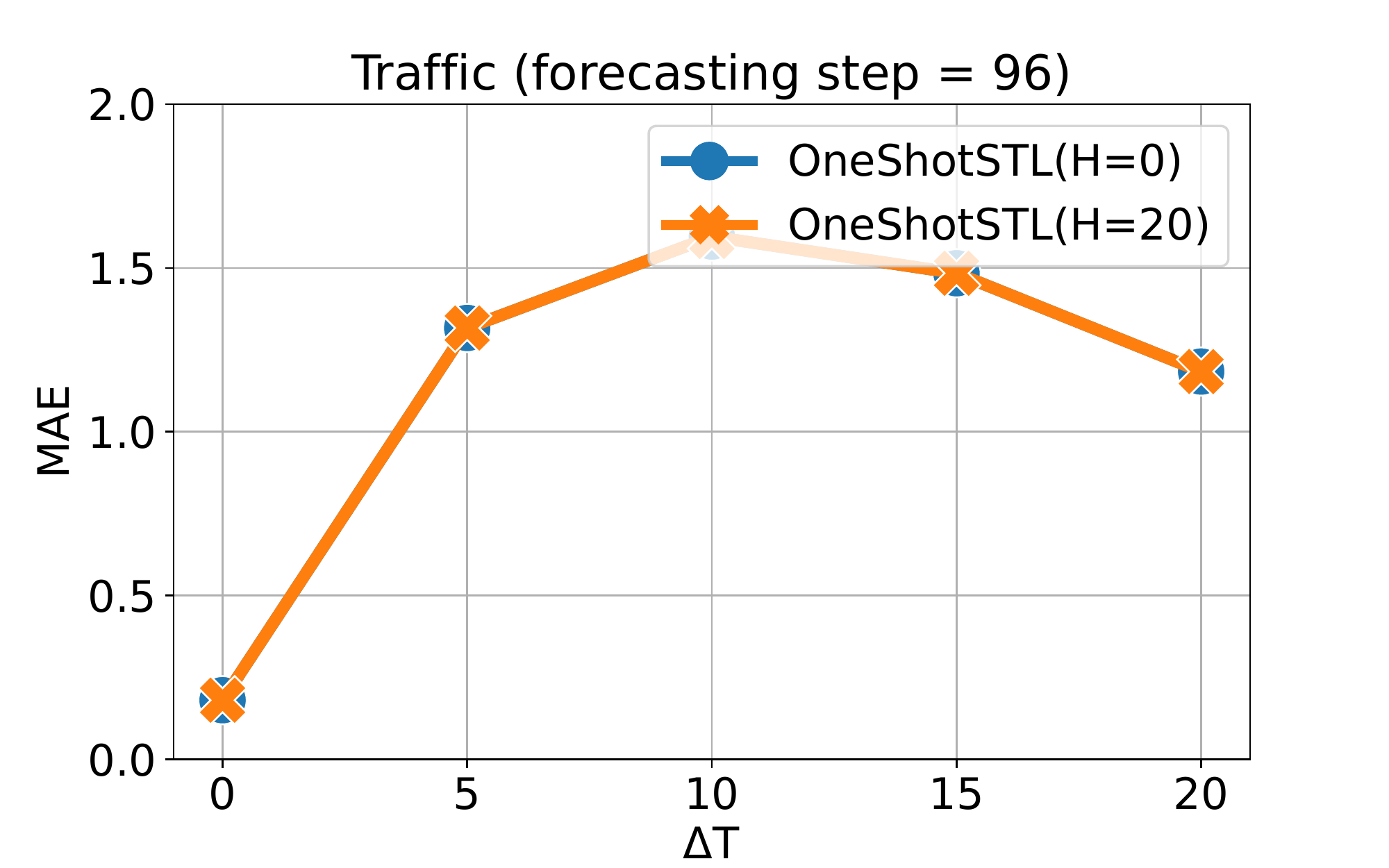}	
		\vspace{-5mm}
	\end{subfigure}	
	\begin{subfigure}[b]{0.245\textwidth}
		\includegraphics[width=\textwidth]{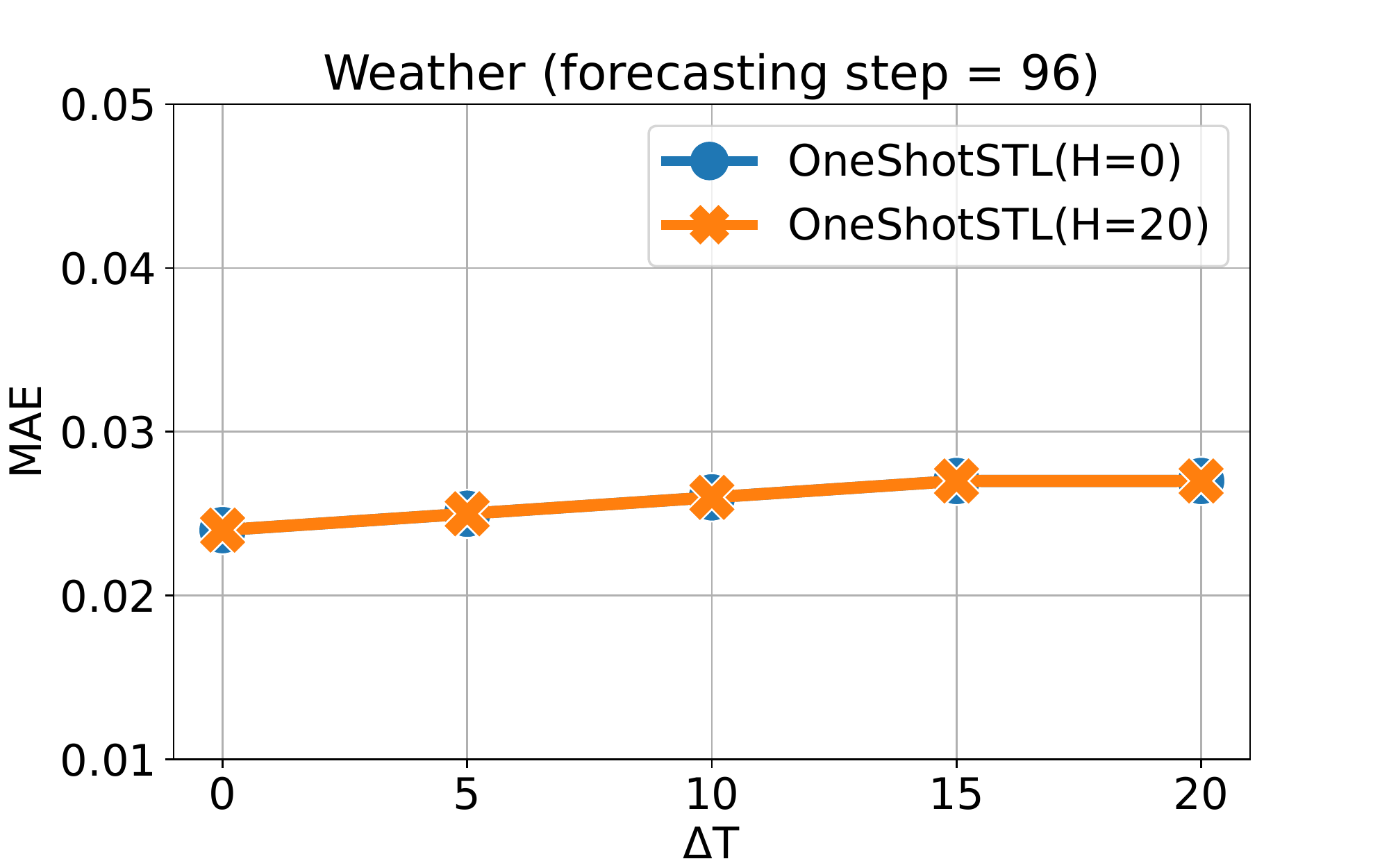}
		\vspace{-5mm}
	\end{subfigure}
	\vspace{-2mm}
	\caption{Ablation Study of hyper-parameters $T$ and $H$ of OneShotSTL on TSF task. We add noise to $T$ by $\Delta T \in \{0, 5, 10, 15, 20\}$. Y-axis represents the MAE (lower the better).}
	\vspace{-3mm}	
	\label{fig:ablation_TSF_T}
\end{figure*}

\begin{figure*}[t]	
	\begin{subfigure}[b]{0.245\textwidth}
		\includegraphics[width=\textwidth]{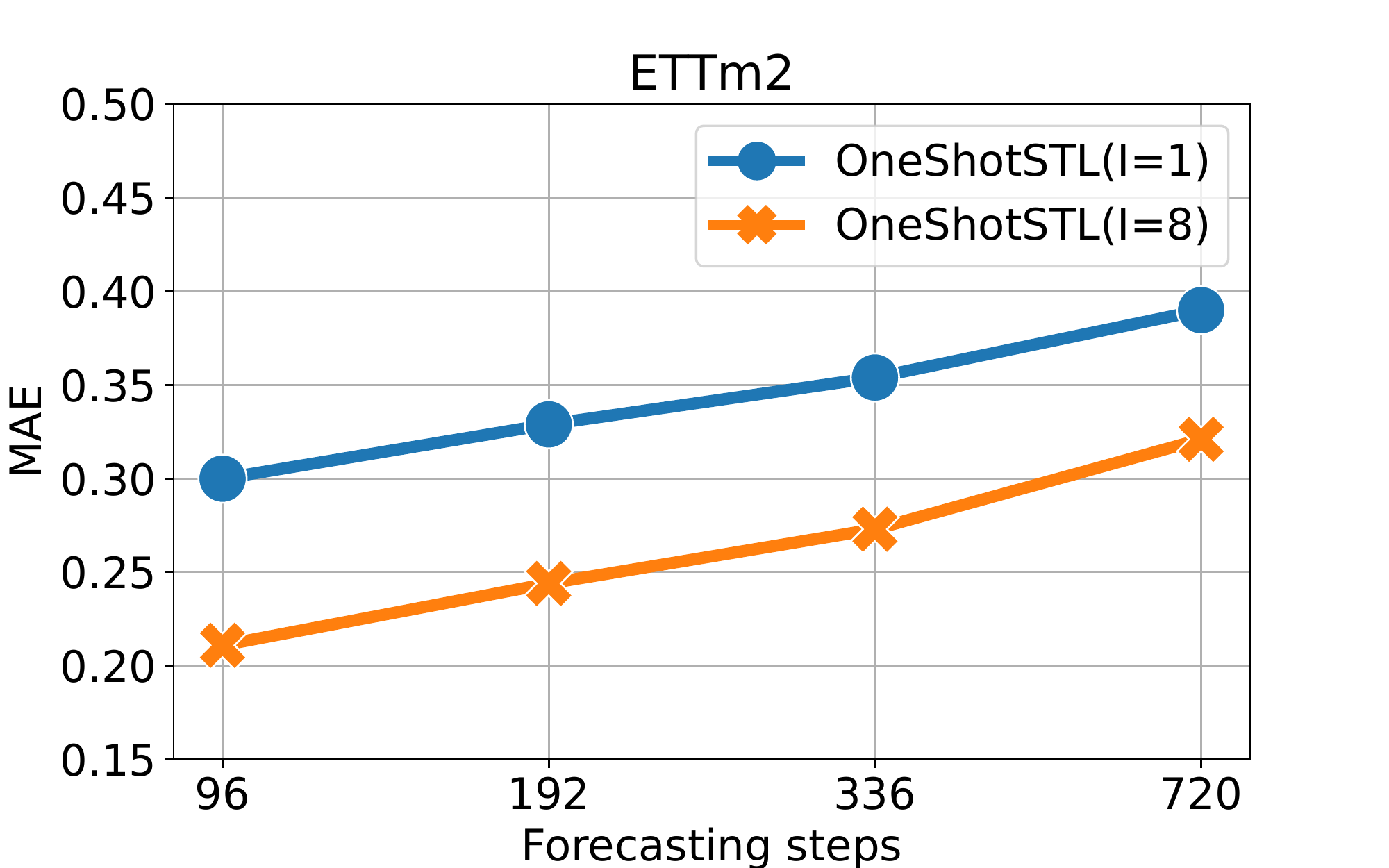}	
		\vspace{-5mm}
	\end{subfigure}	
	\begin{subfigure}[b]{0.245\textwidth}
		\includegraphics[width=\textwidth]{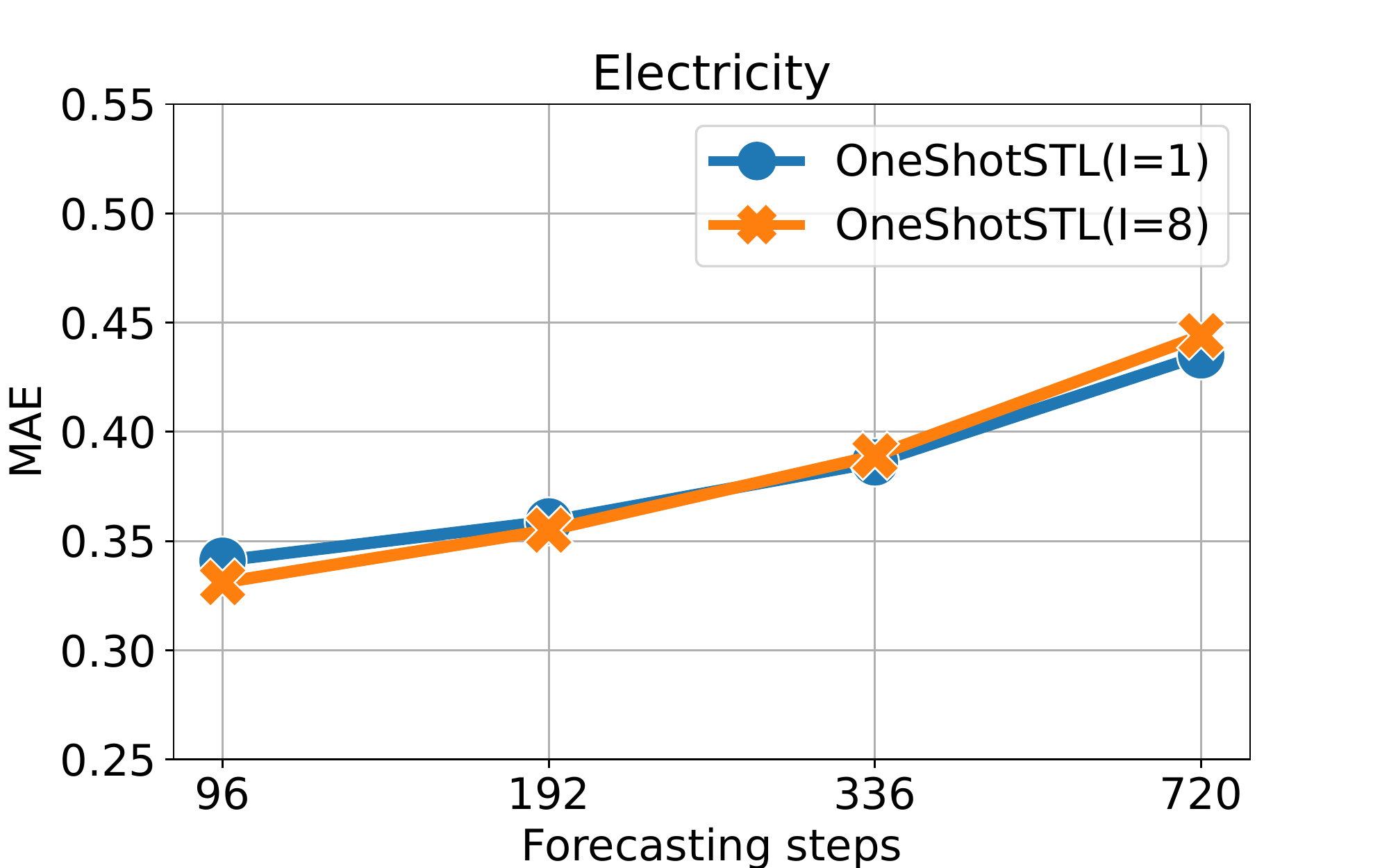}
		\vspace{-5mm}
	\end{subfigure}
	\begin{subfigure}[b]{0.245\textwidth}
		\includegraphics[width=\textwidth]{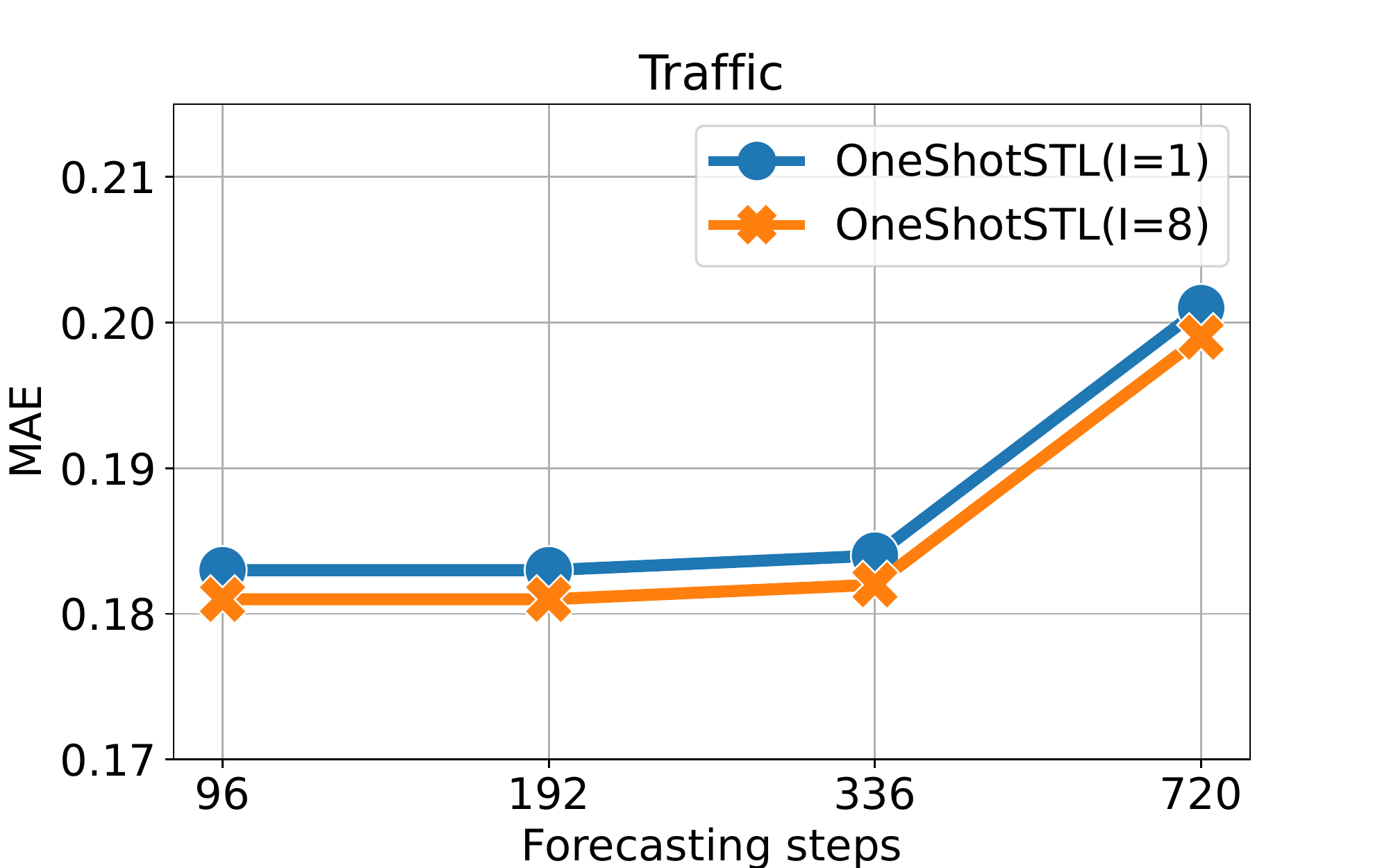}
		\vspace{-5mm}
	\end{subfigure}	
	\begin{subfigure}[b]{0.245\textwidth}
		\includegraphics[width=\textwidth]{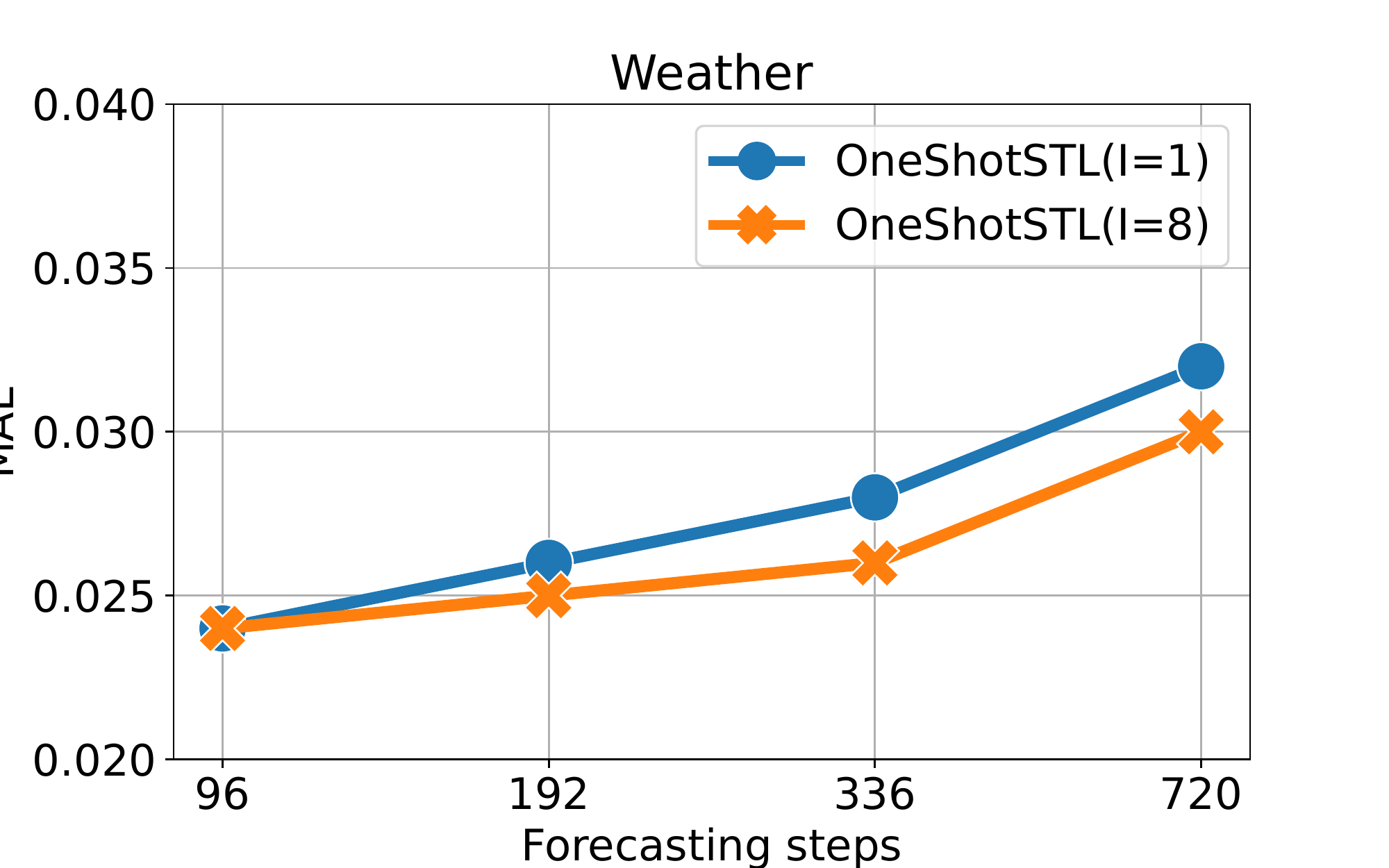}
		\vspace{-5mm}
	\end{subfigure}	
	\vspace{-2mm}
	\caption{Ablation Study of hyper-parameter $I$ of OneShotSTL on TSF task. Y-axis represents the MAE (lower the better).}
	\vspace{-3mm}	
	\label{fig:ablation_TSF}
\end{figure*}

\subsection{How do the methods compare in TSF?}
In this subsection, we apply STD methods to the TSF task and compare them with the corresponding state-of-the-art methods. 

Table \ref{tab:tsf} depicts the performance of $8$ comparison methods on $6$ benchmark datasets with forecasting length $\{96, 192, 336, 720\}$ ($\{24, 36, 48, 60\}$ for Illness). Each row of the table shows the prediction errors (measured by MAE) and the ranking of each method for each setting is shown in the bracket with the best results highlighted in bold. The last $3$ rows of the table represent the average MAE, ranking and execution time (including both training and testing time) across all the settings. First, we can see from the table that, regarding prediction error, FiLM and OneShotSTL are the top-two performers with average MAE of $0.308$ and $0.337$ respectively, followed by FEDformer ($0.368$) and NBEATS ($0.373$). In terms of average ranking, OneShotSTL is the best method with an average ranking of $2.08$, followed by FiLM ($2.37$), NBEATS ($3.29$), and FEDformer ($4.45$). Specifically, OneShotSTL performs better on datasets with strong seasonality (ETTm2, Electricity, Traffic, and Weather) and predicts relatively worse on Exchange and Illness with weak or no seasonality. 
Furthermore, STD methods have great advantages in speed over comparison TSF methods. The average runtime of OnlineSTL and OneShotSTL is less than $1$ second using a single CPU core on a commodity machine, while the processing time for deep learning-based methods is from hundreds to thousands of seconds using an NVIDIA V100 GPU (more than $1000$ times slower). 

In summary, the forecasting accuracies of OneShotSTL are comparable to the state-of-the-art deep learning method FiLM on the evaluated datasets. Meanwhile, its runtime is more than $1000$ times faster than the non-STD methods.

\subsection{Ablation Study of OneShotSTL}
In this subsection, we make ablation studies on three input parameters of OneShotSTL: periodicity length $T$, maximum seasonality shift length $H$, and number of iterations $I$.

Firstly, we  investigate how the errors in estimating $T$ can impact the final performance. Remember that $T$ is automatically discovered during the initialization phase of OneShotSTL. Thus, we use different values of $\Delta T = \{0, 5, 10, 15, 20\}$ and apply $T \leftarrow T + \Delta T$ as the periodicity length parameter. Meanwhile, for each dataset, we compare OneShotSTL with a seasonality shift window of $H=0$ and $H=20$ respectively. Figure \ref{fig:ablation_TSAD_T} shows the performance of OneShotSTL with different $\Delta T$ on four TSAD datasets. As can be seen from Figure \ref{fig:ablation_TSAD_T}, in all cases, OneShotSTL performs better with $H=20$. Moreover, the performance of OneShotSTL degrades with the increase of $\Delta T$ for all four datasets. However, on KDD21 the score of OneShotSTL drops quickly immediately after $\Delta T = 5$, while on the other three datasets the performance of OneShotSTL drops slowly. Further, we evaluate the impact of $T$ on TSF tasks in the same way, with the results shown in Figure \ref{fig:ablation_TSF_T}. From the figure, we can see that the prediction error of OneShotSTL significantly increases on all datasets for both cases with $H=0$ and $H=20$. 
This is probably because we can only correct $T_c$ by searching for the optimal value within $[T-H, T+H]$, which works for TSAD tasks since we only target on the newly arrived data point. However, for TSF tasks, since we cannot correct the future $T_c$, and thus just use $T$ directly to forecast the future data ($\hat{y}_{t+i} = \tau_{t-1}+v_{t \% T + i}$ in Section \ref{sec:extension}).

Next, we evaluate the impact of input parameter $I$ for OneShotSTL. 
Figure \ref{fig:ablation_TSF} shows the MAE of OneShotSTL with $I=8$ and $I=1$ using different forecasting steps with the same $H=20$. Due to limited space, we omit the results of the ablation study of $I$ on TSAD tasks, where we get similar observations. From Figure \ref{fig:ablation_TSF}, we can see that with $I=8$ OneShotSTL more often produces lower prediction errors than that with $I=1$. Specifically, on ETTm2 datasets, the margin is large, while on the other three datasets we observe minor improvements with $I=8$. In general, setting a larger $I$ will improve the quality of decomposition, but sacrifice the speed.

\section{Conclusion}
In this work, we propose an accurate and efficient online STD algorithm OneShotSTL that can decompose time series online with an update complexity of $O(1)$. It can significantly reduce the processing time complexity of the existing methods that often require $O(T)$, e.g., OnlineSTL~\cite{DBLP:journals/pvldb/MishraSZ22}. Resultantly, the online update time of a single time point using OneShotSTL is more than $1,000$ times faster than the batch STD methods, with accuracy comparable with the best counterparts, e.g., RobustSTL~\cite{DBLP:journals/corr/abs-1812-01767}. Extensive experiments on real-world benchmark datasets for downstream tasks, e.g., TSAD and TSF problems, demonstrate that OneShotSTL can be ranging from $10$ to more than $1000$ times faster than the state-of-the-art TSAD and TSF methods while still providing comparable or even better results.

The method has its limitations. The proposed OneShotSTL 
assumes the underlying $T_c\in [T-H, T+H]$ for two constants $T$ and $H$. This may not hold especially when $T_c$ drastically varies across multiple cycles. Adaptively detecting the seasonal cycle $T_c$ online would be of great importance for the online STD problem. In addition, handling missing points or irregularly sampled time series is crucial in practice, which cannot be handled by most of the current STD methods. 
In the future, we plan to explore the potentials of OneShotSTL for TSF tasks since decomposed trend and seasonal signals could help long sequence forecasting when combined with other techniques. Moreover, integrating OneShotSTL with matrix profile-based methods for online TSAD tasks is also an interesting direction. All these lead to the future research directions of the online STD problem.


\newpage
\bibliographystyle{ACM-Reference-Format}
\bibliography{main}


\begin{thebibliography}{47}


\ifx \showCODEN    \undefined \def \showCODEN     #1{\unskip}     \fi
\ifx \showDOI      \undefined \def \showDOI       #1{#1}\fi
\ifx \showISBNx    \undefined \def \showISBNx     #1{\unskip}     \fi
\ifx \showISBNxiii \undefined \def \showISBNxiii  #1{\unskip}     \fi
\ifx \showISSN     \undefined \def \showISSN      #1{\unskip}     \fi
\ifx \showLCCN     \undefined \def \showLCCN      #1{\unskip}     \fi
\ifx \shownote     \undefined \def \shownote      #1{#1}          \fi
\ifx \showarticletitle \undefined \def \showarticletitle #1{#1}   \fi
\ifx \showURL      \undefined \def \showURL       {\relax}        \fi
\providecommand\bibfield[2]{#2}
\providecommand\bibinfo[2]{#2}
\providecommand\natexlab[1]{#1}
\providecommand\showeprint[2][]{arXiv:#2}

\bibitem[\protect\citeauthoryear{??}{clo}{[n.d.]a}]%
        {cloudmonitor}
 \bibinfo{year}{[n.d.]}\natexlab{a}.
\newblock \bibinfo{title}{{Alibaba Cloudmonitor}}.
\newblock
  \bibinfo{howpublished}{https://www.alibabacloud.com/product/cloud-monitor}.
\newblock
\newblock
\shownote{Last accessed 2022-09-21.}


\bibitem[\protect\citeauthoryear{??}{clo}{[n.d.]b}]%
        {cloudwatch}
 \bibinfo{year}{[n.d.]}\natexlab{b}.
\newblock \bibinfo{title}{{AWS Cloudwatch}}.
\newblock \bibinfo{howpublished}{https://aws.amazon.com/cloudwatch/}.
\newblock
\newblock
\shownote{Last accessed 2022-09-21.}


\bibitem[\protect\citeauthoryear{??}{fin}{[n.d.]}]%
        {findlength}
 \bibinfo{year}{[n.d.]}\natexlab{}.
\newblock \bibinfo{title}{{Find length function of TSB-UAD}}.
\newblock
  \bibinfo{howpublished}{\url{https://github.com/TheDatumOrg/TSB-UAD/blob/main/TSB_UAD/utils/slidingWindows.py}}.
\newblock
\newblock
\shownote{Last accessed 2022-12-05.}


\bibitem[\protect\citeauthoryear{??}{aio}{[n.d.]}]%
        {aiops}
 \bibinfo{year}{[n.d.]}\natexlab{}.
\newblock \bibinfo{title}{{How to Get Started With AIOps}}.
\newblock
  \bibinfo{howpublished}{https://www.gartner.com/smarterwithgartner/how-to-get-started-with-aiops}.
\newblock
\newblock
\shownote{Last accessed 2022-09-21.}


\bibitem[\protect\citeauthoryear{??}{tsf}{[n.d.]}]%
        {tsfdata}
 \bibinfo{year}{[n.d.]}\natexlab{}.
\newblock \bibinfo{title}{{Informer Datasets}}.
\newblock
  \bibinfo{howpublished}{\url{https://github.com/zhouhaoyi/Informer2020}}.
\newblock
\newblock
\shownote{Last accessed 2022-12-05.}


\bibitem[\protect\citeauthoryear{??}{jav}{[n.d.]}]%
        {javastl}
 \bibinfo{year}{[n.d.]}\natexlab{}.
\newblock \bibinfo{title}{{Java implementaion of STL in stl-decomp-4j
  package}}.
\newblock
  \bibinfo{howpublished}{\url{https://github.com/ServiceNow/stl-decomp-4j}}.
\newblock
\newblock
\shownote{Last accessed 2022-09-21.}


\bibitem[\protect\citeauthoryear{??}{sre}{[n.d.]}]%
        {sreworks}
 \bibinfo{year}{[n.d.]}\natexlab{}.
\newblock \bibinfo{title}{{Python implementation of FastRobustSTL in
  SREWorks}}.
\newblock \bibinfo{howpublished}{\url{https://github.com/alibaba/SREWorks/}}.
\newblock
\newblock
\shownote{Last accessed 2022-09-21.}


\bibitem[\protect\citeauthoryear{??}{pyt}{[n.d.]}]%
        {pytorch-forecast}
 \bibinfo{year}{[n.d.]}\natexlab{}.
\newblock \bibinfo{title}{{Pytorch-forecasting package}}.
\newblock
  \bibinfo{howpublished}{\url{https://pytorch-forecasting.readthedocs.io/en/stable/index.html}}.
\newblock
\newblock
\shownote{Last accessed 2022-09-21.}


\bibitem[\protect\citeauthoryear{??}{sta}{[n.d.]}]%
        {statsforecast}
 \bibinfo{year}{[n.d.]}\natexlab{}.
\newblock \bibinfo{title}{{Statsforecast package}}.
\newblock
  \bibinfo{howpublished}{\url{https://nixtla.github.io/statsforecast/}}.
\newblock
\newblock
\shownote{Last accessed 2022-09-21.}


\bibitem[\protect\citeauthoryear{??}{sup}{[n.d.]}]%
        {supp}
 \bibinfo{year}{[n.d.]}\natexlab{}.
\newblock \bibinfo{title}{{Supplementary material}}.
\newblock \bibinfo{howpublished}{\url{https://github.com/xiao-he/OneShotSTL/}}.
\newblock
\newblock
\shownote{Last accessed 2023-02-17.}


\bibitem[\protect\citeauthoryear{Ariyo, Adewumi, and Ayo}{Ariyo
  et~al\mbox{.}}{2014}]%
        {10.1109/UKSim.2014.67}
\bibfield{author}{\bibinfo{person}{Adebiyi~A. Ariyo},
  \bibinfo{person}{Adewumi~O. Adewumi}, {and} \bibinfo{person}{Charles~K.
  Ayo}.} \bibinfo{year}{2014}\natexlab{}.
\newblock \showarticletitle{Stock Price Prediction Using the ARIMA Model}. In
  \bibinfo{booktitle}{\emph{Proceedings of the 2014 UKSim-AMSS 16th
  International Conference on Computer Modelling and Simulation}}
  \emph{(\bibinfo{series}{UKSIM '14})}. \bibinfo{publisher}{IEEE Computer
  Society}, \bibinfo{address}{USA}, \bibinfo{pages}{106–112}.
\newblock
\showISBNx{9781479949229}
\urldef\tempurl%
\url{https://doi.org/10.1109/UKSim.2014.67}
\showDOI{\tempurl}


\bibitem[\protect\citeauthoryear{Audibert, Michiardi, Guyard, Marti, and
  Zuluaga}{Audibert et~al\mbox{.}}{2020}]%
        {DBLP:conf/kdd/AudibertMGMZ20}
\bibfield{author}{\bibinfo{person}{Julien Audibert}, \bibinfo{person}{Pietro
  Michiardi}, \bibinfo{person}{Fr{\'{e}}d{\'{e}}ric Guyard},
  \bibinfo{person}{S{\'{e}}bastien Marti}, {and} \bibinfo{person}{Maria~A.
  Zuluaga}.} \bibinfo{year}{2020}\natexlab{}.
\newblock \showarticletitle{{USAD:} UnSupervised Anomaly Detection on
  Multivariate Time Series}. In \bibinfo{booktitle}{\emph{{KDD} '20: The 26th
  {ACM} {SIGKDD} Conference on Knowledge Discovery and Data Mining, Virtual
  Event, CA, USA, August 23-27, 2020}},
  \bibfield{editor}{\bibinfo{person}{Rajesh Gupta}, \bibinfo{person}{Yan Liu},
  \bibinfo{person}{Jiliang Tang}, {and} \bibinfo{person}{B.~Aditya Prakash}}
  (Eds.). \bibinfo{publisher}{{ACM}}, \bibinfo{pages}{3395--3404}.
\newblock
\urldef\tempurl%
\url{https://doi.org/10.1145/3394486.3403392}
\showDOI{\tempurl}


\bibitem[\protect\citeauthoryear{Beck}{Beck}{2015}]%
        {beck2015convergence}
\bibfield{author}{\bibinfo{person}{Amir Beck}.}
  \bibinfo{year}{2015}\natexlab{}.
\newblock \showarticletitle{On the convergence of alternating minimization for
  convex programming with applications to iteratively reweighted least squares
  and decomposition schemes}.
\newblock \bibinfo{journal}{\emph{SIAM Journal on Optimization}}
  \bibinfo{volume}{25}, \bibinfo{number}{1} (\bibinfo{year}{2015}),
  \bibinfo{pages}{185--209}.
\newblock


\bibitem[\protect\citeauthoryear{Boniol, Linardi, Roncallo, Palpanas, Meftah,
  and Remy}{Boniol et~al\mbox{.}}{2021a}]%
        {DBLP:journals/vldb/BoniolLRPMR21}
\bibfield{author}{\bibinfo{person}{Paul Boniol}, \bibinfo{person}{Michele
  Linardi}, \bibinfo{person}{Federico Roncallo}, \bibinfo{person}{Themis
  Palpanas}, \bibinfo{person}{Mohammed Meftah}, {and} \bibinfo{person}{Emmanuel
  Remy}.} \bibinfo{year}{2021}\natexlab{a}.
\newblock \showarticletitle{Unsupervised and scalable subsequence anomaly
  detection in large data series}.
\newblock \bibinfo{journal}{\emph{{VLDB} J.}} \bibinfo{volume}{30},
  \bibinfo{number}{6} (\bibinfo{year}{2021}), \bibinfo{pages}{909--931}.
\newblock
\urldef\tempurl%
\url{https://doi.org/10.1007/s00778-021-00655-8}
\showDOI{\tempurl}


\bibitem[\protect\citeauthoryear{Boniol, Paparrizos, Palpanas, and
  Franklin}{Boniol et~al\mbox{.}}{2021b}]%
        {DBLP:journals/pvldb/BoniolPPF21}
\bibfield{author}{\bibinfo{person}{Paul Boniol}, \bibinfo{person}{John
  Paparrizos}, \bibinfo{person}{Themis Palpanas}, {and}
  \bibinfo{person}{Michael~J. Franklin}.} \bibinfo{year}{2021}\natexlab{b}.
\newblock \showarticletitle{{SAND:} Streaming Subsequence Anomaly Detection}.
\newblock \bibinfo{journal}{\emph{Proc. {VLDB} Endow.}} \bibinfo{volume}{14},
  \bibinfo{number}{10} (\bibinfo{year}{2021}), \bibinfo{pages}{1717--1729}.
\newblock
\urldef\tempurl%
\url{https://doi.org/10.14778/3467861.3467863}
\showDOI{\tempurl}


\bibitem[\protect\citeauthoryear{Cleveland, Cleveland, McRae, and
  Terpenning}{Cleveland et~al\mbox{.}}{1990}]%
        {cleveland90}
\bibfield{author}{\bibinfo{person}{Robert~B. Cleveland},
  \bibinfo{person}{William~S. Cleveland}, \bibinfo{person}{Jean~E. McRae},
  {and} \bibinfo{person}{Irma Terpenning}.} \bibinfo{year}{1990}\natexlab{}.
\newblock \showarticletitle{STL: A Seasonal-Trend Decomposition Procedure Based
  on Loess (with Discussion)}.
\newblock \bibinfo{journal}{\emph{Journal of Official Statistics}}
  \bibinfo{volume}{6} (\bibinfo{year}{1990}), \bibinfo{pages}{3--73}.
\newblock


\bibitem[\protect\citeauthoryear{{Dau, Hoang Anh and Keogh, Eamonn and Kamgar,
  Kaveh and Yeh, Chin-Chia Michael and Zhu, Yan and Gharghabi, Shaghayegh and
  Ratanamahatana, Chotirat Ann and Yanping and Hu, Bing and Begum, Nurjahan and
  Bagnall, Anthony and Mueen, Abdullah and Batista, Gustavo, and
  Hexagon-ML}}{{Dau, Hoang Anh and Keogh, Eamonn and Kamgar, Kaveh and Yeh,
  Chin-Chia Michael and Zhu, Yan and Gharghabi, Shaghayegh and Ratanamahatana,
  Chotirat Ann and Yanping and Hu, Bing and Begum, Nurjahan and Bagnall,
  Anthony and Mueen, Abdullah and Batista, Gustavo, and Hexagon-ML}}{2018}]%
        {UCRArchive2018}
\bibfield{author}{\bibinfo{person}{{Dau, Hoang Anh and Keogh, Eamonn and
  Kamgar, Kaveh and Yeh, Chin-Chia Michael and Zhu, Yan and Gharghabi,
  Shaghayegh and Ratanamahatana, Chotirat Ann and Yanping and Hu, Bing and
  Begum, Nurjahan and Bagnall, Anthony and Mueen, Abdullah and Batista,
  Gustavo, and Hexagon-ML}}.} \bibinfo{year}{2018}\natexlab{}.
\newblock \bibinfo{title}{The UCR Time Series Archive}.
\newblock
\newblock
\newblock
\shownote{\url{https://www.cs.ucr.edu/~eamonn/time_series_data_2018/}.}


\bibitem[\protect\citeauthoryear{De~Livera, Hyndman, and Snyder}{De~Livera
  et~al\mbox{.}}{2011}]%
        {de2011forecasting}
\bibfield{author}{\bibinfo{person}{Alysha~M De~Livera}, \bibinfo{person}{Rob~J
  Hyndman}, {and} \bibinfo{person}{Ralph~D Snyder}.}
  \bibinfo{year}{2011}\natexlab{}.
\newblock \showarticletitle{Forecasting time series with complex seasonal
  patterns using exponential smoothing}.
\newblock \bibinfo{journal}{\emph{Journal of the American statistical
  association}} \bibinfo{volume}{106}, \bibinfo{number}{496}
  (\bibinfo{year}{2011}), \bibinfo{pages}{1513--1527}.
\newblock


\bibitem[\protect\citeauthoryear{Dokumentov, Hyndman, et~al\mbox{.}}{Dokumentov
  et~al\mbox{.}}{2015}]%
        {dokumentov2015str}
\bibfield{author}{\bibinfo{person}{Alexander Dokumentov},
  \bibinfo{person}{Rob~J Hyndman}, {et~al\mbox{.}}}
  \bibinfo{year}{2015}\natexlab{}.
\newblock \showarticletitle{STR: A seasonal-trend decomposition procedure based
  on regression}.
\newblock \bibinfo{journal}{\emph{Monash University: Melbourne, Australia}}
  (\bibinfo{year}{2015}), \bibinfo{pages}{23}.
\newblock


\bibitem[\protect\citeauthoryear{Galati, Audibert, and Zuluaga}{Galati
  et~al\mbox{.}}{[n.d.]}]%
        {usad}
\bibfield{author}{\bibinfo{person}{Francesco Galati}, \bibinfo{person}{Julien
  Audibert}, {and} \bibinfo{person}{Maria~A. Zuluaga}.}
  \bibinfo{year}{[n.d.]}\natexlab{}.
\newblock \bibinfo{title}{{Python implementation of USAD}}.
\newblock \bibinfo{howpublished}{\url{https://github.com/manigalati/usad}}.
\newblock
\newblock
\shownote{Last accessed 2022-09-21.}


\bibitem[\protect\citeauthoryear{George}{George}{2019}]%
        {iot}
\bibfield{author}{\bibinfo{person}{Sam George}.}
  \bibinfo{year}{2019}\natexlab{}.
\newblock \bibinfo{title}{{IoT Signals report: IoT’s promise will be unlocked
  by addressing skills shortage,complexity and security}}.
\newblock \bibinfo{howpublished}{https://blogs.microsoft.com/blog/2019/07/30/}.
\newblock
\newblock
\shownote{Last accessed 2022-09-21.}


\bibitem[\protect\citeauthoryear{Green}{Green}{1984}]%
        {10.2307/2345503}
\bibfield{author}{\bibinfo{person}{P.~J. Green}.}
  \bibinfo{year}{1984}\natexlab{}.
\newblock \showarticletitle{Iteratively Reweighted Least Squares for Maximum
  Likelihood Estimation, and some Robust and Resistant Alternatives}.
\newblock \bibinfo{journal}{\emph{Journal of the Royal Statistical Society.
  Series B (Methodological)}} \bibinfo{volume}{46}, \bibinfo{number}{2}
  (\bibinfo{year}{1984}), \bibinfo{pages}{149--192}.
\newblock
\showISSN{00359246}
\urldef\tempurl%
\url{http://www.jstor.org/stable/2345503}
\showURL{%
\tempurl}


\bibitem[\protect\citeauthoryear{He, Alesiani, and Shaker}{He
  et~al\mbox{.}}{2019}]%
        {DBLP:conf/aaai/HeAS19}
\bibfield{author}{\bibinfo{person}{Xiao He}, \bibinfo{person}{Francesco
  Alesiani}, {and} \bibinfo{person}{Ammar Shaker}.}
  \bibinfo{year}{2019}\natexlab{}.
\newblock \showarticletitle{Efficient and Scalable Multi-Task Regression on
  Massive Number of Tasks}. In \bibinfo{booktitle}{\emph{The Thirty-Third
  {AAAI} Conference on Artificial Intelligence, {AAAI} 2019, Honolulu, Hawaii,
  USA, January 27 - February 1, 2019}}. \bibinfo{publisher}{{AAAI} Press},
  \bibinfo{pages}{3763--3770}.
\newblock
\urldef\tempurl%
\url{https://doi.org/10.1609/aaai.v33i01.33013763}
\showDOI{\tempurl}


\bibitem[\protect\citeauthoryear{Kim, Choi, Choi, Lee, and Yoon}{Kim
  et~al\mbox{.}}{2022}]%
        {DBLP:conf/aaai/KimCCLY22}
\bibfield{author}{\bibinfo{person}{Siwon Kim}, \bibinfo{person}{Kukjin Choi},
  \bibinfo{person}{Hyun{-}Soo Choi}, \bibinfo{person}{Byunghan Lee}, {and}
  \bibinfo{person}{Sungroh Yoon}.} \bibinfo{year}{2022}\natexlab{}.
\newblock \showarticletitle{Towards a Rigorous Evaluation of Time-Series
  Anomaly Detection}. In \bibinfo{booktitle}{\emph{Thirty-Sixth {AAAI}
  Conference on Artificial Intelligence, {AAAI} 2022, Thirty-Fourth Conference
  on Innovative Applications of Artificial Intelligence, {IAAI} 2022, The
  Twelveth Symposium on Educational Advances in Artificial Intelligence, {EAAI}
  2022 Virtual Event, February 22 - March 1, 2022}}. \bibinfo{publisher}{{AAAI}
  Press}, \bibinfo{pages}{7194--7201}.
\newblock
\urldef\tempurl%
\url{https://ojs.aaai.org/index.php/AAAI/article/view/20680}
\showURL{%
\tempurl}


\bibitem[\protect\citeauthoryear{Kim, Koh, Boyd, and Gorinevsky}{Kim
  et~al\mbox{.}}{2009}]%
        {kim2009ell_1}
\bibfield{author}{\bibinfo{person}{Seung-Jean Kim}, \bibinfo{person}{Kwangmoo
  Koh}, \bibinfo{person}{Stephen Boyd}, {and} \bibinfo{person}{Dimitry
  Gorinevsky}.} \bibinfo{year}{2009}\natexlab{}.
\newblock \showarticletitle{$\backslash$ell\_1 trend filtering}.
\newblock \bibinfo{journal}{\emph{SIAM review}} \bibinfo{volume}{51},
  \bibinfo{number}{2} (\bibinfo{year}{2009}), \bibinfo{pages}{339--360}.
\newblock


\bibitem[\protect\citeauthoryear{Law}{Law}{2019}]%
        {law2019stumpy}
\bibfield{author}{\bibinfo{person}{Sean~M. Law}.}
  \bibinfo{year}{2019}\natexlab{}.
\newblock \showarticletitle{{STUMPY: A Powerful and Scalable Python Library for
  Time Series Data Mining}}.
\newblock \bibinfo{journal}{\emph{{The Journal of Open Source Software}}}
  \bibinfo{volume}{4}, \bibinfo{number}{39} (\bibinfo{year}{2019}),
  \bibinfo{pages}{1504}.
\newblock


\bibitem[\protect\citeauthoryear{Lu, Wu, Mueen, Zuluaga, and Keogh}{Lu
  et~al\mbox{.}}{2022}]%
        {DBLP:conf/kdd/Lu00ZK22}
\bibfield{author}{\bibinfo{person}{Yue Lu}, \bibinfo{person}{Renjie Wu},
  \bibinfo{person}{Abdullah Mueen}, \bibinfo{person}{Maria~A. Zuluaga}, {and}
  \bibinfo{person}{Eamonn~J. Keogh}.} \bibinfo{year}{2022}\natexlab{}.
\newblock \showarticletitle{Matrix Profile {XXIV:} Scaling Time Series Anomaly
  Detection to Trillions of Datapoints and Ultra-fast Arriving Data Streams}.
  In \bibinfo{booktitle}{\emph{{KDD} '22: The 28th {ACM} {SIGKDD} Conference on
  Knowledge Discovery and Data Mining, Washington, DC, USA, August 14 - 18,
  2022}}, \bibfield{editor}{\bibinfo{person}{Aidong Zhang} {and}
  \bibinfo{person}{Huzefa Rangwala}} (Eds.). \bibinfo{publisher}{{ACM}},
  \bibinfo{pages}{1173--1182}.
\newblock
\urldef\tempurl%
\url{https://doi.org/10.1145/3534678.3539271}
\showDOI{\tempurl}


\bibitem[\protect\citeauthoryear{Mishra, Sriharsha, and Zhong}{Mishra
  et~al\mbox{.}}{2022}]%
        {DBLP:journals/pvldb/MishraSZ22}
\bibfield{author}{\bibinfo{person}{Abhinav Mishra}, \bibinfo{person}{Ram
  Sriharsha}, {and} \bibinfo{person}{Sichen Zhong}.}
  \bibinfo{year}{2022}\natexlab{}.
\newblock \showarticletitle{OnlineSTL: Scaling Time Series Decomposition by
  100x}.
\newblock \bibinfo{journal}{\emph{Proc. {VLDB} Endow.}} \bibinfo{volume}{15},
  \bibinfo{number}{7} (\bibinfo{year}{2022}), \bibinfo{pages}{1417--1425}.
\newblock
\urldef\tempurl%
\url{https://www.vldb.org/pvldb/vol15/p1417-mishra.pdf}
\showURL{%
\tempurl}


\bibitem[\protect\citeauthoryear{Oreshkin, Carpov, Chapados, and
  Bengio}{Oreshkin et~al\mbox{.}}{2020}]%
        {DBLP:conf/iclr/OreshkinCCB20}
\bibfield{author}{\bibinfo{person}{Boris~N. Oreshkin}, \bibinfo{person}{Dmitri
  Carpov}, \bibinfo{person}{Nicolas Chapados}, {and} \bibinfo{person}{Yoshua
  Bengio}.} \bibinfo{year}{2020}\natexlab{}.
\newblock \showarticletitle{{N-BEATS:} Neural basis expansion analysis for
  interpretable time series forecasting}. In \bibinfo{booktitle}{\emph{8th
  International Conference on Learning Representations, {ICLR} 2020, Addis
  Ababa, Ethiopia, April 26-30, 2020}}. \bibinfo{publisher}{OpenReview.net}.
\newblock
\urldef\tempurl%
\url{https://openreview.net/forum?id=r1ecqn4YwB}
\showURL{%
\tempurl}


\bibitem[\protect\citeauthoryear{Paparrizos, Boniol, Palpanas, Tsay, Elmore,
  and Franklin}{Paparrizos et~al\mbox{.}}{2022a}]%
        {paparrizos2022volume}
\bibfield{author}{\bibinfo{person}{John Paparrizos}, \bibinfo{person}{Paul
  Boniol}, \bibinfo{person}{Themis Palpanas}, \bibinfo{person}{Ruey~S Tsay},
  \bibinfo{person}{Aaron Elmore}, {and} \bibinfo{person}{Michael~J Franklin}.}
  \bibinfo{year}{2022}\natexlab{a}.
\newblock \showarticletitle{{Volume Under the Surface: A New Accuracy
  Evaluation Measure for Time-Series Anomaly Detection}}.
\newblock \bibinfo{journal}{\emph{Proceedings of the VLDB Endowment}}
  \bibinfo{volume}{15}, \bibinfo{number}{11} (\bibinfo{year}{2022}),
  \bibinfo{pages}{2774--2787}.
\newblock


\bibitem[\protect\citeauthoryear{Paparrizos, Kang, Boniol, Tsay, Palpanas, and
  Franklin}{Paparrizos et~al\mbox{.}}{2022b}]%
        {DBLP:journals/pvldb/PaparrizosKBTPF22}
\bibfield{author}{\bibinfo{person}{John Paparrizos}, \bibinfo{person}{Yuhao
  Kang}, \bibinfo{person}{Paul Boniol}, \bibinfo{person}{Ruey~S. Tsay},
  \bibinfo{person}{Themis Palpanas}, {and} \bibinfo{person}{Michael~J.
  Franklin}.} \bibinfo{year}{2022}\natexlab{b}.
\newblock \showarticletitle{{TSB-UAD:} An End-to-End Benchmark Suite for
  Univariate Time-Series Anomaly Detection}.
\newblock \bibinfo{journal}{\emph{Proc. {VLDB} Endow.}} \bibinfo{volume}{15},
  \bibinfo{number}{8} (\bibinfo{year}{2022}), \bibinfo{pages}{1697--1711}.
\newblock
\urldef\tempurl%
\url{https://www.vldb.org/pvldb/vol15/p1697-paparrizos.pdf}
\showURL{%
\tempurl}


\bibitem[\protect\citeauthoryear{Park, Hoshi, and Kemp}{Park
  et~al\mbox{.}}{2018}]%
        {DBLP:journals/ral/ParkHK18}
\bibfield{author}{\bibinfo{person}{Daehyung Park}, \bibinfo{person}{Yuuna
  Hoshi}, {and} \bibinfo{person}{Charles~C. Kemp}.}
  \bibinfo{year}{2018}\natexlab{}.
\newblock \showarticletitle{A Multimodal Anomaly Detector for Robot-Assisted
  Feeding Using an LSTM-Based Variational Autoencoder}.
\newblock \bibinfo{journal}{\emph{{IEEE} Robotics Autom. Lett.}}
  \bibinfo{volume}{3}, \bibinfo{number}{2} (\bibinfo{year}{2018}),
  \bibinfo{pages}{1544--1551}.
\newblock
\urldef\tempurl%
\url{https://doi.org/10.1109/LRA.2018.2801475}
\showDOI{\tempurl}


\bibitem[\protect\citeauthoryear{Puech, Boussard, D'Amato, and Millerand}{Puech
  et~al\mbox{.}}{2019}]%
        {DBLP:conf/pkdd/PuechBDM19}
\bibfield{author}{\bibinfo{person}{Tom Puech}, \bibinfo{person}{Matthieu
  Boussard}, \bibinfo{person}{Anthony D'Amato}, {and}
  \bibinfo{person}{Ga{\"{e}}tan Millerand}.} \bibinfo{year}{2019}\natexlab{}.
\newblock \showarticletitle{A Fully Automated Periodicity Detection in Time
  Series}. In \bibinfo{booktitle}{\emph{Advanced Analytics and Learning on
  Temporal Data - 4th {ECML} {PKDD} Workshop, {AALTD} 2019, W{\"{u}}rzburg,
  Germany, September 20, 2019, Revised Selected Papers}}
  \emph{(\bibinfo{series}{Lecture Notes in Computer Science})},
  \bibfield{editor}{\bibinfo{person}{Vincent Lemaire}, \bibinfo{person}{Simon
  Malinowski}, \bibinfo{person}{Anthony~J. Bagnall}, \bibinfo{person}{Alexis
  Bondu}, \bibinfo{person}{Thomas Guyet}, {and} \bibinfo{person}{Romain
  Tavenard}} (Eds.), Vol.~\bibinfo{volume}{11986}.
  \bibinfo{publisher}{Springer}, \bibinfo{pages}{43--54}.
\newblock
\urldef\tempurl%
\url{https://doi.org/10.1007/978-3-030-39098-3\_4}
\showDOI{\tempurl}


\bibitem[\protect\citeauthoryear{Salinas, Flunkert, Gasthaus, and
  Januschowski}{Salinas et~al\mbox{.}}{2020}]%
        {salinas2020deepar}
\bibfield{author}{\bibinfo{person}{David Salinas}, \bibinfo{person}{Valentin
  Flunkert}, \bibinfo{person}{Jan Gasthaus}, {and} \bibinfo{person}{Tim
  Januschowski}.} \bibinfo{year}{2020}\natexlab{}.
\newblock \showarticletitle{DeepAR: Probabilistic forecasting with
  autoregressive recurrent networks}.
\newblock \bibinfo{journal}{\emph{International Journal of Forecasting}}
  \bibinfo{volume}{36}, \bibinfo{number}{3} (\bibinfo{year}{2020}),
  \bibinfo{pages}{1181--1191}.
\newblock


\bibitem[\protect\citeauthoryear{Taylor and Letham}{Taylor and Letham}{2017}]%
        {DBLP:journals/peerjpre/TaylorL17}
\bibfield{author}{\bibinfo{person}{Sean~J. Taylor} {and}
  \bibinfo{person}{Benjamin Letham}.} \bibinfo{year}{2017}\natexlab{}.
\newblock \showarticletitle{Forecasting at Scale}.
\newblock \bibinfo{journal}{\emph{PeerJ Prepr.}}  \bibinfo{volume}{5}
  (\bibinfo{year}{2017}), \bibinfo{pages}{e3190}.
\newblock
\urldef\tempurl%
\url{https://doi.org/10.7287/peerj.preprints.3190v1}
\showDOI{\tempurl}


\bibitem[\protect\citeauthoryear{Toller, Santos, and Kern}{Toller
  et~al\mbox{.}}{2019}]%
        {DBLP:journals/datamine/TollerSK19}
\bibfield{author}{\bibinfo{person}{Maximilian Toller}, \bibinfo{person}{Tiago
  Santos}, {and} \bibinfo{person}{Roman Kern}.}
  \bibinfo{year}{2019}\natexlab{}.
\newblock \showarticletitle{{SAZED:} parameter-free domain-agnostic season
  length estimation in time series data}.
\newblock \bibinfo{journal}{\emph{Data Min. Knowl. Discov.}}
  \bibinfo{volume}{33}, \bibinfo{number}{6} (\bibinfo{year}{2019}),
  \bibinfo{pages}{1775--1798}.
\newblock
\urldef\tempurl%
\url{https://doi.org/10.1007/s10618-019-00645-z}
\showDOI{\tempurl}


\bibitem[\protect\citeauthoryear{Tuli}{Tuli}{[n.d.]}]%
        {tranad}
\bibfield{author}{\bibinfo{person}{Shreshth Tuli}.}
  \bibinfo{year}{[n.d.]}\natexlab{}.
\newblock \bibinfo{title}{{Python implementation of TranAD}}.
\newblock
  \bibinfo{howpublished}{\url{https://github.com/imperial-qore/TranAD}}.
\newblock
\newblock
\shownote{Last accessed 2022-09-21.}


\bibitem[\protect\citeauthoryear{Tuli, Casale, and Jennings}{Tuli
  et~al\mbox{.}}{2022}]%
        {DBLP:journals/pvldb/TuliCJ22}
\bibfield{author}{\bibinfo{person}{Shreshth Tuli}, \bibinfo{person}{Giuliano
  Casale}, {and} \bibinfo{person}{Nicholas~R. Jennings}.}
  \bibinfo{year}{2022}\natexlab{}.
\newblock \showarticletitle{TranAD: Deep Transformer Networks for Anomaly
  Detection in Multivariate Time Series Data}.
\newblock \bibinfo{journal}{\emph{Proc. {VLDB} Endow.}} \bibinfo{volume}{15},
  \bibinfo{number}{6} (\bibinfo{year}{2022}), \bibinfo{pages}{1201--1214}.
\newblock
\urldef\tempurl%
\url{https://www.vldb.org/pvldb/vol15/p1201-tuli.pdf}
\showURL{%
\tempurl}


\bibitem[\protect\citeauthoryear{Vlachos, Yu, and Castelli}{Vlachos
  et~al\mbox{.}}{2005}]%
        {DBLP:conf/sdm/VlachosYC05}
\bibfield{author}{\bibinfo{person}{Michail Vlachos}, \bibinfo{person}{Philip~S.
  Yu}, {and} \bibinfo{person}{Vittorio Castelli}.}
  \bibinfo{year}{2005}\natexlab{}.
\newblock \showarticletitle{On Periodicity Detection and Structural Periodic
  Similarity}. In \bibinfo{booktitle}{\emph{Proceedings of the 2005 {SIAM}
  International Conference on Data Mining, {SDM} 2005, Newport Beach, CA, USA,
  April 21-23, 2005}}, \bibfield{editor}{\bibinfo{person}{Hillol Kargupta},
  \bibinfo{person}{Jaideep Srivastava}, \bibinfo{person}{Chandrika Kamath},
  {and} \bibinfo{person}{Arnold Goodman}} (Eds.). \bibinfo{publisher}{{SIAM}},
  \bibinfo{pages}{449--460}.
\newblock
\urldef\tempurl%
\url{https://doi.org/10.1137/1.9781611972757.40}
\showDOI{\tempurl}


\bibitem[\protect\citeauthoryear{Wen, Gao, Song, Sun, Xu, and Zhu}{Wen
  et~al\mbox{.}}{2018}]%
        {DBLP:journals/corr/abs-1812-01767}
\bibfield{author}{\bibinfo{person}{Qingsong Wen}, \bibinfo{person}{Jingkun
  Gao}, \bibinfo{person}{Xiaomin Song}, \bibinfo{person}{Liang Sun},
  \bibinfo{person}{Huan Xu}, {and} \bibinfo{person}{Shenghuo Zhu}.}
  \bibinfo{year}{2018}\natexlab{}.
\newblock \showarticletitle{RobustSTL: {A} Robust Seasonal-Trend Decomposition
  Algorithm for Long Time Series}.
\newblock \bibinfo{journal}{\emph{CoRR}}  \bibinfo{volume}{abs/1812.01767}
  (\bibinfo{year}{2018}).
\newblock
\showeprint[arXiv]{1812.01767}
\urldef\tempurl%
\url{http://arxiv.org/abs/1812.01767}
\showURL{%
\tempurl}


\bibitem[\protect\citeauthoryear{Wen, He, Sun, Zhang, Ke, and Xu}{Wen
  et~al\mbox{.}}{2021}]%
        {DBLP:conf/sigmod/WenH0ZKX21}
\bibfield{author}{\bibinfo{person}{Qingsong Wen}, \bibinfo{person}{Kai He},
  \bibinfo{person}{Liang Sun}, \bibinfo{person}{Yingying Zhang},
  \bibinfo{person}{Min Ke}, {and} \bibinfo{person}{Huan Xu}.}
  \bibinfo{year}{2021}\natexlab{}.
\newblock \showarticletitle{RobustPeriod: Robust Time-Frequency Mining for
  Multiple Periodicity Detection}. In \bibinfo{booktitle}{\emph{{SIGMOD} '21:
  International Conference on Management of Data, Virtual Event, China, June
  20-25, 2021}}, \bibfield{editor}{\bibinfo{person}{Guoliang Li},
  \bibinfo{person}{Zhanhuai Li}, \bibinfo{person}{Stratos Idreos}, {and}
  \bibinfo{person}{Divesh Srivastava}} (Eds.). \bibinfo{publisher}{{ACM}},
  \bibinfo{pages}{2328--2337}.
\newblock
\urldef\tempurl%
\url{https://doi.org/10.1145/3448016.3452779}
\showDOI{\tempurl}


\bibitem[\protect\citeauthoryear{Wen, Zhang, Li, and Sun}{Wen
  et~al\mbox{.}}{2020}]%
        {DBLP:conf/kdd/WenZL020}
\bibfield{author}{\bibinfo{person}{Qingsong Wen}, \bibinfo{person}{Zhe Zhang},
  \bibinfo{person}{Yan Li}, {and} \bibinfo{person}{Liang Sun}.}
  \bibinfo{year}{2020}\natexlab{}.
\newblock \showarticletitle{Fast RobustSTL: Efficient and Robust Seasonal-Trend
  Decomposition for Time Series with Complex Patterns}. In
  \bibinfo{booktitle}{\emph{{KDD} '20: The 26th {ACM} {SIGKDD} Conference on
  Knowledge Discovery and Data Mining, Virtual Event, CA, USA, August 23-27,
  2020}}, \bibfield{editor}{\bibinfo{person}{Rajesh Gupta},
  \bibinfo{person}{Yan Liu}, \bibinfo{person}{Jiliang Tang}, {and}
  \bibinfo{person}{B.~Aditya Prakash}} (Eds.). \bibinfo{publisher}{{ACM}},
  \bibinfo{pages}{2203--2213}.
\newblock
\urldef\tempurl%
\url{https://doi.org/10.1145/3394486.3403271}
\showDOI{\tempurl}


\bibitem[\protect\citeauthoryear{Wu, Mueen, Zuluaga, and Keogh}{Wu
  et~al\mbox{.}}{[n.d.]}]%
        {damp}
\bibfield{author}{\bibinfo{person}{Renjie Wu}, \bibinfo{person}{Abdullah
  Mueen}, \bibinfo{person}{Maria~A. Zuluaga}, {and} \bibinfo{person}{Eamonn~J.
  Keogh}.} \bibinfo{year}{[n.d.]}\natexlab{}.
\newblock \bibinfo{title}{{Matlab implementaion of DAMP}}.
\newblock
  \bibinfo{howpublished}{\url{https://sites.google.com/view/discord-aware-matrix-profile}}.
\newblock
\newblock
\shownote{Last accessed 2022-09-21.}


\bibitem[\protect\citeauthoryear{Yeh, Zhu, Ulanova, Begum, Ding, Dau,
  Zimmerman, Silva, Mueen, and Keogh}{Yeh et~al\mbox{.}}{2018}]%
        {DBLP:journals/datamine/YehZUBDDZSMK18}
\bibfield{author}{\bibinfo{person}{Chin{-}Chia~Michael Yeh},
  \bibinfo{person}{Yan Zhu}, \bibinfo{person}{Liudmila Ulanova},
  \bibinfo{person}{Nurjahan Begum}, \bibinfo{person}{Yifei Ding},
  \bibinfo{person}{Hoang~Anh Dau}, \bibinfo{person}{Zachary Zimmerman},
  \bibinfo{person}{Diego~Furtado Silva}, \bibinfo{person}{Abdullah Mueen},
  {and} \bibinfo{person}{Eamonn~J. Keogh}.} \bibinfo{year}{2018}\natexlab{}.
\newblock \showarticletitle{Time series joins, motifs, discords and shapelets:
  a unifying view that exploits the matrix profile}.
\newblock \bibinfo{journal}{\emph{Data Min. Knowl. Discov.}}
  \bibinfo{volume}{32}, \bibinfo{number}{1} (\bibinfo{year}{2018}),
  \bibinfo{pages}{83--123}.
\newblock
\urldef\tempurl%
\url{https://doi.org/10.1007/s10618-017-0519-9}
\showDOI{\tempurl}


\bibitem[\protect\citeauthoryear{Zhou, Zhang, Peng, Zhang, Li, Xiong, and
  Zhang}{Zhou et~al\mbox{.}}{2021}]%
        {zhou2021informer}
\bibfield{author}{\bibinfo{person}{Haoyi Zhou}, \bibinfo{person}{Shanghang
  Zhang}, \bibinfo{person}{Jieqi Peng}, \bibinfo{person}{Shuai Zhang},
  \bibinfo{person}{Jianxin Li}, \bibinfo{person}{Hui Xiong}, {and}
  \bibinfo{person}{Wancai Zhang}.} \bibinfo{year}{2021}\natexlab{}.
\newblock \showarticletitle{Informer: Beyond efficient transformer for long
  sequence time-series forecasting}. In \bibinfo{booktitle}{\emph{Proceedings
  of the AAAI Conference on Artificial Intelligence}},
  Vol.~\bibinfo{volume}{35}. \bibinfo{pages}{11106--11115}.
\newblock


\bibitem[\protect\citeauthoryear{Zhou, Ma, Wang, Wen, Sun, Yao, Yin, and
  Jin}{Zhou et~al\mbox{.}}{2022a}]%
        {DBLP:journals/corr/abs-2205-08897}
\bibfield{author}{\bibinfo{person}{Tian Zhou}, \bibinfo{person}{Ziqing Ma},
  \bibinfo{person}{Xue Wang}, \bibinfo{person}{Qingsong Wen},
  \bibinfo{person}{Liang Sun}, \bibinfo{person}{Tao Yao},
  \bibinfo{person}{Wotao Yin}, {and} \bibinfo{person}{Rong Jin}.}
  \bibinfo{year}{2022}\natexlab{a}.
\newblock \showarticletitle{FiLM: Frequency improved Legendre Memory Model for
  Long-term Time Series Forecasting}.
\newblock \bibinfo{journal}{\emph{CoRR}}  \bibinfo{volume}{abs/2205.08897}
  (\bibinfo{year}{2022}).
\newblock
\urldef\tempurl%
\url{https://doi.org/10.48550/arXiv.2205.08897}
\showDOI{\tempurl}
\showeprint[arXiv]{2205.08897}


\bibitem[\protect\citeauthoryear{Zhou, Ma, Wen, Wang, Sun, and Jin}{Zhou
  et~al\mbox{.}}{2022b}]%
        {DBLP:conf/icml/ZhouMWW0022}
\bibfield{author}{\bibinfo{person}{Tian Zhou}, \bibinfo{person}{Ziqing Ma},
  \bibinfo{person}{Qingsong Wen}, \bibinfo{person}{Xue Wang},
  \bibinfo{person}{Liang Sun}, {and} \bibinfo{person}{Rong Jin}.}
  \bibinfo{year}{2022}\natexlab{b}.
\newblock \showarticletitle{FEDformer: Frequency Enhanced Decomposed
  Transformer for Long-term Series Forecasting}. In
  \bibinfo{booktitle}{\emph{International Conference on Machine Learning,
  {ICML} 2022, 17-23 July 2022, Baltimore, Maryland, {USA}}}
  \emph{(\bibinfo{series}{Proceedings of Machine Learning Research})},
  \bibfield{editor}{\bibinfo{person}{Kamalika Chaudhuri},
  \bibinfo{person}{Stefanie Jegelka}, \bibinfo{person}{Le~Song},
  \bibinfo{person}{Csaba Szepesv{\'{a}}ri}, \bibinfo{person}{Gang Niu}, {and}
  \bibinfo{person}{Sivan Sabato}} (Eds.), Vol.~\bibinfo{volume}{162}.
  \bibinfo{publisher}{{PMLR}}, \bibinfo{pages}{27268--27286}.
\newblock


\end{thebibliography}

\end{document}